%% file: rob-fingernail.tex
\documentclass{rob}%

\usepackage{color}
\usepackage[colorlinks]{hyperref}
\usepackage{subfigure}
\usepackage{makecell}
\usepackage{multirow}
\usepackage{tabularx}
\usepackage{caption2}
\usepackage{todonotes}
\usepackage{CJK}
\usepackage[super]{cite}
\usepackage{amsmath}
\usepackage{amsfonts}

\usepackage{xcolor}

\begin{document}

\title[Estimating Fingertip Forces, Torques, and Local Curvatures from Fingernail Images]{Estimating Fingertip Forces, Torques, and Local Curvatures from Fingernail Images}

\author{Nutan Chen$\dagger$\thanks{Corresponding author. E-mail:
nutan.chen@gmail.com}, G\"oran Westling$\ddagger$, Benoni B. Edin$\ddagger$ and Patrick van der Smagt$\dagger$}
\affil{$\dagger$Machine Learning Research Lab, Volkswagen Group, Germany\\
$\ddagger$Department of Integrative Medical Biology, Physiology Section, Ume\aa\ University, Sweden}

\ADaccepted{MONTH DAY, YEAR. First published online: MONTH DAY, YEAR}

\maketitle

\graphicspath{{figures/}}

\input{sections/abstract}
\input{sections/introduction}
\input{sections/alignment}
\input{sections/predictor}

\input{sections/results}

\input{sections/Conclusions}

\section*{Acknowledgements}
This work has been supported in part by the Swedish Research Council, VR 2011-3128.

\clearpage
\newpage
\pagebreak


\end{document}

%% file: sections/abstract.tex
\begin{summary}

The study of dexterous manipulation has provided important insights in humans sensorimotor control as well as inspiration for manipulation strategies in robotic hands. 
Previous work focused on experimental environment with restrictions.
Here we describe a method using the deformation and color distribution of the fingernail and its surrounding skin, to estimate the fingertip forces, torques and contact surface curvatures for various objects, including the shape and material of the contact surfaces and the weight of the objects. The proposed method circumvents limitations associated with sensorized objects, gloves or fixed contact surface type. 
In addition, compared with previous single finger estimation in an experimental environment, we extend the approach to multiple finger force estimation, which can be used for applications such as human grasping analysis.
Four algorithms are used, c.q., Gaussian process (GP), Convolutional Neural Networks (CNN), Neural Networks with Fast Dropout (NN-FD) and Recurrent Neural Networks with Fast Dropout (RNN-FD), to model a mapping from images to the corresponding labels. The results further show that the proposed method has high accuracy to predict force, torque and contact surface.

\end{summary}

\begin{keywords}
Fingertip forces; Machine learning; Image processing; Fingernail images.
\end{keywords}

%% file: sections/introduction.tex
\section{Introduction}

When humans grasp and manipulate objects, control of the fingertip forces is critical \cite{Johansson240765,Panarese405710}. While the positioning of fingertips on objects are important, the generated force vectors not only define the manipulative action but are strictly controlled to ensure grasp stability.
Measuring these forces is, however, prohibitively difficult. Attaching measurements instruments to objects is an obvious solution but makes it practically impossible to include more than a few objects in experiments and also restricts the position of contact between fingertips and objects. While gloves with force sensors at the fingers make it possible to use a range of objects but with the obvious drawback that tactile sensibility is impeded. 

\begin{figure}
	\centerline{\includegraphics[height=3.5cm]{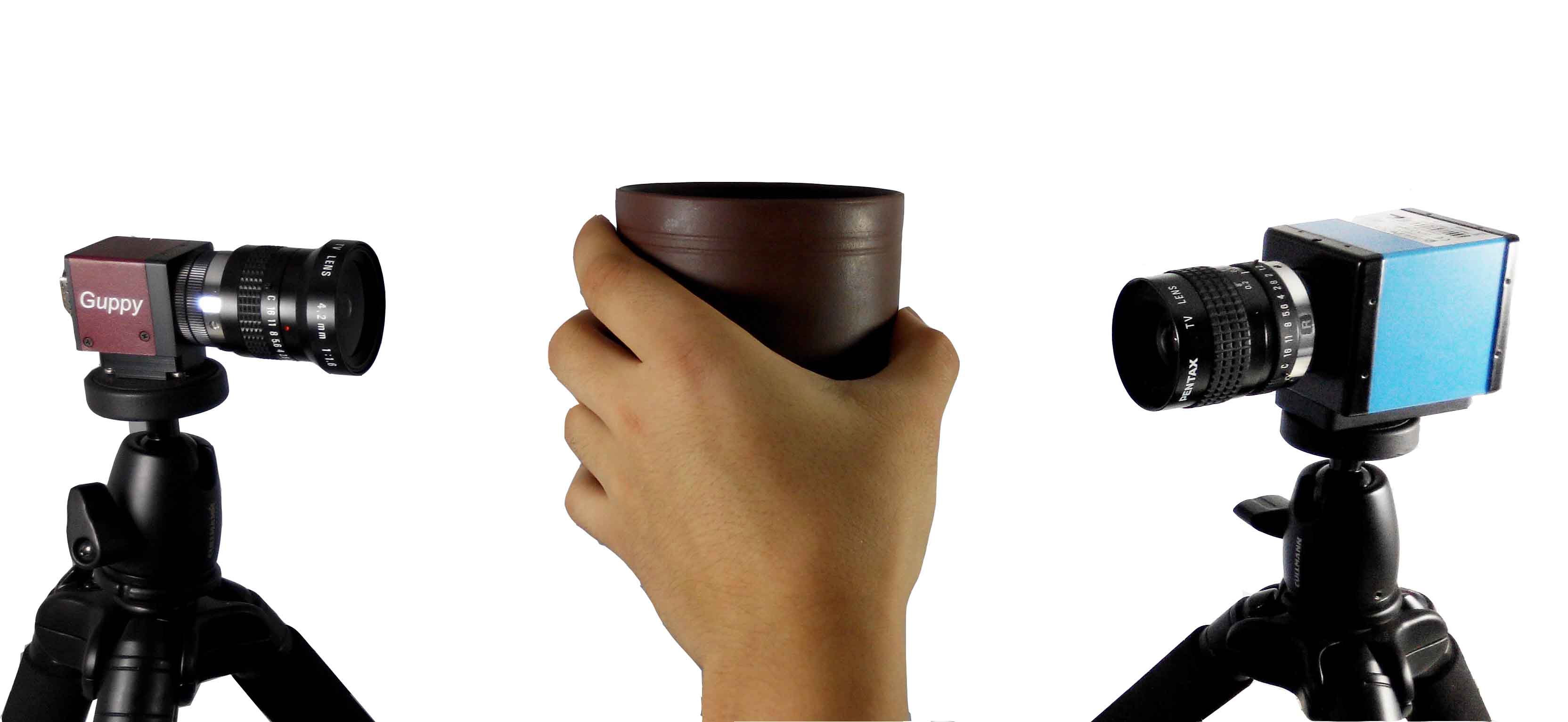}}
	\caption{Estimation of the finger force and torque from videos.}\label{fig:grip}
\end{figure}

The starting point of the solution we propose is the changes in nail coloration that are obvious when a fingertip applies different forces and torques to a contact surface.
Following up on the seminal work by Mascaro \textsl{et al.}\ (see, e.g., [\citen{Sun2009}]), we use steady color cameras to observe the nails of the fingers while in contact with an object (Fig.~\ref{fig:grip}), and learn the relationship between 
the nail coloration, the prevailing force vector, and the curvature of the contact surface.
We have investigated these in three previous publications \cite{2013Sebastian,chen2014a,chen2015a}.

Our method does not require mounting a sensor at the finger or the 
object. Instead, 
we localize the finger in an 
image, perform appropriate image transformations and extractions, and predict fingertip force and torque along with the curvature of the contact surface.
A crucial aspect therefore is image alignment. Prior methods of fingernail image registration are 2D-to-3D registration with a grid pattern and fiducial markers onto the finger \cite{Sun2008}, 2D-to-3D registration using Convolutional Neural Networks \cite{chen2014a}, rigid body transformation including the Harris feature point based method \cite{Sun2009}, Canny edge detection \cite{Grieve2010}, template matching using markers \cite{2013Sebastian}, non-rigid registration fitting a finger model \cite{Sun2009}, and Active Appearance Models (AAM)\cite{grieve2016optimizing}. Other methods use sensors mounted on the finger \cite{2013Sebastian} or require restrictions such as a bracket to support the hand \cite{Sun2008} or the finger \cite{Grieve2010}. 
Analyses of "natural" human grasping requires, however, a robust and generally applicable system that does not obstruct movements or otherwise interferes with human performance. To address this challenge, we align the images using non-rigid alignment techniques \cite{tmi_S10}.

Following preprocessing, various methods have been developed to estimate the force. Model-based methods \cite{tomas2013} contain linearized sigmoid models, EigenNail models\cite{Sun2007}, and a linear least squared method (LSM) \cite{yoshimoto2018estimation}. [\citen{2013Sebastian}] estimates force and torque using Gaussian processes (GP) and neural networks. 
Given
the high accuracy obtained in [\citen{2013Sebastian,chen2014a,chen2015a}] we use Gaussian processes to estimate force from the aligned images in this paper. We also explore other methods for the estimation such as Neural Networks, Convolutional Neural Networks and Recurrent Neural Networks.

Moving away from a constrained lab setting with perfect conditions and comfortable restrictions -- e.g., a finger brace, a fixed force/torque sensor on a table, or a fixed finger location w.r.t. the camera \cite{grieve2016optimizing, yoshimoto2018estimation} -- 
we extend this method to a more universal environment. Prior works focused on a certain contact surface material, shape, recording time and object weight \cite{grieve2016optimizing,yoshimoto2018estimation}; in this study, with these variables, we find our approach performs accurately. The results show the first application of  
video-based finger force prediction. 

%% file: sections/alignment.tex

\section{Methods}

\subsection{Setup}

\begin{figure*}
\centering
  \subfigure[Recording system.]{
      \label{setup01}
    \begin{minipage}[b]{0.5\textwidth}
      \centering
      \includegraphics[width=\textwidth]{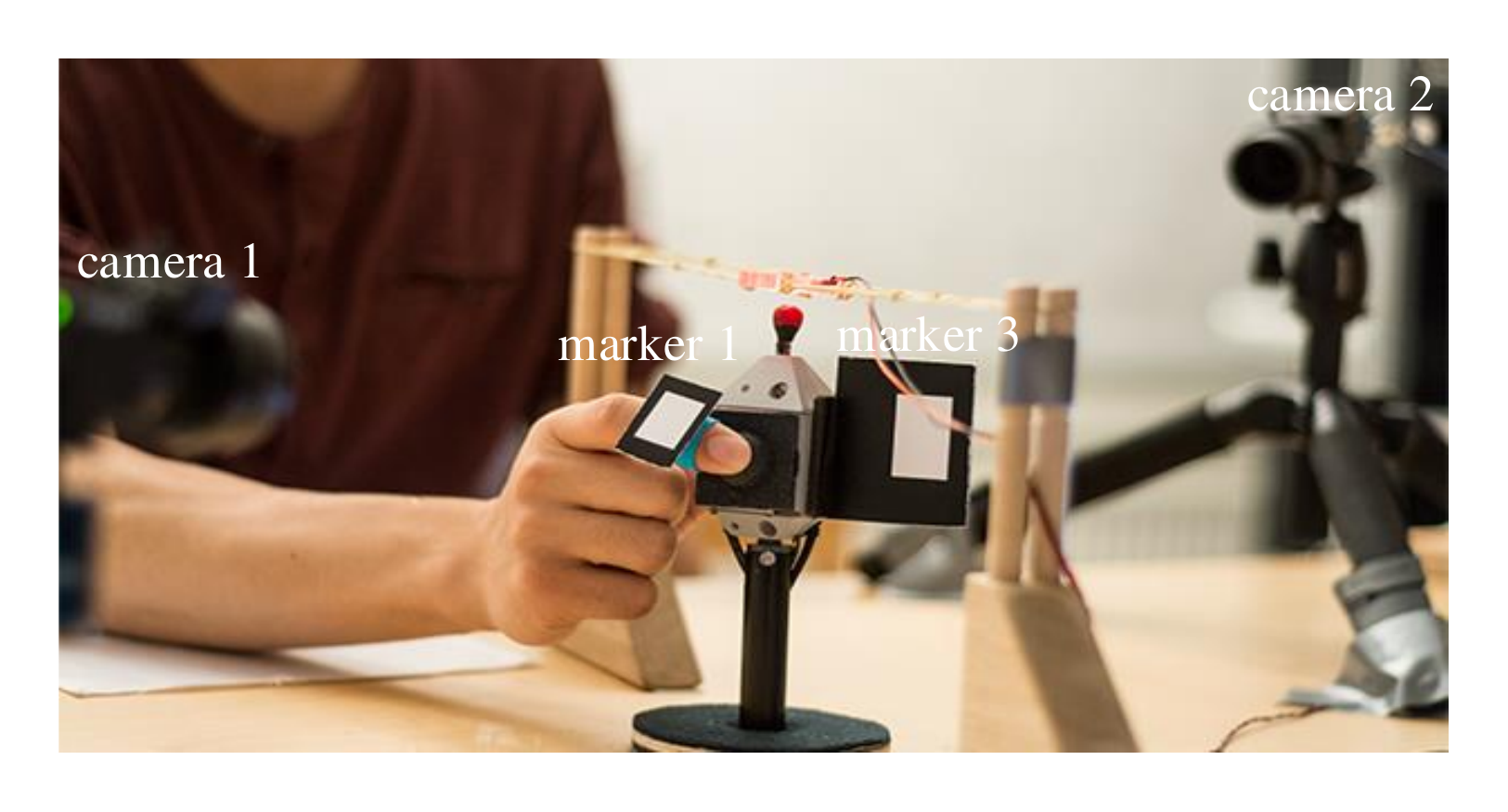}
    \end{minipage}} \hfill
  \subfigure[Recording system (Cont'd).]{
        \label{setup02}
    \begin{minipage}[b]{0.43\textwidth}
      \centering
      \includegraphics[width=0.6\textwidth]{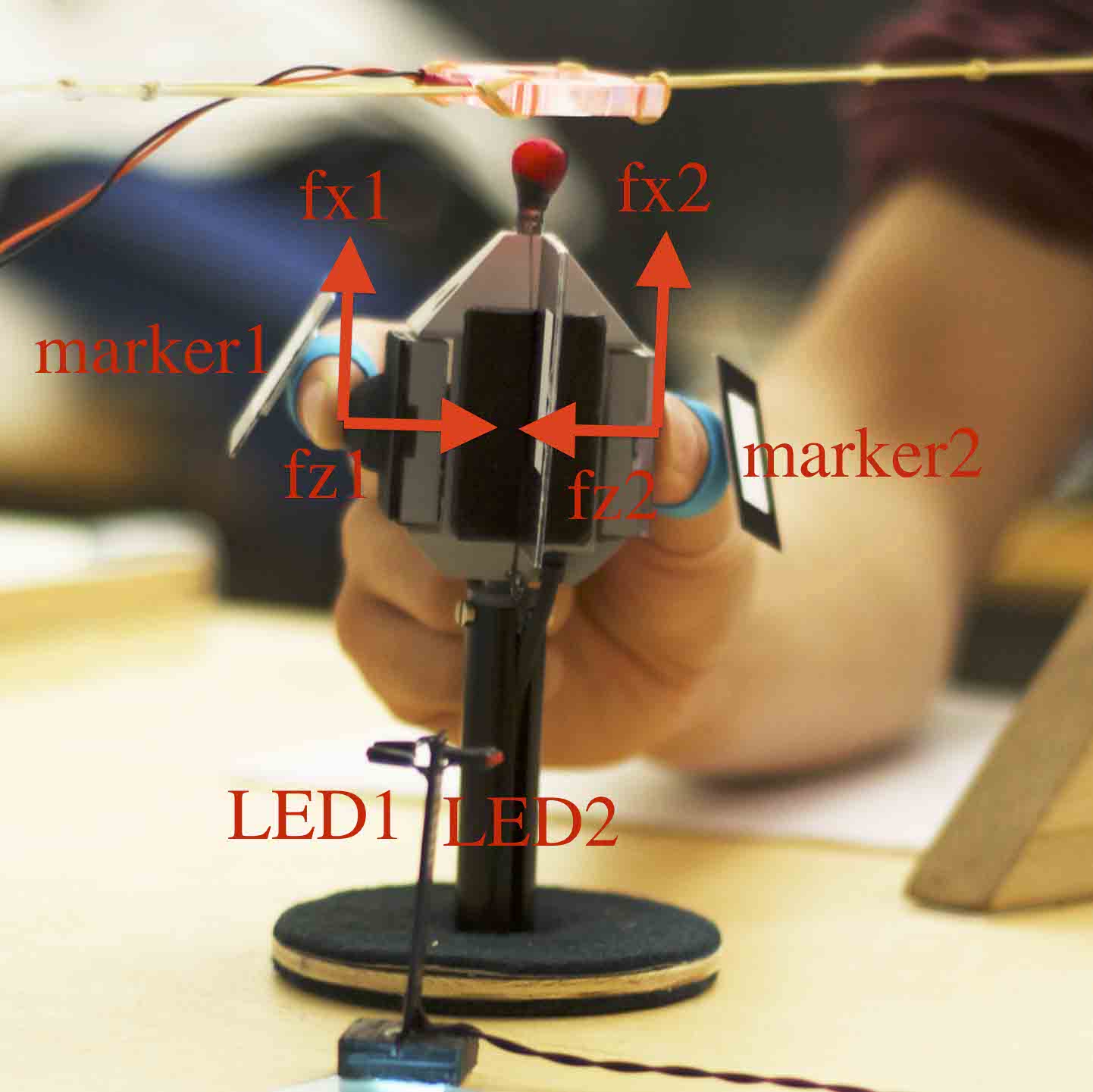}
    \end{minipage}}\\
  \subfigure[The grasped object. The object above the table is visible for the subjects.]{
      \label{setup03}
    \begin{minipage}[b]{0.52\textwidth}
      \centering
      \includegraphics[width=\textwidth]{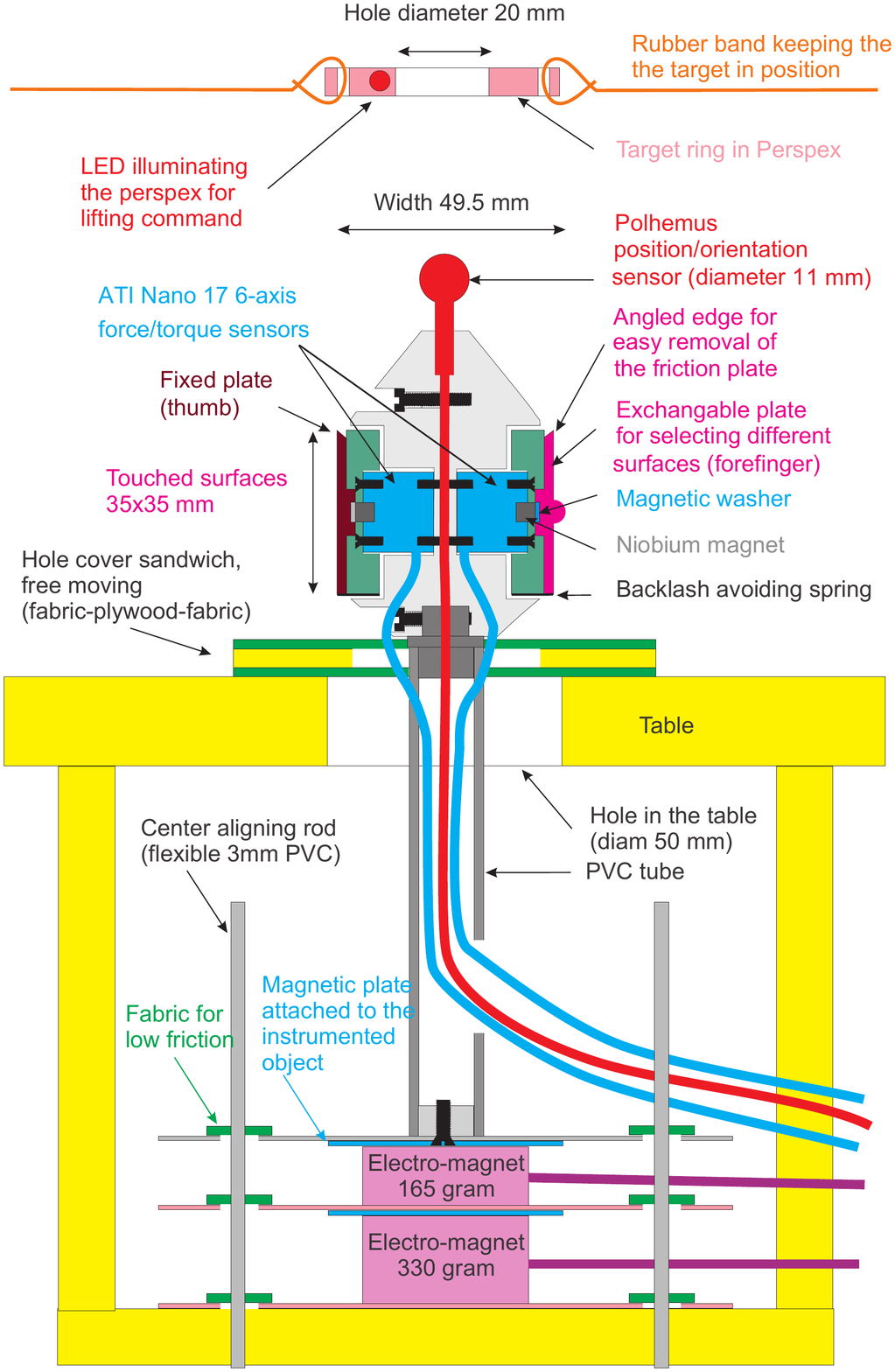}
    \end{minipage}} \hfill
  \subfigure[Contact surfaces for the index finger. Odd rows are the front view, and even rows are the top view. \#\,1 to \#\,11 are sandpaper and \#\,12 is silk.  $r$ represents the radius of the surface.  \#\,5 to \#\,8 are flat in the $x$ or $y$ direction; therefore, $r$ only represents the radius of the non-flat direction.]{
  	\centering
        \label{setup04}
    \begin{minipage}[b]{0.45\textwidth}
      \centering
      \includegraphics[width=\textwidth]{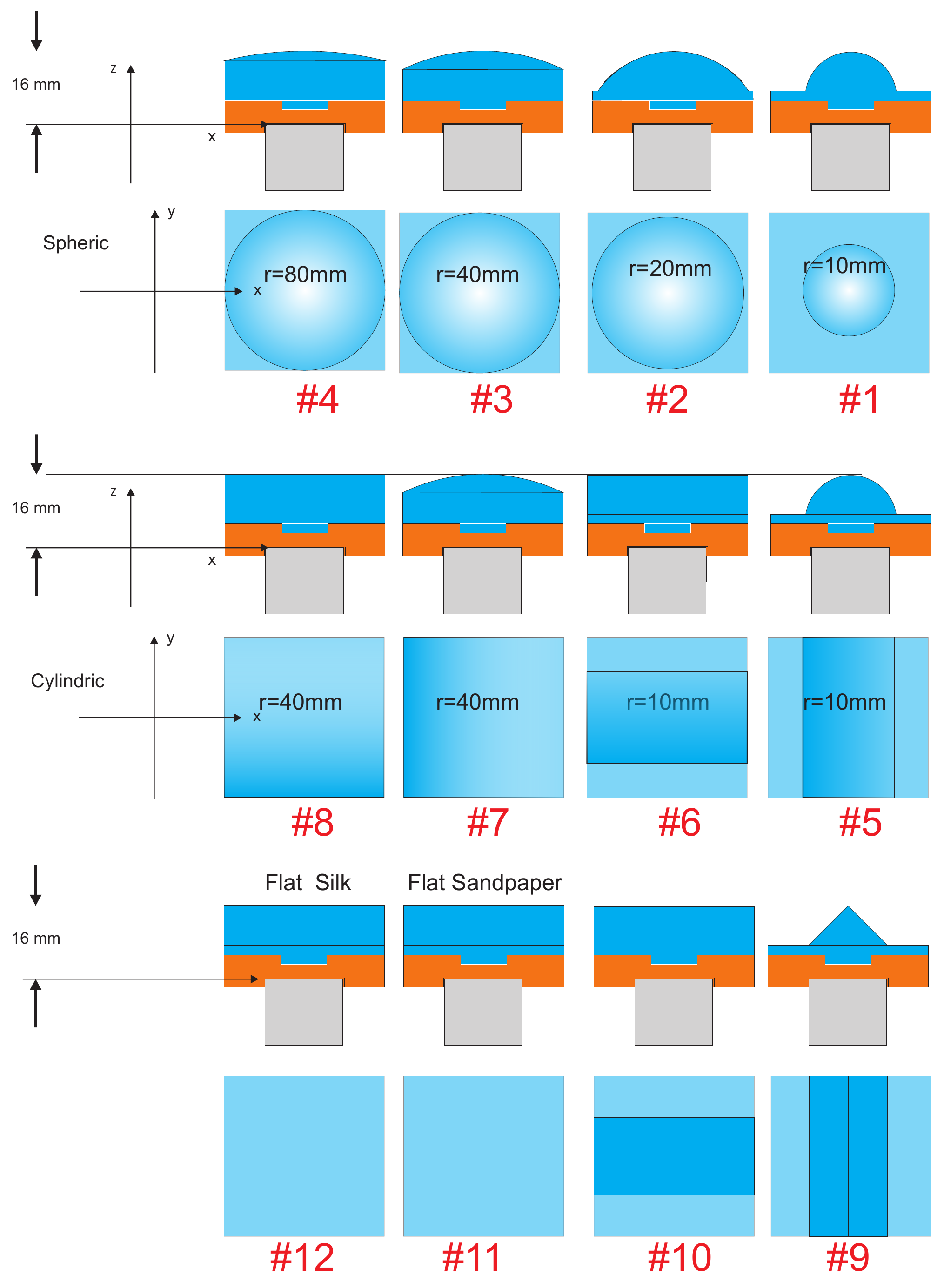}
\end{minipage}}\\
  \caption{Setup. (b) $F$ refers to the forces of the fingertips, where $x$ and $z$ represent lift and grip directions, respectively.}
  \label{setup}
\end{figure*}

\noindent{\textbf{Hardware setup}}.
The
setup consisted of two stationary cameras and two force/torque sensors (see Fig.~\ref{setup01}). The cameras recorded the distal phalanges of the index finger and thumb, illuminated with diffuse light, as well as calibration markers on the respective nails. Video data was captured by two POINTGREY cameras at c:a 24\,fps 
with a resolution of $1280\times980$ pixels.
At the same time, ATI Nano-17 six-axis force/torque sensors
located under the two contact surfaces measured ground truth forces and torques at 100\,Hz.

The whole setup is depicted in Fig.~\ref{setup}. 
Camera\,1 captured the images of the rectangular marker\,1, LED\,1 and the index finger, while camera\,2 captured the images of marker\,2, LED\,2 and the thumb. Marker\,3 was attached to the side of object (Figs.~\ref{setup01},~\ref{setup02}). Niobium magnets guarantee the distance between the two contact surfaces to be 49.5\,mm. 
The object (Fig.~\ref{setup03}) allowed easy adjustment of its weight and contact surfaces. The 12 types of curvatures and surface used in the experiments (Fig.~\ref{setup04}), represent a wide range of commonly grasped objects. The surface on the sensor of the thumb is flat sandpaper.

\noindent{\textbf{Data synchronization}}.
The image data and force/torque data were synchronized using two light-emitting dioides (LEDs) for each camera. For each pair, one LED was always
\emph{on} and the other \emph{off} but they 
switched at random with a mean frequency of 4Hz. The state of the LEDs was stored along the force data. The sequences of LED signals collected from images and along with the forces data were scaled to the same frequency and then synchronized based on the cross-correlation.

\subsection{Image Alignment}

Image alignment was developed to reduce the variance caused by the finger orientation and location in the visual data. Before this, however, the mean shift algorithm \cite{Comaniciu00real-timetracking} was used to track the finger in the video stream. The fingernail and its surrounding skin were then segmented from the background based on the edge, and the fingernail geometry centers were shifted to the same position in the images. To segment the finger image robustly, we transferred the image from RGB to HSV, and changed hue and saturation to distinguish the finger from the background and then transferred the segmented image back to RGB. With this approach it was possible to segment a fingernail from a background color very close to skin color.

Earlier we have demonstrated alignment with convolutional neural networks \cite{chen2014a} but the method we now propose achieves a high quality without reconstruction of a 3D finger model. To this end -- applying the method described below -- alignment transformations were generated using the blue channel of the image since it (as well as the red channel) varied little with contact forces.

Given a reference image $\mathbf{R}$, every subsequent image $\mathbf{J}$ was aligned using non-rigid image alignment \cite{tmi_S10}. In short, assume that the two images have the following intensity relationship:
\begin{align}
\mathbf{R} = \mathbf{J}(T) + \mathbf{v} + z
\end{align}
where $\mathbf{v}$ is an intensity correction field,  $z \sim \mathcal{N}(0, \sigma^2)$ is zero-mean Gaussian noise, and $T$ is the geometric transformation that registers $J$ onto $R$. To estimate $\mathbf{v}$ and $T$, we minimized the objective function
\begin{align}
E(\mathbf{v}, T)= D(\mathbf{v}, T) + w\| \mathbf{P} \mathbf{v} \| ^2
\end{align}
where
\begin{align}
D(\mathbf{v}, T) = \| \mathbf{R} -\mathbf{J}(T)-\mathbf{v} \|^2
\end{align}
is a measure of the similarity of $\mathbf{R}$ and $\mathbf{J}$, and $\| \mathbf{P}\mathbf{v} \|^2$ is a regularization term that penalizes some properties of $\mathbf{v}$, e.g., unsmoothness (the scalar $w$ thus parameterizes the trade-off between the data fitness and regularization).

We modeled the transformation $T$ by using the free-form deformation transformation with three hierarchical levels of B-spline control  
points and  updated the transformation parameters via gradient-descent optimization.

%% file: sections/predictor.tex
\subsection{Predictors}

The fingernail and surrounding skin color distribution and deformation reflected the changes of contact force. Several variants of predictors were developed to construct the mappings from the finger images to the fingertip contact force/torque and contact surface curvatures. Below we describe in detail the four prediction methods: Gaussian Process (GP) regression and Convolutional Neural Networks (CNN), Neural Networks (NN) and Recurrent Neural Networks (RNN).\\

\noindent{\bf{Gaussian Process Regression (GP)}}. GP is a widely used stochastic process \cite{Rasmussen2006}. It is completely defined by its mean $m(\mathbf{x})$ and covariance $k(\mathbf{x},\mathbf{x'})$ functions,
\begin{align}
m(\mathbf{x})&=\mathbf{E}[f(\mathbf{x})],\\
k(\mathbf{x},\mathbf{x}')&=\mathbf{E}\Bigl[\bigl(f(\mathbf{x})-m(\mathbf{x})\bigr)\bigl(f(\mathbf{x}')-m(\mathbf{x}')\bigr)\Bigr].
\end{align}
A basic assumption of GPs is that the target values are similar for similar inputs. A smooth function that measures the closeness or similarity of inputs is the squared exponential (SE) covariance:
\begin{equation}
k(\mathbf{x},\mathbf{x}') = \sigma_f^2 \exp\Bigl( -\frac{1}{2l^2} \| \mathbf{x} - \mathbf{x}' \|_2 \Bigr),
\end{equation}
where the signal variance $\sigma_f^2$ and length-scale $l$ are hyperparameters. 

In our implementation, the inputs $\mathbf{X} :=(\mathbf{x}_1,\mathbf{x}_2, \dots, \mathbf{x}_N )^T$ were aligned images reshaped to 1D vectors. The associated targets $\mathbf{y} := (\mathbf{y}_1, \mathbf{y}_2, \dots, \mathbf{y}_N)^T$ were the measured forces and torques and the curvature of the surface in contact with the fingertip. Based on a training set $\{(\mathbf{x}_i, \mathbf{y}_i),i = 1, 2, \dots , N\}$, we obtained the predictive mean and variance of the function values $f^*$ at testing locations of the data points $\mathbf{x}^*$ by using
\begin{align}
\mathbf{E}[f^*] &= \mathbf{k}^{*T} (K + \sigma_n^2 I)^{-1} \mathbf{y},\\
\mathrm{Var}[f^*] &= k(\mathbf{x}^*, \mathbf{x}^*) - \mathbf{k}^{*T} (K + \sigma_n^2 I)^{-1} \mathbf{k}^*,
\end{align}
where $K_{ij} = k(\mathbf{x_i}, \mathbf{x_j})$, $k^*_i = k(\mathbf{x_i}, \mathbf{x^*})$, 
$\sigma_n^2$ is the noise variance, and $I$ represents identity matrix.

The optimal values for the hyperparameters $\{\sigma^2_f, l, \sigma^2_n\}$ were evaluated from the training set by maximizing the log likelihood function
\begin{align}
\log p(\mathbf{y}|X)=-\frac{1}{2} \mathbf{y}^T (K+\sigma_n^2I)^{-1}\mathbf{y} 
	- \frac{1}{2} \log \left| K +\sigma_n^2 I \right| - \frac{N}{2} \log 2 \pi.
\end{align}

The GP predictor was able to train one model for all participants but then the data set increased, of course, to tens of thousands of images. A critical issue with GP methods is that large computations are required: $O(N^3)$ for training and $O(N^2)$ for per testing case, where $N$ is the number of training samples \cite{Rasmussen2006}. To reduce the computational costs, we implemented the \textit{fully independent training conditional approximation} (FITC) (a.k.a. \textit{sparse pseudo-input GP}\cite{NIPS2005_2857}). The \textit{inducing points} are a small amount of inputs $M$ that summarize a large number of inputs $N$. By using inducing points we reduced the training and testing cost to $O(NM^2)$ and $O(M^2)$, respectively. FITC was implemented by randomly selecting a subset of the training data as the inducing points: $\overline{\mathbf{X}} := ( \overline{\mathbf{x}}_1, \overline{\mathbf{x}}_2,\dots, \overline{\mathbf{x}}_m )^T$. A more efficient likelihood approximation is then given by
\begin{align}
p(\mathbf{y} | \mathbf{f})  \simeq q(\mathbf{y} | \mathbf{u}) 
  = \mathcal{N} (K_{f,u}K^{-1}_{u,u}u, \textmd{diag}[K_{f,f}-Q_{f,f}+\sigma^2_\mathrm{noise}I]),
\end{align}
where $\mathbf{u}$ is the corresponding latent values of $\overline{\mathbf{X}}$, $\mathbf{f} = \{f_n\}^N_{n=1}$ are latent values based on $\mathbf{x}_n \in \mathbf{X}$, the covariance function $K_{f,f}$ is the Gram matrix of all pairs $(\mathbf{x}_i, \mathbf{x}_j)$, and diag[*] is a diagonal matrix, and $Q_{f,f} \doteq K_{f,u}$.\\

\noindent{\bf{Convolutional Neural Networks (CNN).}}
\label{sec:cnn}
CNN \cite{lecun1998} is another approach for regression and classification that potentially suitable for predicting forces from our video streams. It is relatively robust to shifts, scales and distortions of the input data and, importantly, can be efficiently trained on large data sets.

   \begin{figure}
      \centering
      \includegraphics[width=9.5cm]{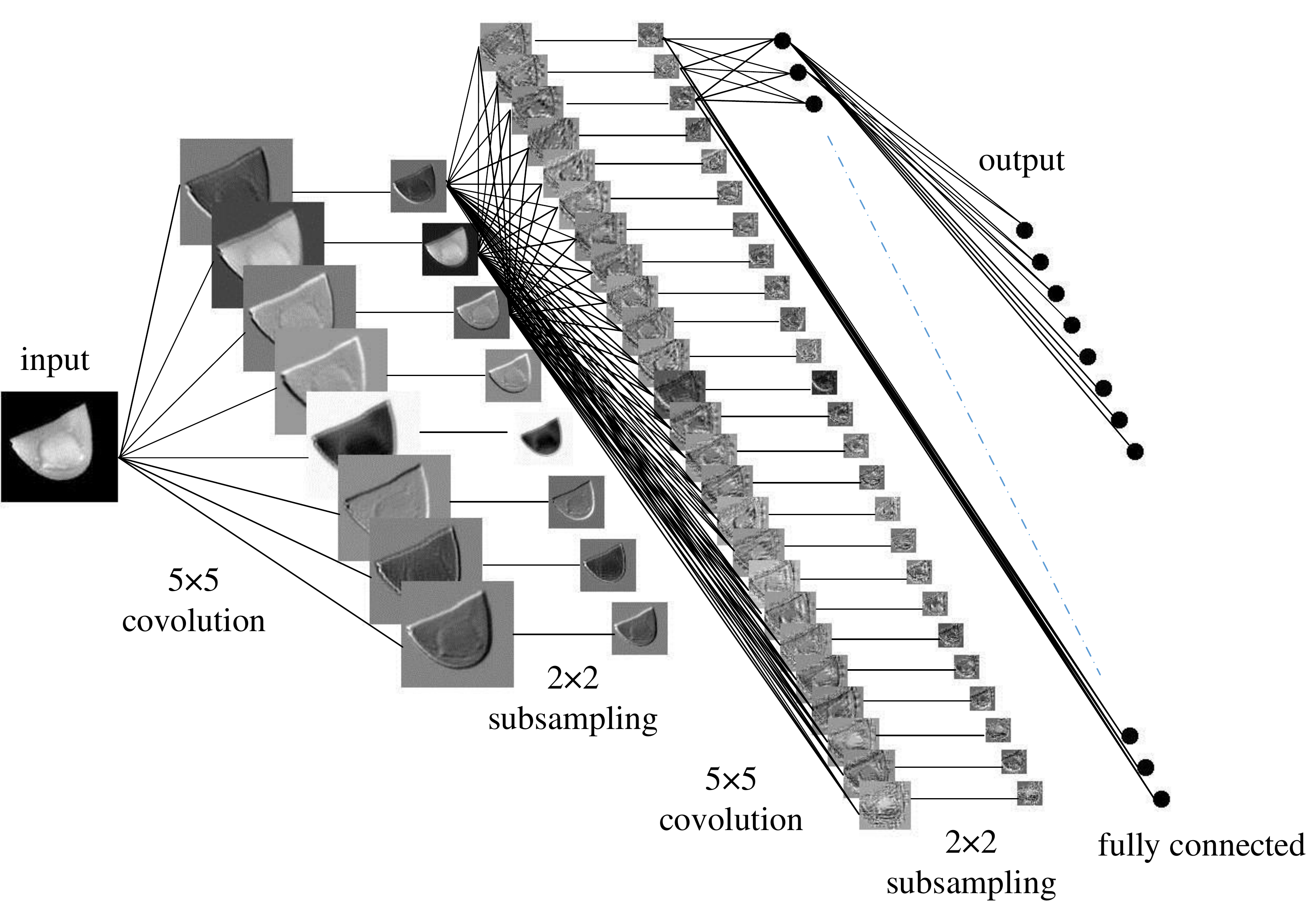}
      \caption{Architecture of convolutional neural networks for fingernail images.}
      \label{cnn}
   \end{figure}

Fig.~\ref{cnn} illustrates the architecture of the proposed network for finger images.
It contains six layers: a first convolutional layer followed by a first max-pooling layer, another convolutional layer followed by a second max-pooling layer, and two fully connected layers.
In our experiments, both convolutional layers had 5$\times$5 sized filters.
The first convolutional layer employed 8 kernels, the second layer used 25.

We now describe convolutional neural networks more formally.
Typically, a CNN is designed as subsequent stages of convolution and max-pooling.
The top layers are usually ordinary multi-layer perceptrons.

The convolution of a 2D image for a feature map $h$ is
\begin{equation}
  \label{lr}
   h(m,n)=\sum_{u=0}^{l_x}\sum_{v=0}^{l_y}w(u,v)\,g(u+m,v+n) + b,
\end{equation}
where $g$ is the input map, $w$ the kernel weights, and $b$ the bias. $(l_x,l_y)$ is the size of the filter. $(m, n)$ is the pixel position on the feature map.

The max-pooling activation can be computed as
\begin{equation}
  \label{}
   p(m,n)=\textstyle{\max_{i=1}^{r_1}}\Bigl(
   			 \textstyle{\max_{j=1}^{r_2}}\,h(r_1m+i,r_2n+j)\Bigr),\end{equation}
where $(r_1,r_2)$ is the pooling size and $p$ is the feature map in the max-pooling layer.
Max-pooling, a non-linear down-sampling method, decreases the computational complexity.
These layers take the output of convolutional layers as input, and reduce the resolution of the input.

The fully-connected MLP contains 100 hidden units.
The last layer is linear and has 8 outputs $\hat{\mathbf{y}}:=\{f_x, f_y, f_z, \tau_x, \tau_y, \tau_z, c_1, c_2\}^T$. In our model, the activation of MLP layer was tanh, and the activation of the output layer was identity. The outputs of the MLP were the input of linear regression. 

We use the chain rule to backpropagate error gradients back into the network to minimize the square error loss:
\begin{align}
L=\frac{1}{N} \sum^N_{i=1} \| \mathbf{y}-\hat{\mathbf{y}} \| _2,
\end{align}
where $\mathbf{y}$ is the ground truth of the outputs.

\noindent{\bf{Fast Dropout Neural Networks.}}
\label{sec:NNFD}
\subsubsection{Neural Networks (NN)}
Given training image vector $\mathbf{x}$, we have a neural network:
\begin{equation}
    \mathbf{f}(\mathbf{x}) = \sigma (\mathbf{W}\mathbf{x}+\mathbf{b})
\end{equation}
where $\{\mathbf{W}, \mathbf{b} \}$ are the parameters, and $\sigma$ is the activation function. It can be extended to multi-layer neural networks with more hidden layers by taking the output of one layer as the input of another layer. The parameters of the network are obtained by minimizing the loss function
\begin{align}
L=\frac{1}{N} \sum^N_{i=1} \| \mathbf{y}-\mathbf{f}(\mathbf{x}) \| ^2_2.
\end{align}

\subsubsection{Fast Dropout (FD)}

Random dropout \cite{corr_dropout} of the neurons during training process in the last layer of neural networks prevents complex co-adaptations and reduces over-fitting. 
Compared with random dropout, fast dropout \cite{wang13a} is more efficient to train a model by sampling from a Gaussian approximation.

$z_i \sim \mathrm{Bernoulli}(p_i)$ is sampled to determine whether the input $x_i$ is dropped out, where $p_i$ is the rate of not dropping out. The output $y$ is derived by
\begin{align}
a&=\mathbf{w}^T \mathbf{D}_z \mathbf{x} \\
y&=\sigma(a)
\end{align}
where $\mathbf{w}$ is a weight vector and $D_z = \mathrm{diag}(z) \in \mathbb{R}^{m\times m}$. $m$ is the data dimension.

The input of the output layer takes a random variable for every hidden unit. Under fast dropout training, we can assume its input as a Gaussian distribution $\mathbf{X} \sim \mathcal{N}(x| \mu, s^2)$. For any hidden unit, the mean and variance of output $y$ are $\nu$ and $\tau^2$. Using sigmoid activation function $\sigma$, we have:
\begin{align}
    \nu &= \int^{\infty}_{-\infty} \sigma(x)\mathcal{N}(x | \mu, s^2) dx \approx \sigma(\frac{\mu}{\sqrt{1+ \pi s^2/8}}), \\
    \tau^2  &= \underset{X\sim \mathcal{N}(\mu, s^2)}{\mathrm{Var}} [\sigma (X)] = E[\sigma (X)^2]-E[\sigma (X)]^2.
\end{align}

We draw samples of a Gaussian approximation for $a=\mathbf{w}^T \mathbf{D}_z \mathbf{x}$. The mean and variance of $a$ can be obtained. We assume that $\mathbf{x}$ components to be independent; therefore, the central limit theorem of Lyapunov condition is satisfied with $m\rightarrow \infty$. Consequently, $a$ is approximately Gaussian.

The neural networks with fast dropout can be trained to update $\mathbf{w}$ through back-propagation.

\noindent{\bf{Recurrent Neural Networks (RNN) with Fast Dropout}.}
Since picking up and placing is a sequence movement, recurrent neural network is employed to process the sequence data.

We have a sequence of finger images $ \mathbf{x}_t \in \mathbb{R}^l (t=1,2,\dots T)$ and corresponding forces, torques and surface curvatures $\mathbf{y}_t \in \mathbb{R}^m$ $(t= 1,2,\dots T)$. $\hat{\mathbf{y}}_t  \in \mathbb{R}^m$ $(t= 1,2,\dots T)$ is the output of RNN which has hidden layers $\mathbf{h}_t  \in \mathbb{R}^n$ $(t=0,1,\dots T)$. $f_h$ and $f_y$ are transfer functions, the constant $T$ is the sequence length and $l$, $m$, $n$ are the input, output and hidden dimensions at every time step.
With one hidden layer, we have
\begin{align}
\mathbf{h}_t & = f_h(\mathbf{x}_t\mathbf{W}_\mathrm{in}+\mathbf{h}_{t-1}\mathbf{W}_\mathrm{rec}+\mathbf{b}_h), \nonumber \\
\hat{\mathbf{y}}_t & = f_y(\mathbf{h}_t\mathbf{W}_\mathrm{out}+\mathbf{b}_y)
\end{align}
where $\theta=\{\mathbf{W}_\mathrm{in}, \mathbf{W}_\mathrm{out}, \mathbf{W}_\mathrm{rec}, \mathbf{b}_h, \mathbf{b}_y\}$ are the parameters.
The gradients are calculated by back-propagation through time. The parameters are obtained by minimizing the loss function
\begin{align}
\mathcal{L}(\theta)=\sum_i \| \hat{\mathbf{y}}^{(i)} - \mathbf{y}^{(i)} \|_2.
\end{align}

RNN with fast dropout (RNN-FD) \cite{bayer2014a} can be straightforwardly implemented as FD applied to NN, both described above.

\subsection{Calibration and postprocessing}

The output of the F/T sensors consisted of the forces and torques applied on each sensor, but these were not exactly equal to the fingertip forces and torques because of the shift of the contact position and the rotation of the finger with respect to the sensor.
Therefore, we calibrated separately the forces and torque to correct the fingertip forces and torques.
After training a force/torque predictor by the calibrated training and validation data, calibration of the testing data was only for the purpose of checking the accuracy of the results.  When the system is used ``in production'', the calibration process is not required.
Our approach to compute fingertip force and torque does not depend on knowing the axis of the fingertip nor on knowledge of the spatial orientation of the manipulated object.

\noindent{\textbf{Torque calibration}}. We summarize forces on a contact surface into a three-dimensional force vector and a three-dimensional torque vector at a reference point. The first contact point of a trial is set to the reference point. To compute the torques at the reference point, the location of the point with respect to the sensor coordinate is required.

The object is static with respect to the contact point of the finger. Therefore, the sum of the torques at the reference point is zero:
\begin{align}
\Delta\tau_x = f_zy-f_yz,\
\Delta\tau_y = -f_zx +f_xz,\
\Delta\tau_z = -f_xy +f_yx,
\label{eq_f2t}
\end{align}
where $\Delta\tau = \tau - \tau'$, and $\{x, y, z\}$ is the contact position in the sensor coordinate. $\tau'$ is the finger torque. 
We calibrate the the torque using the data when the subject starting touching the object. In this case, the contact surface approximates to a point and the friction is not enough to generate a torque at the reference point. Thus, $\tau'$ approximately equals to zero ($\Delta\tau \approx \tau$). 

Given the force and torque, there are infinitely many possibilities for the contact positions computed from (\ref{eq_f2t}). In the training dataset, the contact surface is known, so that we have $z = h(x, y)$, where $h$ is a function of the geometry of the surface. With $h$ and (\ref{eq_f2t}), we can compute the contact position $\{x, y, z\}$. In one trial of the experiment, the contact point does not change. Consequently, after obtaining the contact position, we can calibrate the torques for the entire trial. Additionally, the distribution of the nail blood reflects the contact points, so that the prediction process is able to map the fingernail image to the torque.

\noindent{\textbf{Force calibration}}.
As the fingertips were rotated during grasping, the force vectors on the fingertips were also rotated with respect to the sensor coordinate.
The mapping from the finger image to the force is a surjection for the rotation with respect to the $x$ and $y$ axes, so that we only calibrate $f_x$ and $f_y$, which are rotated with respect to the $z$ axis.
We focused on force rotation calibration using two printed markers (cf., ig.~\ref{setup01}) for each finger.

First of all, a rectangular marker (''Marker 3'' in Fig.~\ref{setup01}) was attached to the object such that its position and orientation could be detected by a 2D camera using HALCON (MVTec). The four borders of the marker
were detected, and the corresponding intersections were taken as corners of the rectangle.
Once the internal camera parameters, the rectangle's physical size, and the detected corners are known, the rectangle's pose in the camera space could be initially estimated.
After that, a nonlinear optimization approach updated the pose through minimization of the 
geometrical distance between the detected borders and the back projection of the space rectangle onto the image. 

The orientation $\theta$ of the fingertips with respect to the object was calculated as the difference in position between "Marker 3" and the marker fixated to the finger (e.g., "Marker 1" in Fig.~\ref{setup01}). 

We selected a 
video
frame as the reference frame and estimate the angle $\theta_r$. Thus, $f_x$ and $f_y$ were calibrated by 
a $(\theta - \theta_r)$ rotation 
with respect to the $z$ axis.

\noindent{\textbf{Marker Calibration}}.
The marker location on the fingers could change between recordings since they were pasted without calibration. This required some pre-processing for the marker orientation calibration. To this end, we
recorded a small data set and rotated the ground truth of $f_x$ and $f_y$ with respect to the $z$ axis in $[-20, 20]$ degrees in one-degree steps. 
The angle that resulted in the minimal mismatch was subsequently used to match the finger marker orientation to the orientation in the training data set.

\noindent{\textbf{Postprocessing}}.
To reduce the prediction noise, we smoothed the outputs from the predictors. Weighted linear least squares and a 2nd-degree polynomial model were used for local regression. In addition, it assigned lower weight to outliers in the regression to smooth the data robustly.

%% file: sections/results.tex
\section{Experiments and results}
\label{sec:2}

\subsection{Data}
Kinematic and kinetic data were acquired with an experimental setup used previously \cite{Luciw:2014}. In short, five healthy, right-handed participants (age 19--65; one female) 
were asked to repeatedly grasp and lift an instrumented object using their thumb and index finger. A light-emitting diode (LED) mounted into a translucent Perspex rectangle above the object signaled the start of a trial and when the object should be replaced on the support table (Fig.~\ref{setup}). 
The light intensity at the nails was 120\,lux at the index finger and 150\,lux at the thumb.

The weight of the object varied between 165, 330 and 660\,g. The contact surfaces at the two digits could easily be changed between any of 12 surfaces all but one covered with sandpaper
(Fig.~\ref{setup04}): 
4 spherical convex surfaces ($c$=12.5, 25, 50 or 100 m$^{-1}$), 4 cylindrical convex surfaces ($c_1$ and $c_2$=0, 25 or 100), 2 surfaces with triangular surfaces, and 2 flat surfaces (one with sandpaper and one with silk). 
There were thus 3 weights~$\times$~12 surfaces~=~36 weight-surface combinations.
Each participant repeated every combination 5 times in an unpredictable order, i.e., each participant completed a total of 180 grasp-and-lift trials.
Each of the two contact surfaces 
-- one for the thumb and one for the index finger --
were coupled to a six-axis force/torque sensor (ATI F/T 17). 
The series of trials were divided into segments by detecting the initial object contact and the moment of object release.

Our experiments required the participants to apply up to $\approx{15}\,$N surface normal force to pick up the object (i.e., with silk surface), and this represents an extension of \cite{2013Sebastian, Grieve2010}, which restricted the normal force to $\le{10}$\,N and the shear force to $\pm{2.5}$\,N.

\begin{figure}
  \centering
  \subfigure[Image alignment with mesh transformation illustration \cite{chen2014a}.The columns from left to right are the reference image, the images before alignment, the aligned images and the mesh transformation. The finger images are in the green channel. The more deformation the image (row 2, column 2) has before alignment, the more mesh transformation (row 2, column 4) it is.]{
      \begin{minipage}[b]{\columnwidth}
       \centering
	\includegraphics[width=0.5\columnwidth]{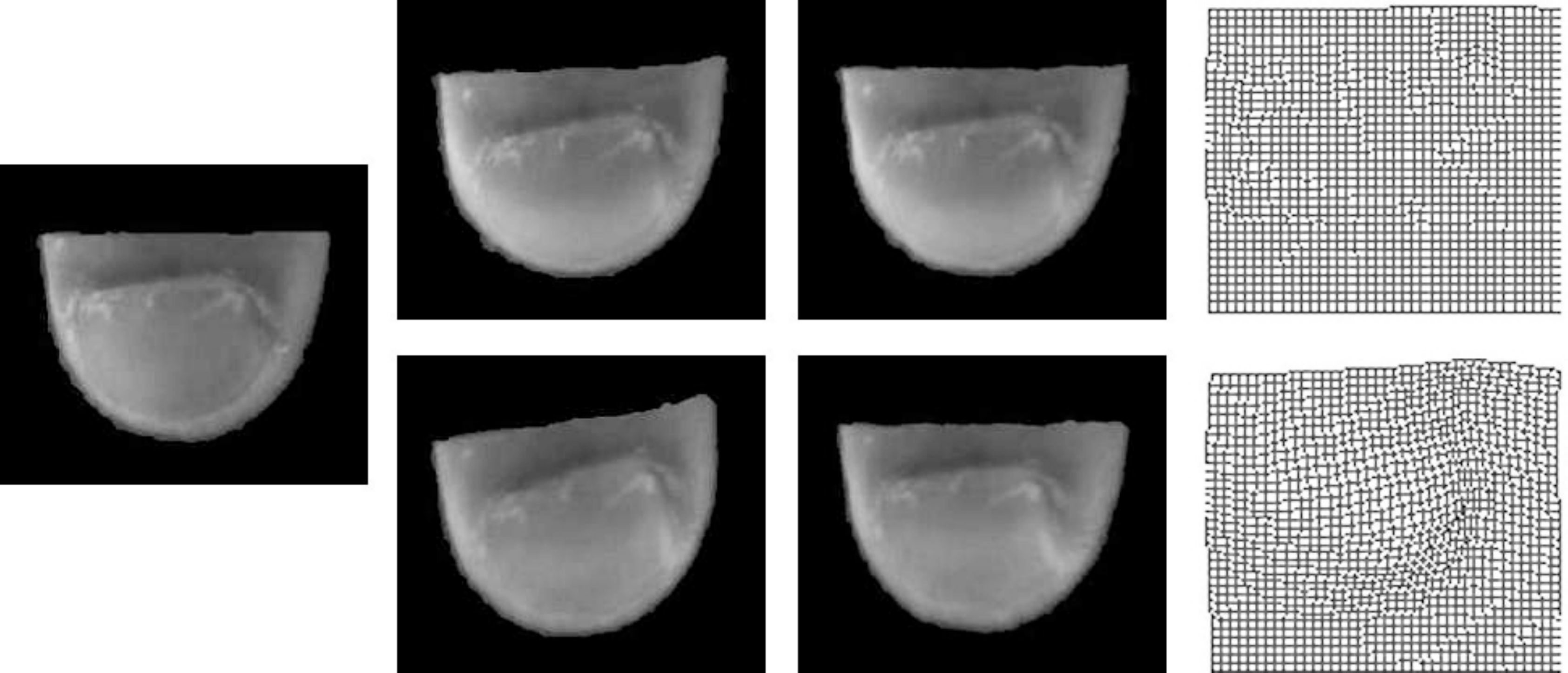}
        \end{minipage}}\\
  \subfigure[Image alignment of three subjects. The left columns of each subfigure are the reference images. The upper and lower lines are images before alignment and aligned images, respectively.]{
    \begin{minipage}[b]{\columnwidth}
     \centering
      \includegraphics[width=0.32\columnwidth]{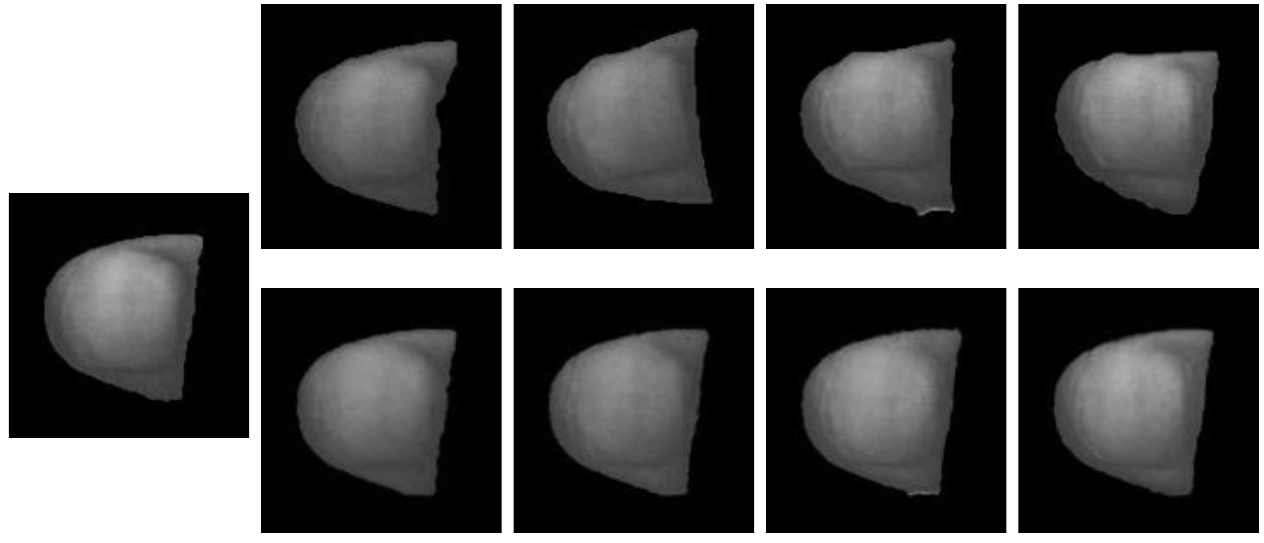}
      \includegraphics[width=0.32\columnwidth]{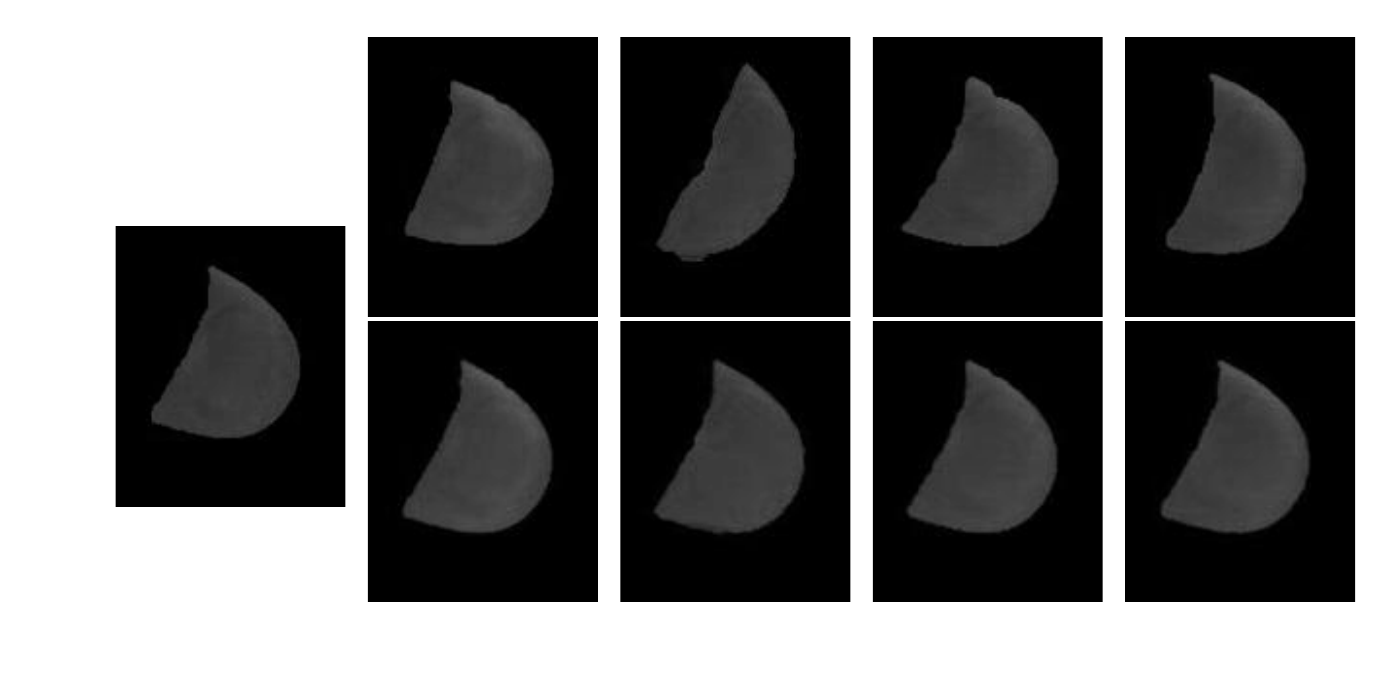}
      \includegraphics[width=0.32\columnwidth]{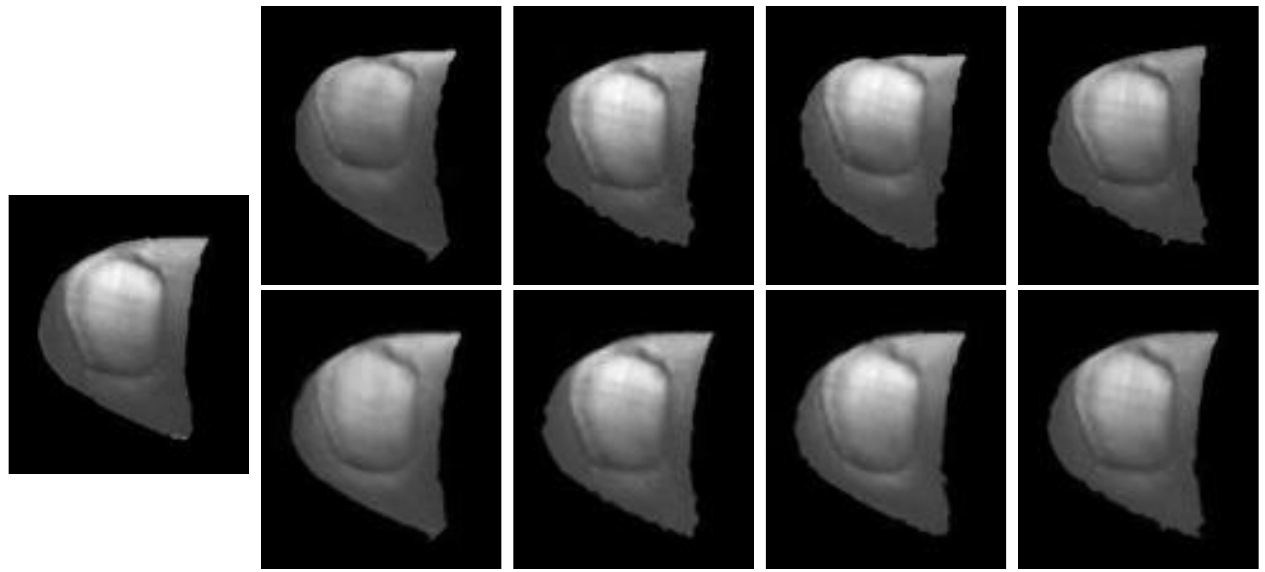}
    \end{minipage}}\\
    \caption{Image alignment.}
     \label{image_alignment_result}
\end{figure}

The testing samples were selected from time-continuous 
data, which increases the prediction difficulty compared to when the testing data is randomly selected from the whole data set as in \cite{tomas2013}.

We used root-mean-square error (RMSE) to evaluate the results. The unit of force and torque are N and Nmm, respectively, 
unless otherwise stated.

\subsection{Image alignment}

We aligned the images for each subject and each finger separately.
Fig.~\ref{image_alignment_result} shows examples of image alignment for visualisation.

\subsection{Force/torque prediction}
\label{result1}

\begin{figure}
\centering
  \subfigure[The output error.]{
    \begin{minipage}[b]{0.45\columnwidth}
      \centering
      \includegraphics[width=\columnwidth]{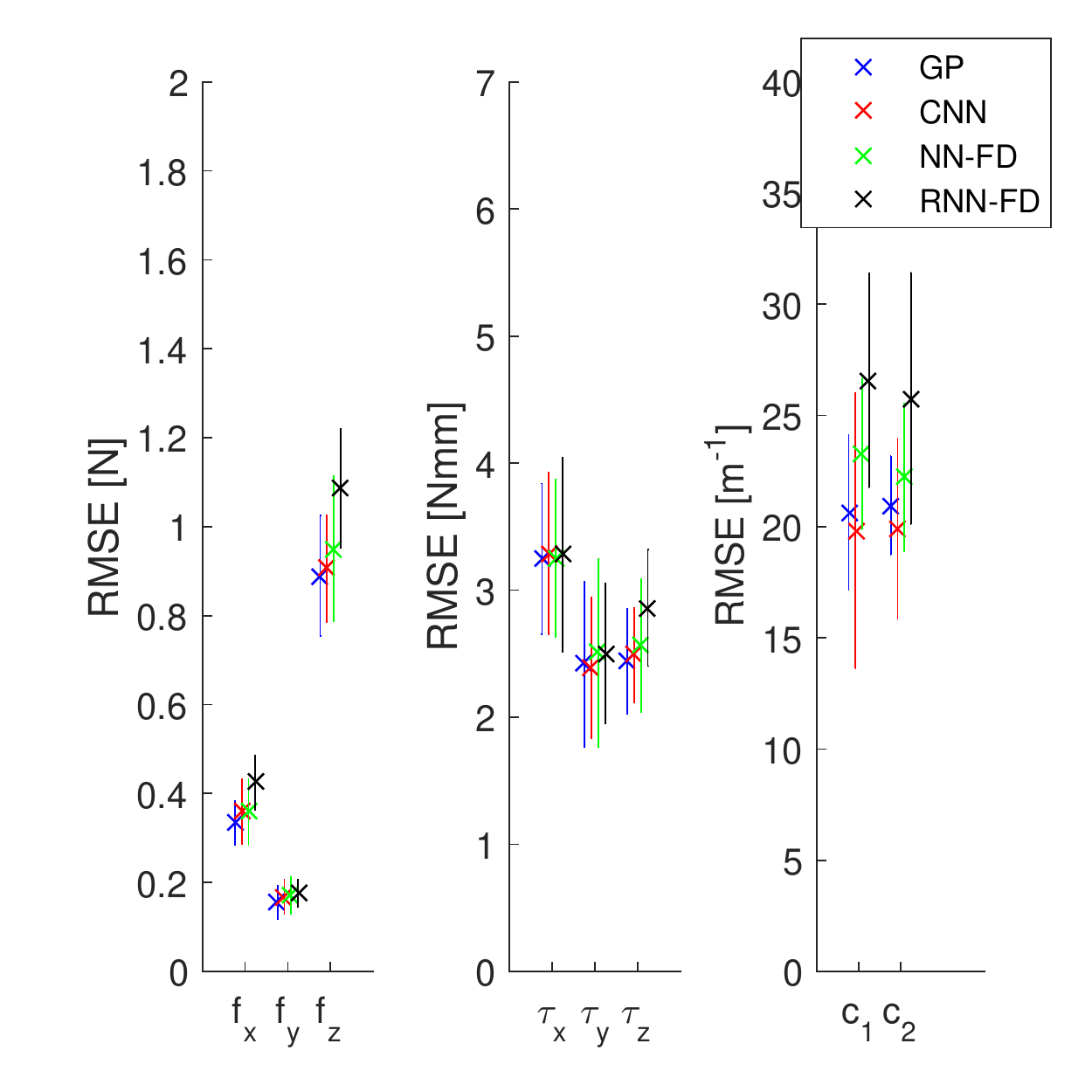}
    \end{minipage}}
  \subfigure[The output standard deviation.]{
    \begin{minipage}[b]{0.45\columnwidth}
      \centering
      \label{method_compar_std}
      \includegraphics[width=\columnwidth]{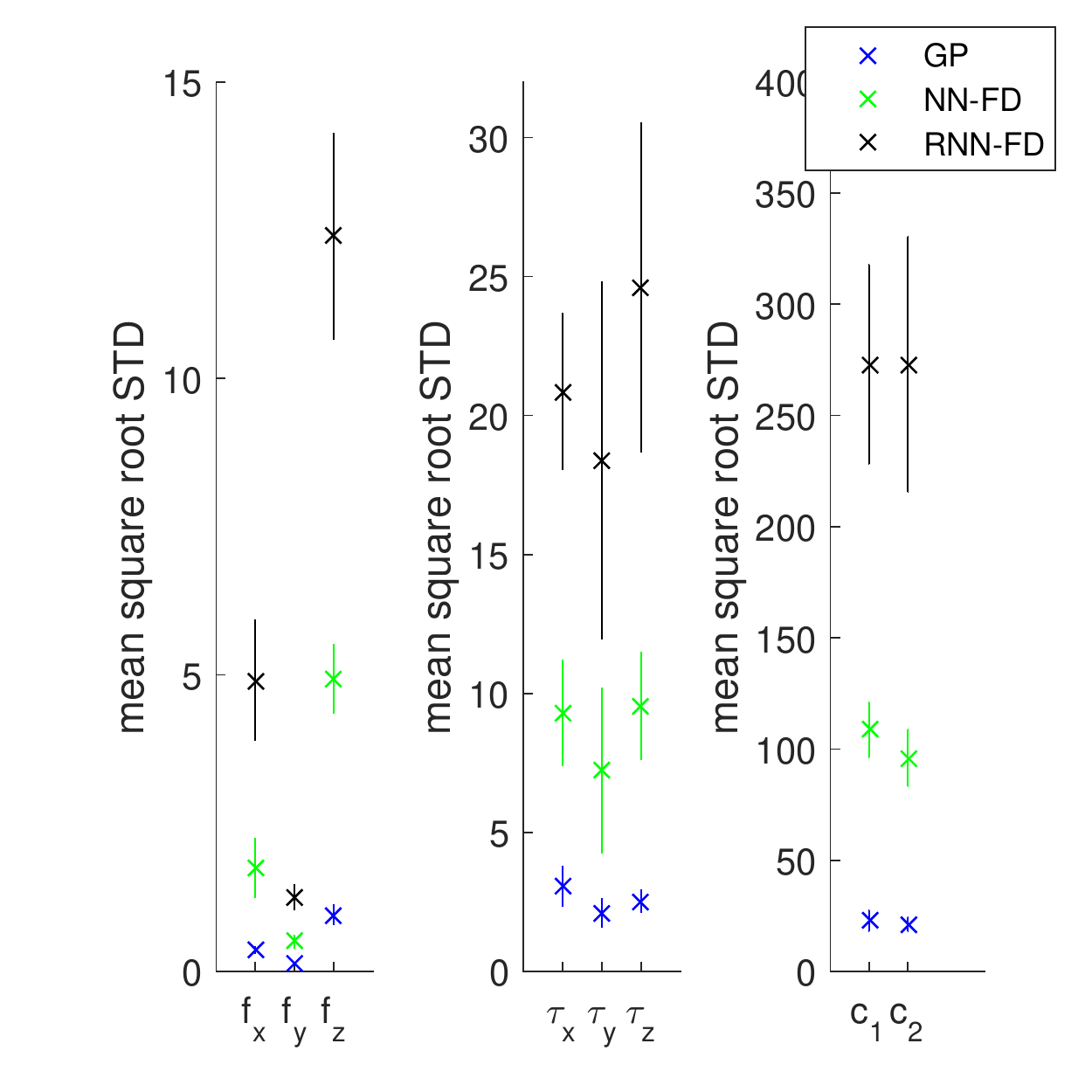}
    \end{minipage}}\\
    \vspace{0 in}
  \caption{Predictor comparison for index fingers of all 5 participants. One of five series was selected as testing data set for each kind of surface and object weight. Each subject has one predictor model. (Since there is no standard deviation of the CNN output, it is not included in the lower panel.)}
  \label{method_compar}
\end{figure}

\begin{figure}
	\centering
	\subfigure[RMS error in force components $f_x$, $f_y$, $f_z$ across all 5 participants. The percentiles are computed separately in $f_x$, $f_y$ and $f_z$.]{
		\begin{minipage}[b]{0.47\columnwidth}
			\centering
			\includegraphics[width=\columnwidth]{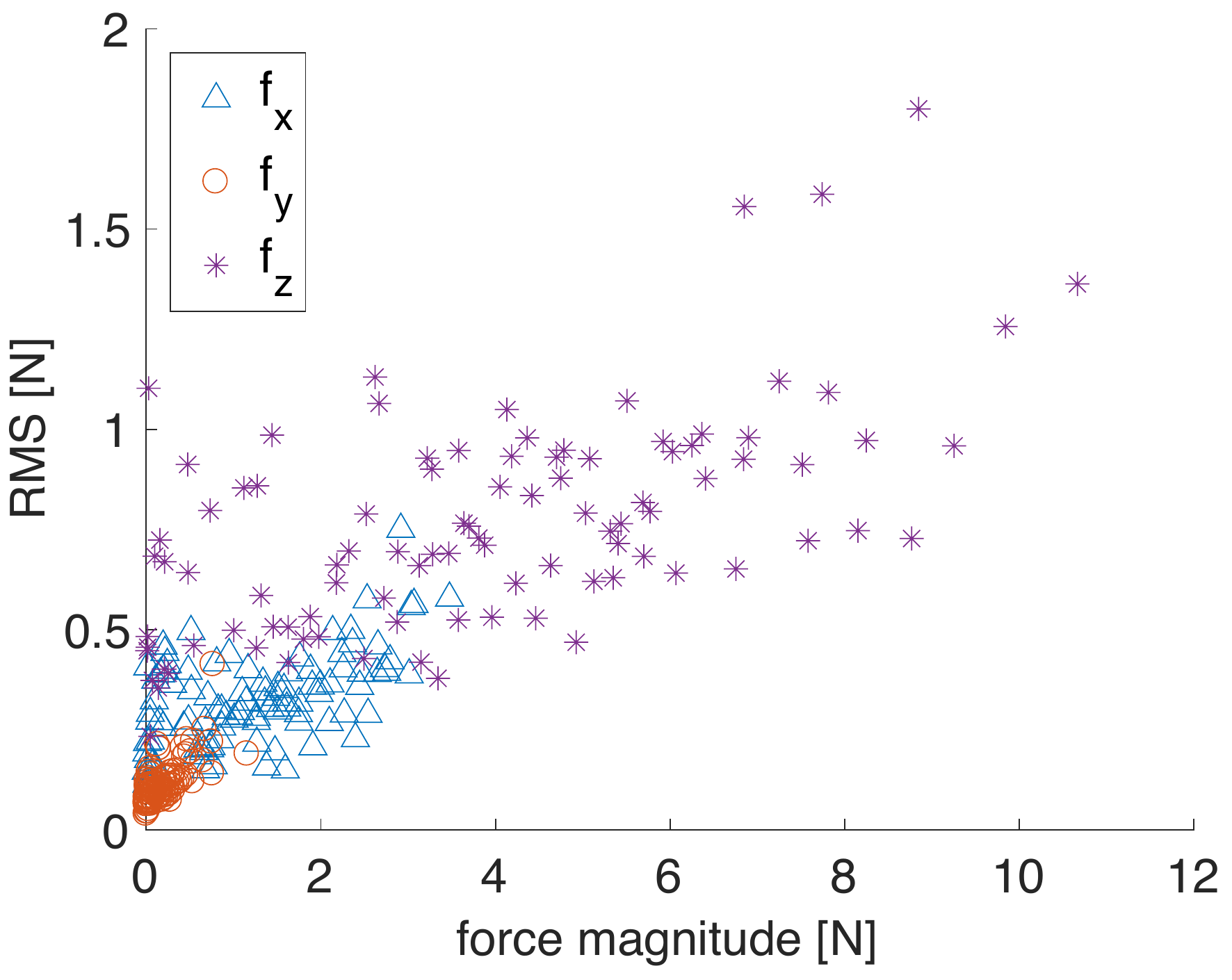}
	\end{minipage}}\hfill
	\subfigure[RMS error in predicted force length magnitude.]{
		\begin{minipage}[b]{0.47\columnwidth}
			\centering
			\includegraphics[width=\columnwidth]{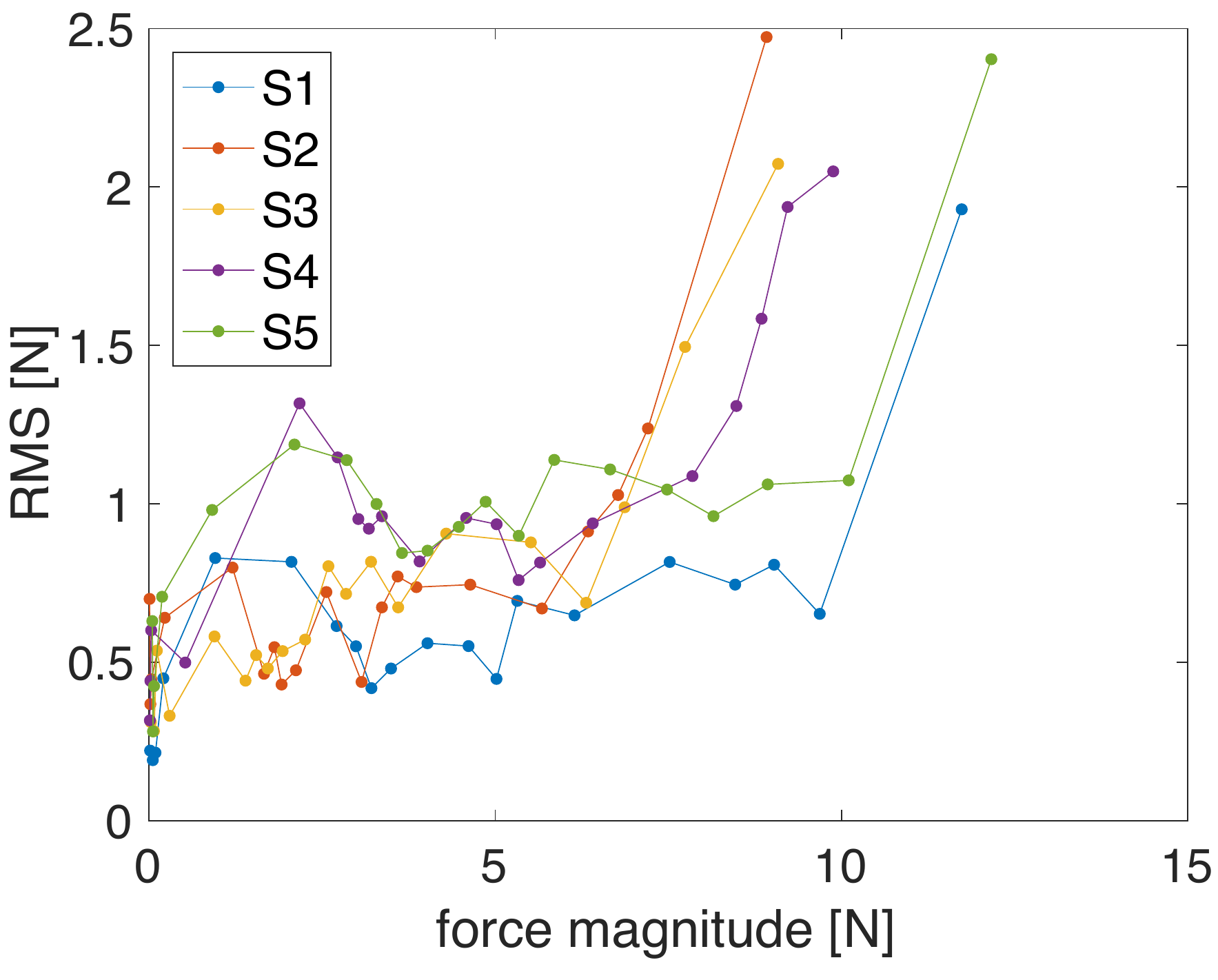}
	\end{minipage}}\\
	\subfigure[RMS error in angle between predicted and surface normal ($f_z$). With increasing the force, the error decreases.]{
		\begin{minipage}[b]{0.47\columnwidth}    
			\centering
			\includegraphics[width=\columnwidth]{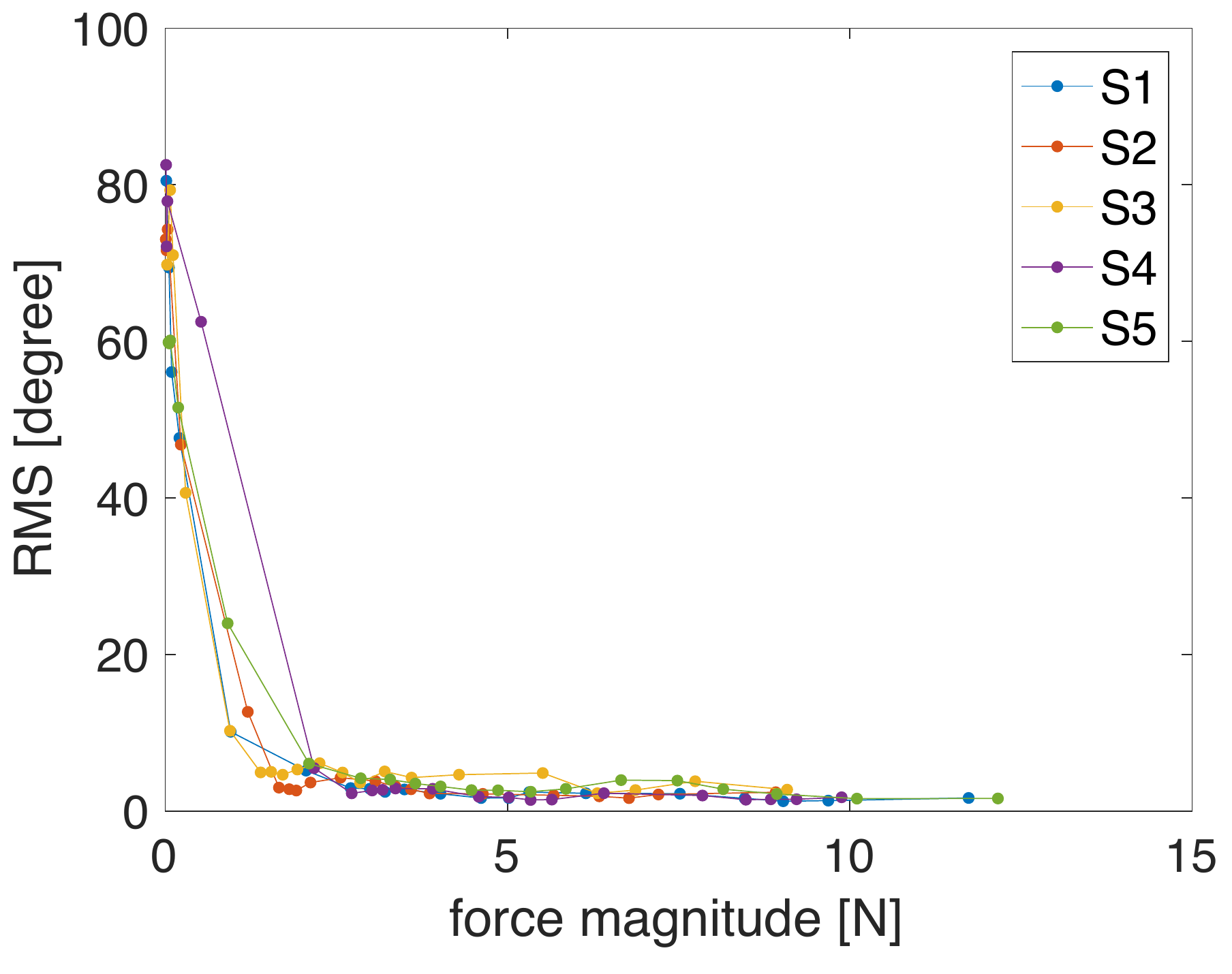}
	\end{minipage}}\hfill
	\subfigure[RMS error in Normal (grip):Tangential (load) force ratio. The ratio is possible extremely large when the tangential force is close to zero. Therefore, the data with tangential force less than 0.2 is excluded.]{
		\begin{minipage}[b]{0.47\columnwidth}
			\centering
			\includegraphics[width=\columnwidth]{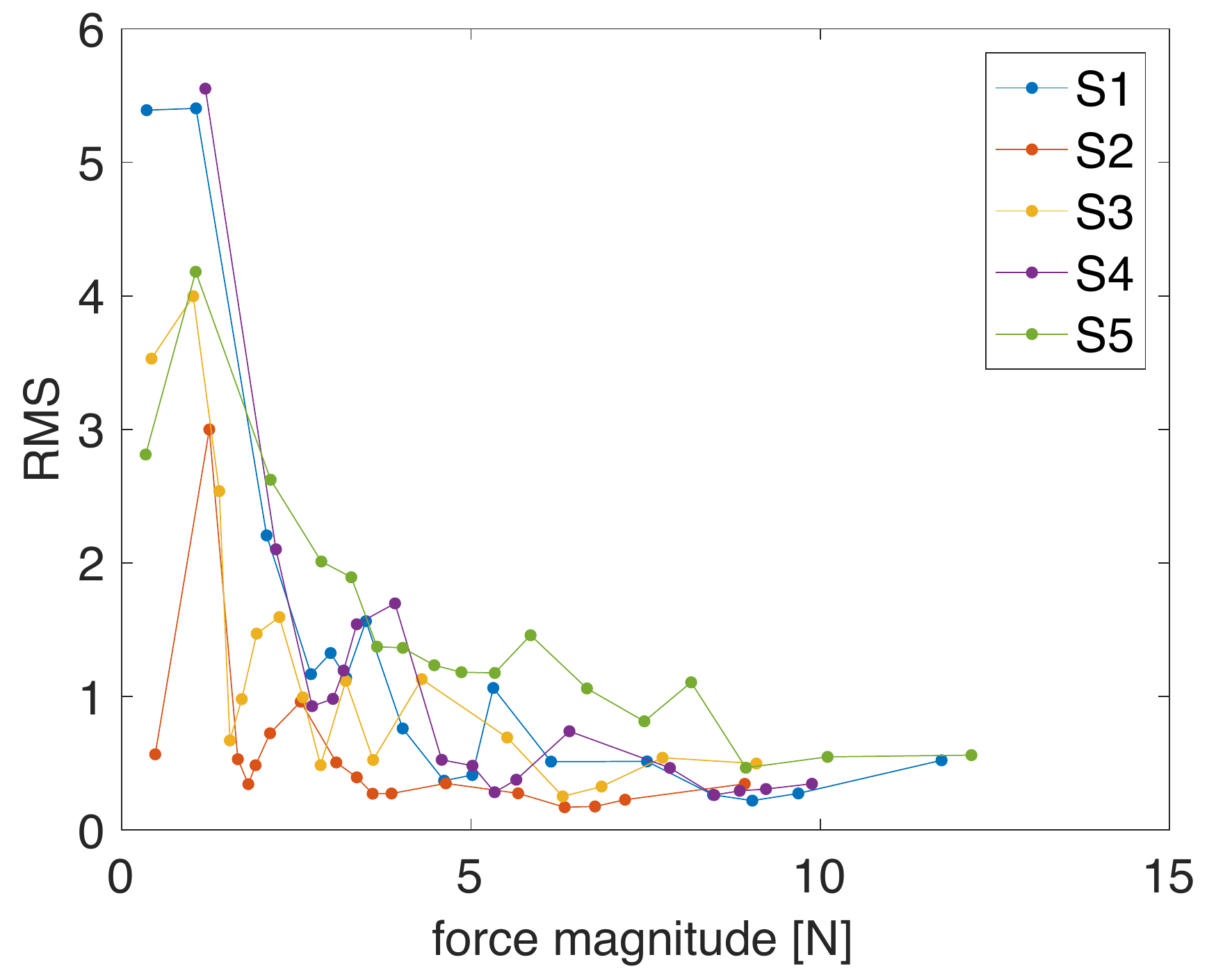}
	\end{minipage}}
	\subfigure[RMS error in torque components $\tau_x$, $\tau_y$, $\tau_z$. The percentiles are computed separately in $\tau_x$, $\tau_y$ and $\tau_z$.]{
		\begin{minipage}[b]{0.47\columnwidth}
			\centering
			\includegraphics[width=\columnwidth]{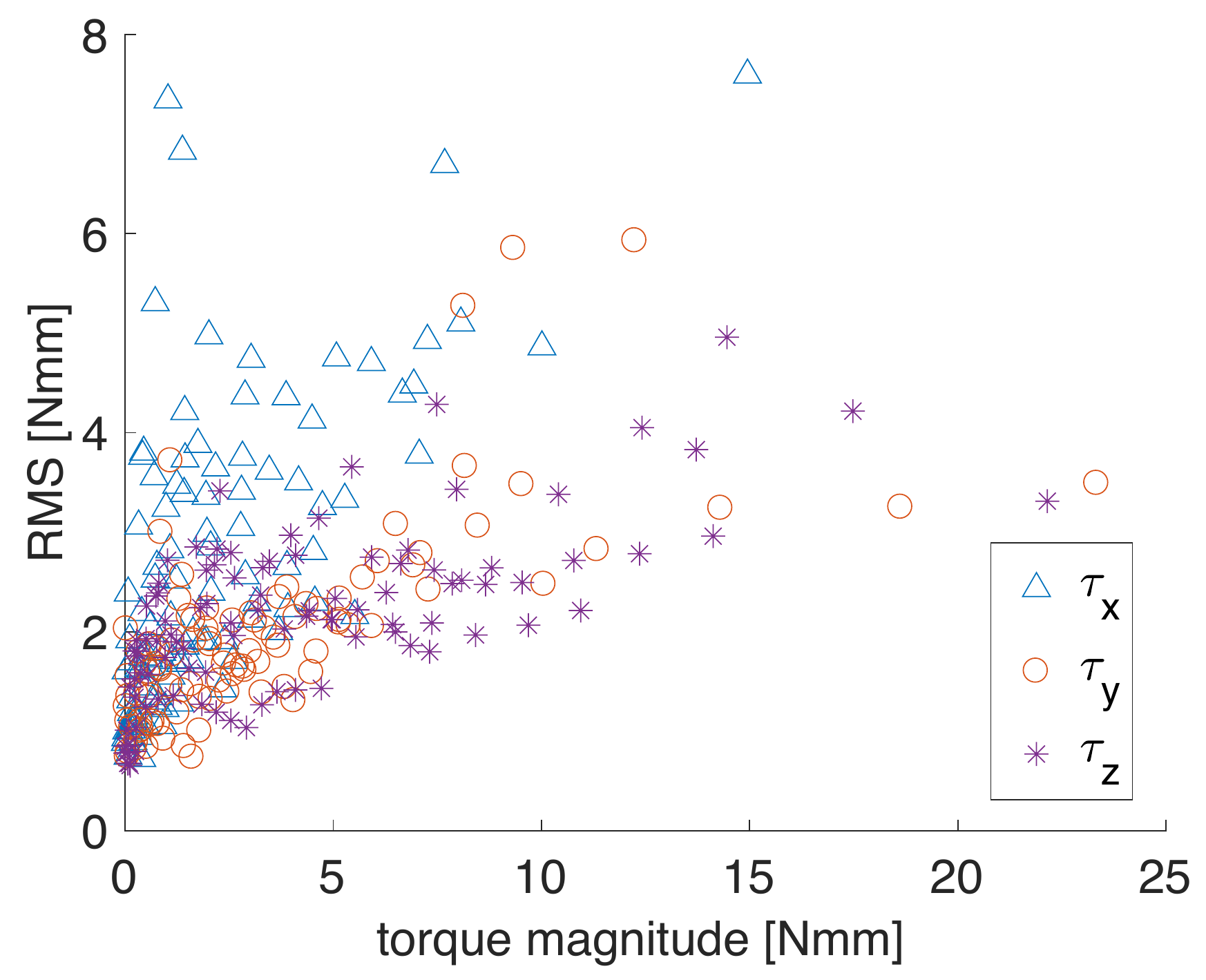}
	\end{minipage}}\\
	\vspace{0 in}
	\caption{Results for index fingers across all 5 participants with individual GP predictor models. One of five series was selected as testing data for each kind of surface and object weight. (a)-(e): Each point represents a percentile of the values in the predicted $f$ or $\tau$ for the percentages in the interval from 0 to 100 with five-percentage steps. The horizontal and the vertical axes are the mean values and the RMS errors of the percentiles respectively.}
	\label{method_compar1}
\end{figure}

\begin{figure}[!ht]
	\centering
	\subfigure[The output error.]{
		\begin{minipage}[b]{0.48\columnwidth}
			\centering
			\includegraphics[width=\columnwidth]{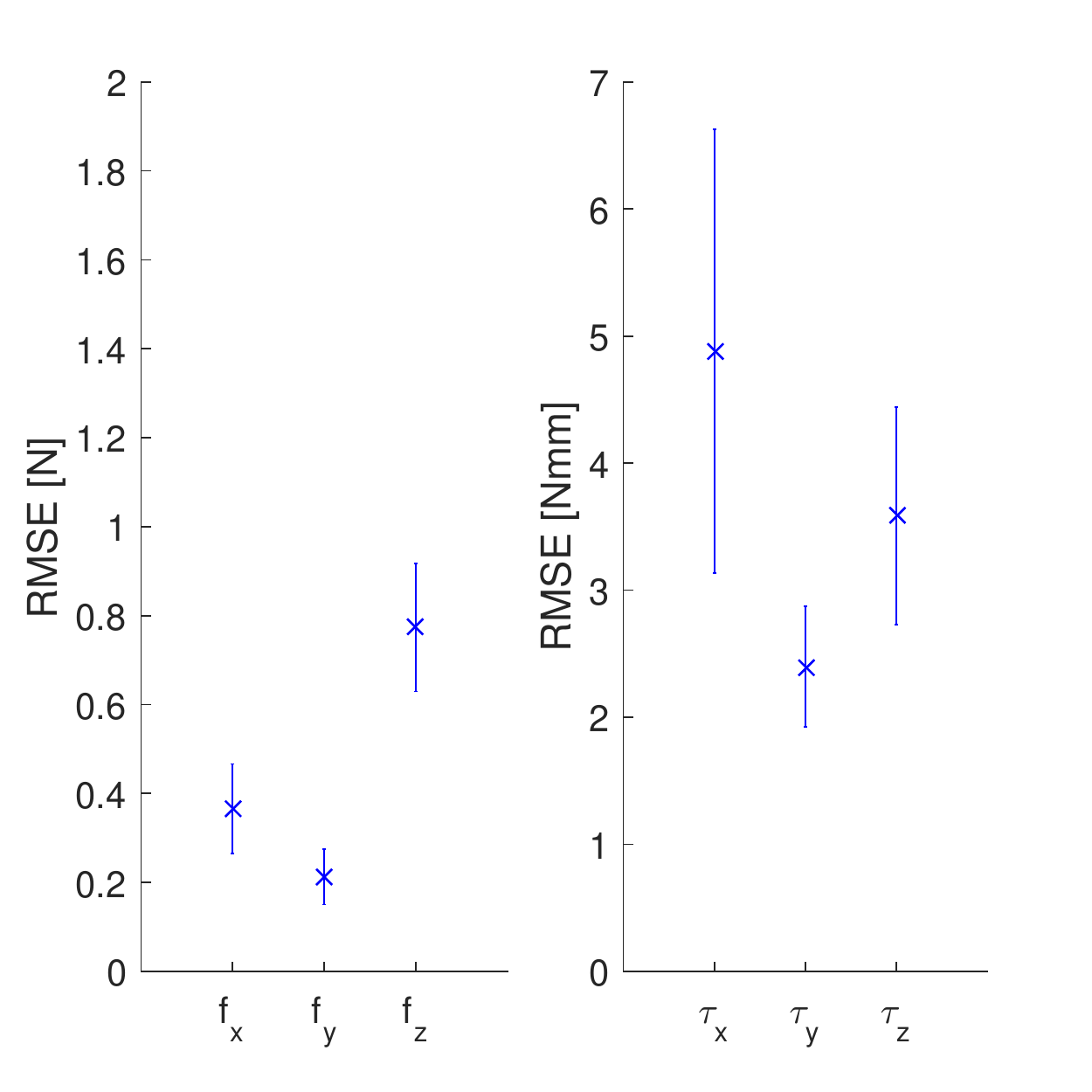}
	\end{minipage}}
	\subfigure[The output standard deviation.]{
		\begin{minipage}[b]{0.48\columnwidth}
			\centering
			\includegraphics[width=\columnwidth]{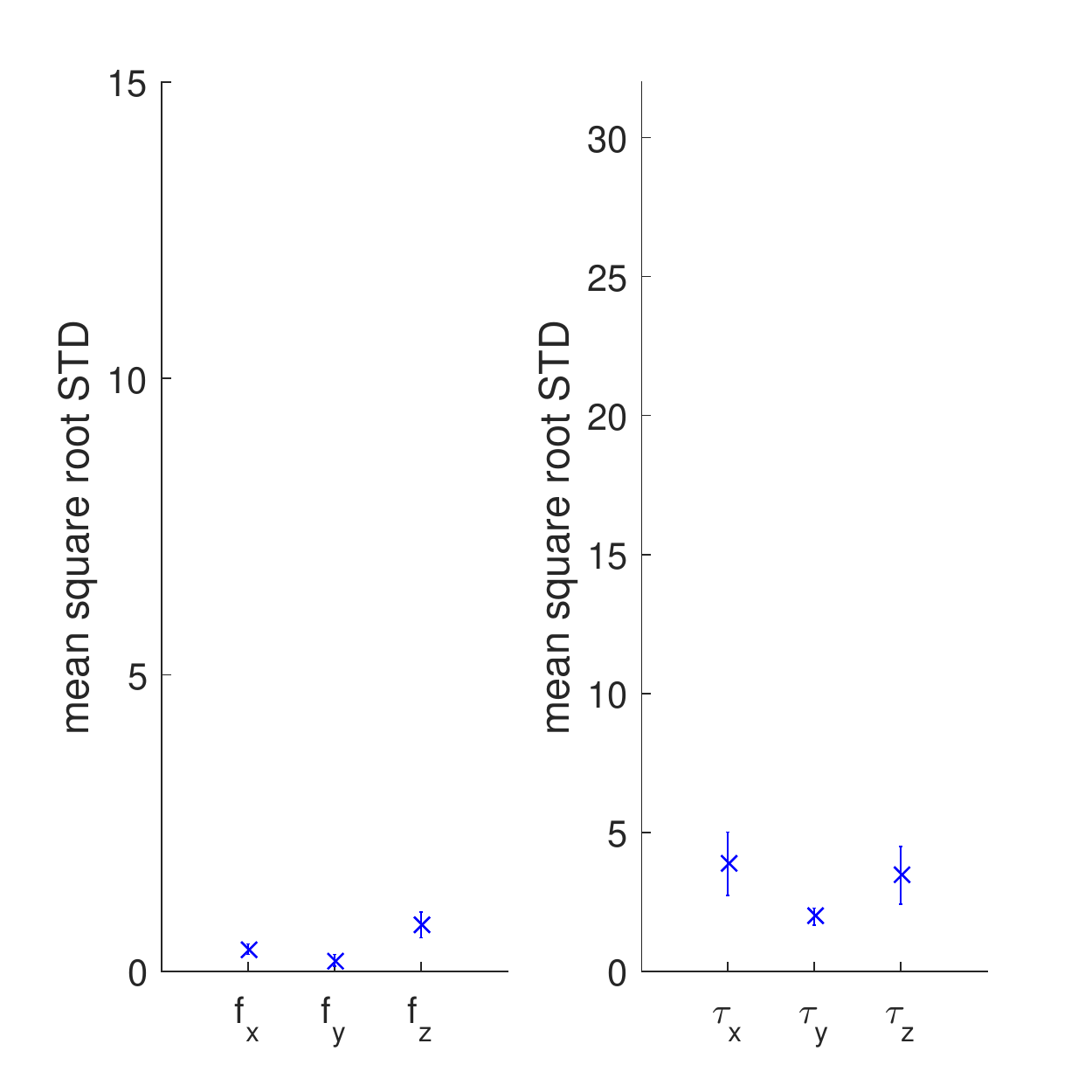}
	\end{minipage}}\\
	\vspace{0 in}
	\caption{Prediction of the thumbs across all 5 participants using GP.}
	\label{thumb_result}
\end{figure}

\begin{figure}
	\centering
	\subfigure[RMS error in force components $f_x$, $f_y$, $f_z$.]{
		\begin{minipage}[b]{0.47\columnwidth}
			\centering
			\includegraphics[width=\columnwidth]{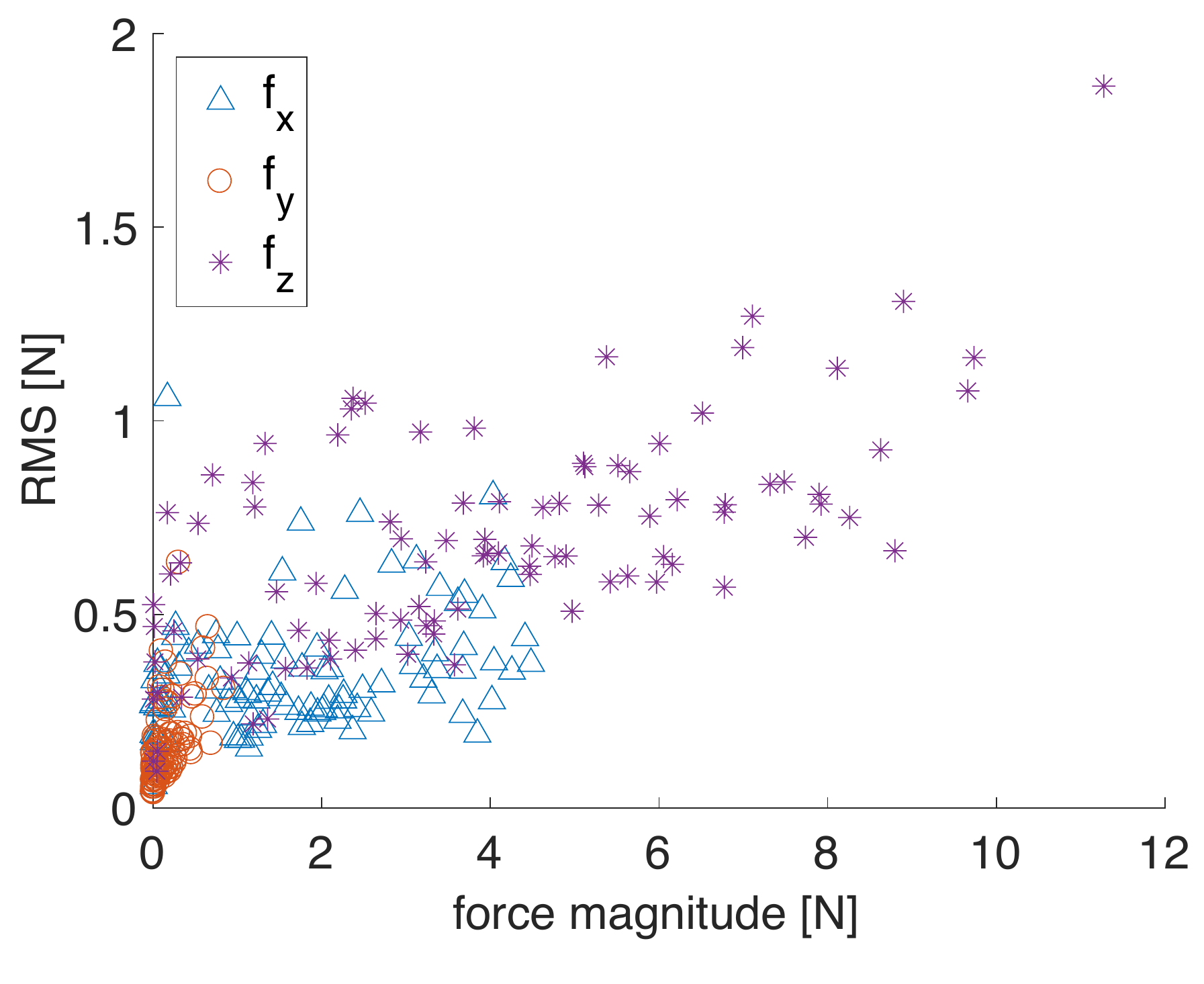}
	\end{minipage}}\hfill
	\subfigure[RMS error in predicted force length magnitude.]{
		\begin{minipage}[b]{0.47\columnwidth}
			\centering
			\includegraphics[width=\columnwidth]{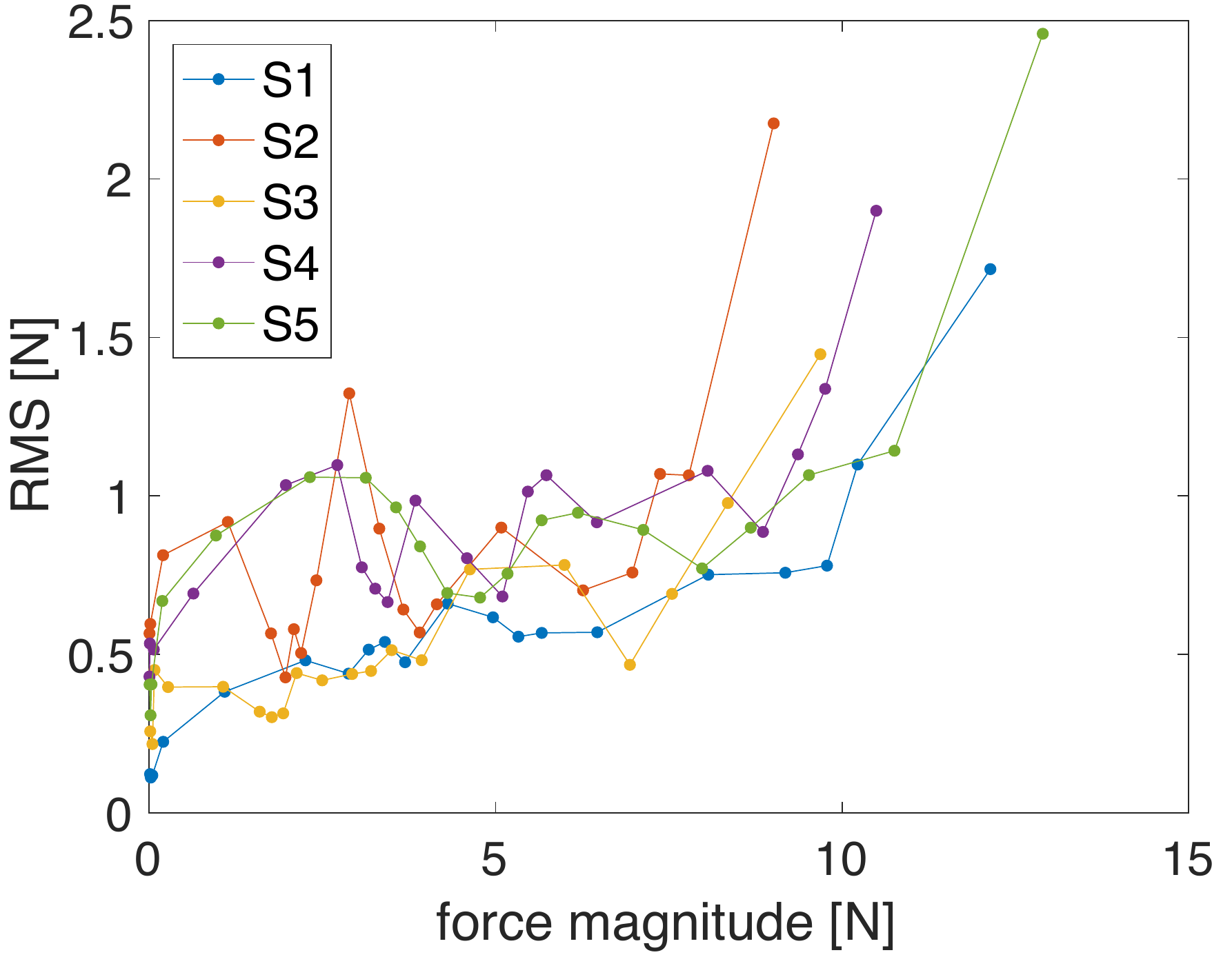}
	\end{minipage}}\\
	\subfigure[RMS error in angle between predicted and surface normal ($f_z$).]{
		\begin{minipage}[b]{0.47\columnwidth}
			\centering
			\includegraphics[width=\columnwidth]{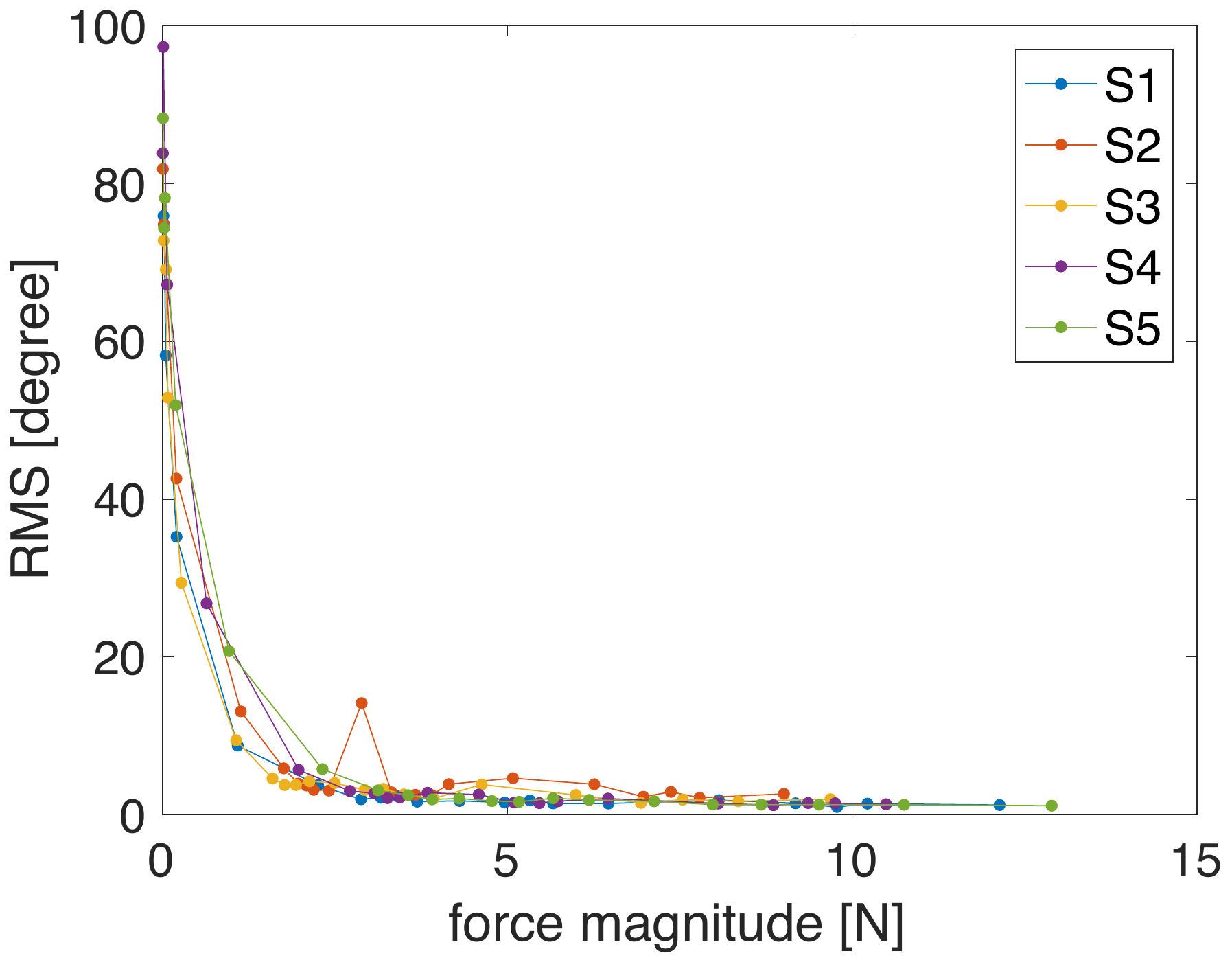}
	\end{minipage}}\hfill
	\subfigure[RMS error in Normal:Tangential force ratio.]{
		\begin{minipage}[b]{0.47\columnwidth}
			\centering
			\includegraphics[width=\columnwidth]{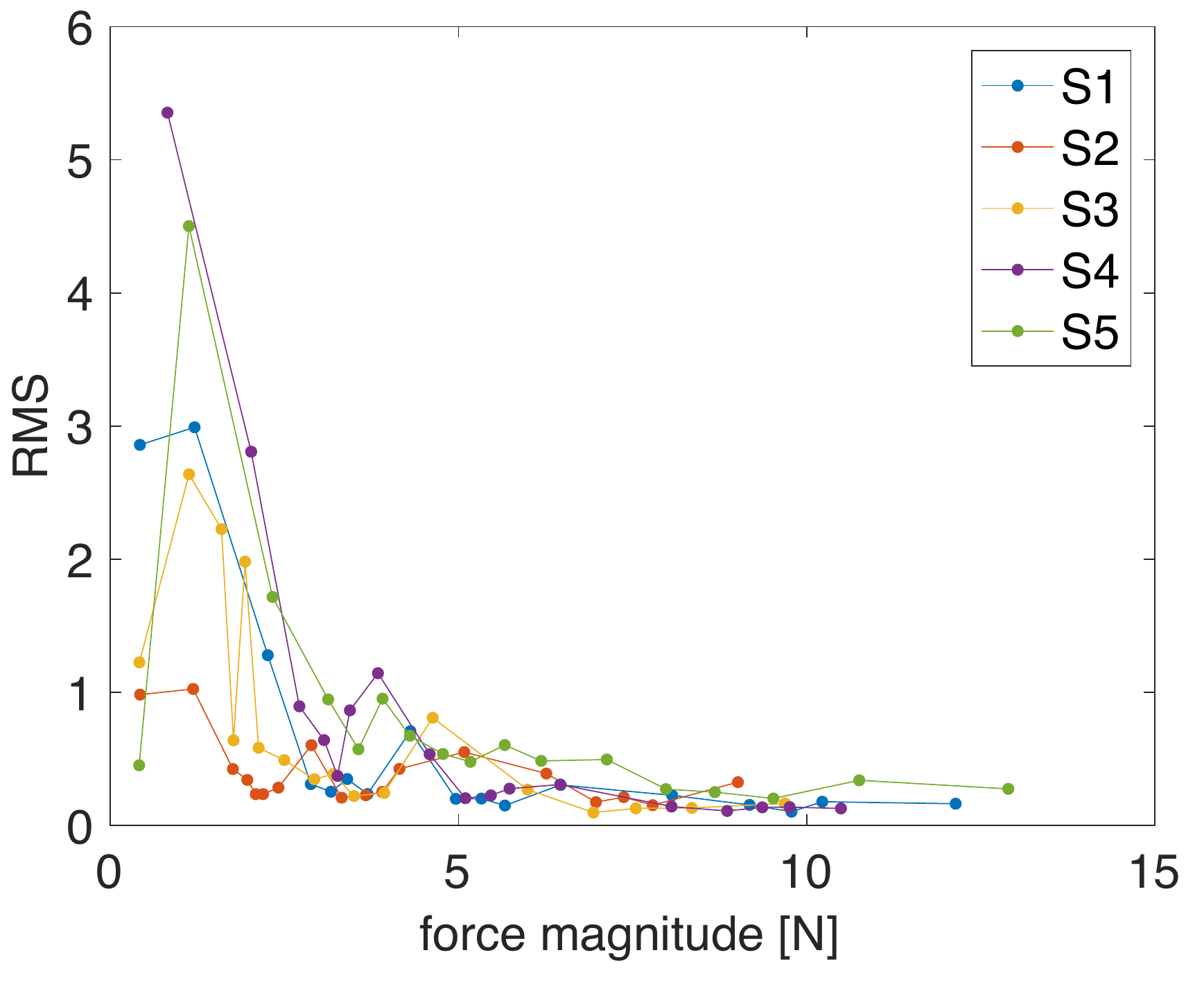}
	\end{minipage}}\hfill
	\subfigure[RMS error in torque components $\tau_x$, $\tau_y$, $\tau_z$.]{
		\begin{minipage}[b]{0.47\columnwidth}
			\centering
			\includegraphics[width=\columnwidth]{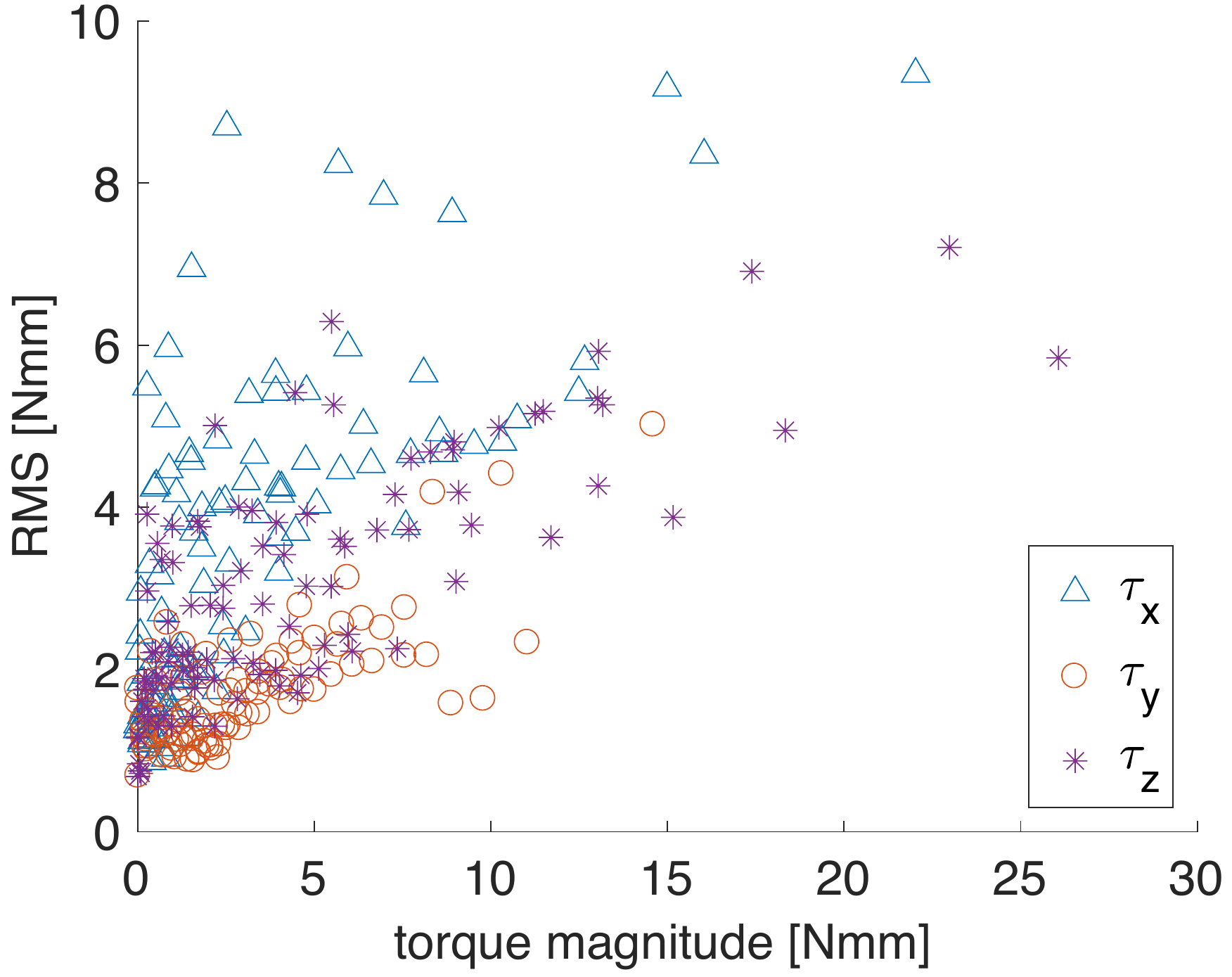}
	\end{minipage}}\\
	\vspace{0 in}
	\caption{Prediction of the thumbs across all 5 participants using GP. One of five series was selected as testing data set for each weight-surface combination. More annotations can be found in Fig.~\ref{method_compar1}.}
	\label{thumb_result1}
\end{figure}

\begin{figure}
\centering
	\subfigure[The result of force/torque prediction. In each surface type the grasping is ordered by three types of weights. For observation, the time between two grasping is set to 0, and the grasping of different surfaces is separated by dashed lines instead.]{
      		\begin{minipage}[b]{0.48\columnwidth}
      			\includegraphics[width=\textwidth]{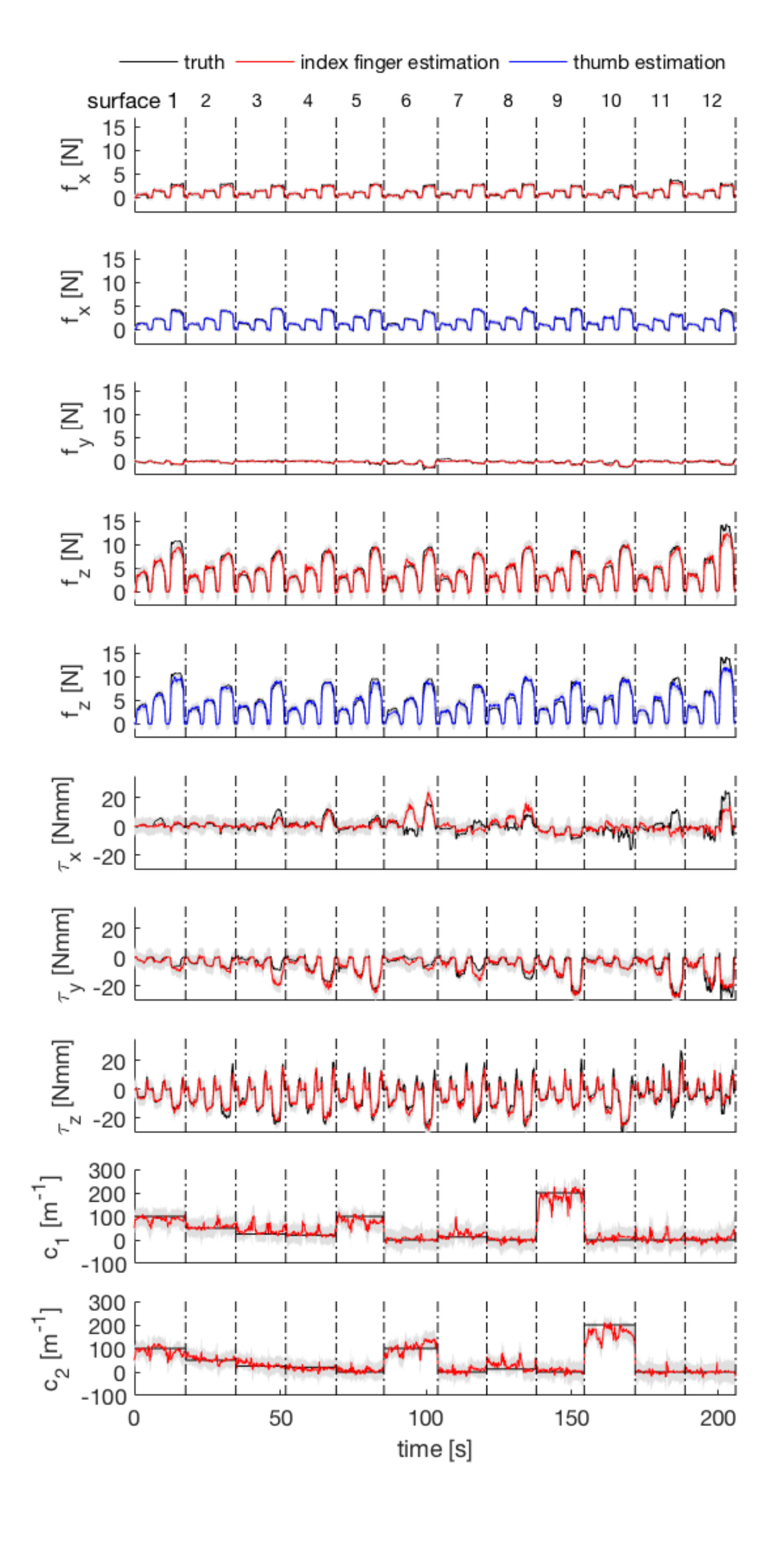}
      			\label{result_f1}
		\end{minipage}}
	 \subfigure[The error of force/torque prediction.]{
	 	\begin{minipage}[b]{0.48\columnwidth}
      			\centering
      			\includegraphics[width=\textwidth]{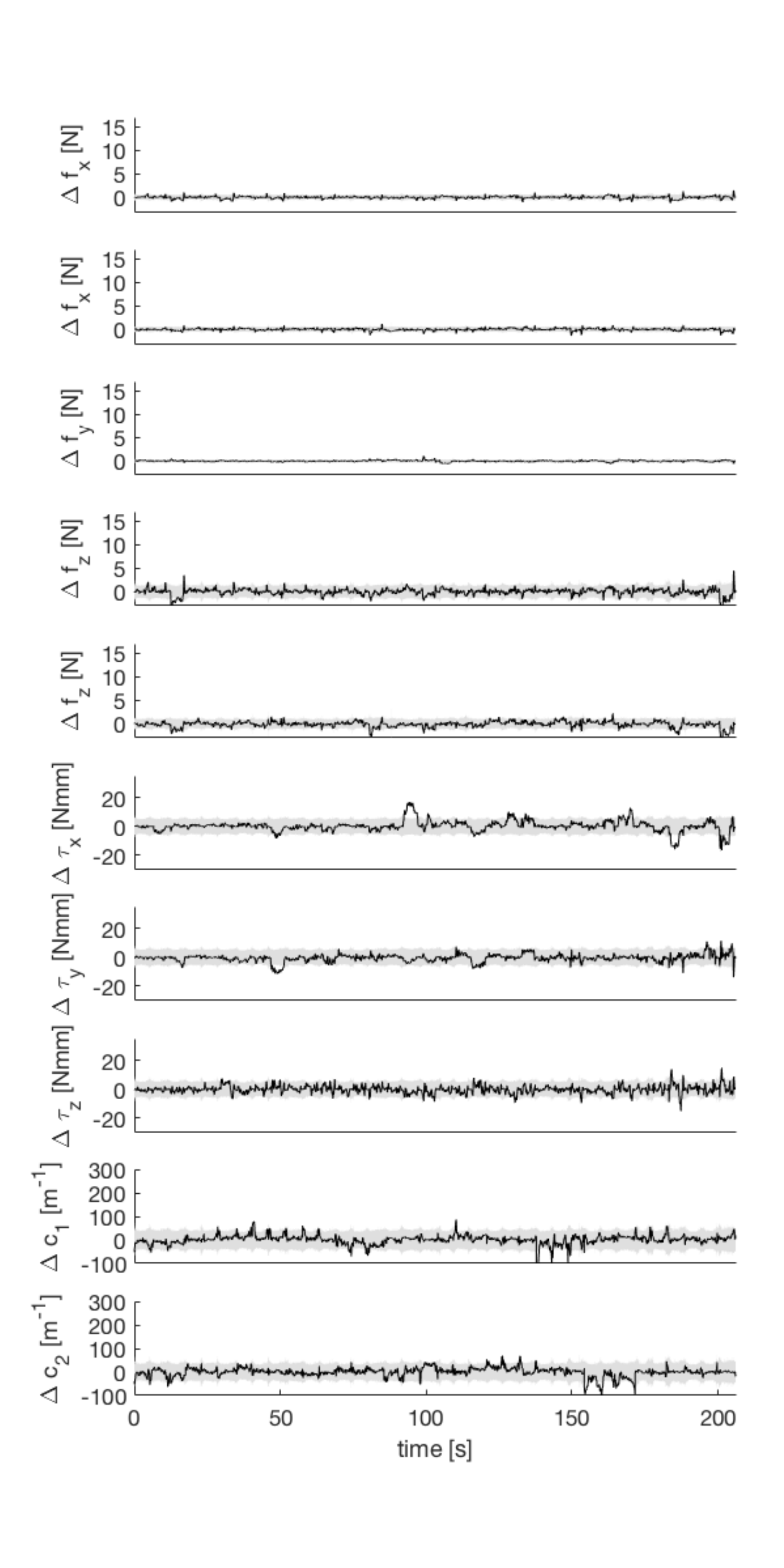}
 			\label{result_f2}
  		\end{minipage}}
	\caption{Prediction for participant $P_1$ using GP. The grey areas are the $95\%$ confidence interval. There are 36 trials.}
\end{figure}

\begin{table}
\caption{Data set range.}
\begin{center}
\begin{tabular}{lllllll}
\hline\\[-0.5ex]
 $f_x$     [-0.9, 4]\,N & $f_y$ [-2, 0.8]\,N   & $f_z$ [0, 15]\,N  \\ [0.5ex]
 $\tau_x$ [-22,26]\,Nmm  & $\tau_y$ [-30, 14]\,Nmm   & $\tau_z$ [-35, 27]\,Nmm \\ [0.5ex]
  $c_1$ $[0, 200]\,\mathrm{m}^{-1}$  & $c_2$  $[0, 200]\,\mathrm{m}^{-1}$    &  \\ [0.5ex]
\hline
\label{table:range}
\end{tabular}
\end{center}
\end{table}

For evaluating GP, one of five series of each weight-surface combination was included in the testing data set, i.e., 20\% of the data was used for testing and 80\% for training. In addition, to avoid over-fitting with CNN, NN-FD and RNN-FD, 25\% of the GP training data was selected as validation data and the remaining 75\% was training data while the testing data set was the same as that for the GP evaluation. The architecture of CNN was described in \ref{sec:cnn}. RNN-FD had one tanh hidden layer with 100 units and an identity output layer. The input and the hidden layers both had 0.5 probability to be dropped. NN-FD had three tanh hidden layers with 170, 120 and 35 units and an identity output layer. The input and the three hidden layers had 0.3 probability to be dropped. The architectures were obtained by hyper-paramter search during training.

Although each subject had a slight different range of force and torque, Table \ref{table:range} provides representative summary data. CNN, NN-FD and RNN-FD were implemented on deep-learning frameworks.
The outputs from the four predictor models for the index fingers across all 5 participants are shown in Fig.~\ref{method_compar}.

The accuracy of GP was slightly higher than that of CNN (Fig.~\ref{method_compar}). On the other hand, while CNN is unable to output error estimates, it was considerably faster than GP. 
Because GP and CNN were more accurate than NN-FD and RNN-FD,
we focused our analyses on the former.
Fig.~\ref{method_compar1} shows the results of individual GP predictor models created for the index finger of each of the five participants. 

Multiple-finger prediction is crucial for manipulation applications. Figs.~\ref{thumb_result1},\ref{thumb_result} show that the GP accurately predicts the forces, torques and surface curvatures of the thumb.
Furthermore, Figs.~\ref{result_f1},\ref{result_f2} show an example of the prediction of participant $P_1$ and Figs.~\ref{pick_replace_material},\ref{pick_replace_} illustrate further details of several trials of picking up and replacing an object. The modle was able to predict normal forces of higher than 10\,N, although when it was close to 15\,N it was not as accurate as at lower forces. 

The predicted surface normal forces, $f_z$, of the thumb and index finger were approximately equal to each other (if not, it would imply that the object was accelerated sideways).
Moreover, when the object was held stationary in air, the sum of the vertical forces applied by the thumb and index finger counteracted the weight of the object.
This also fitted well with model predictions: the
$x$ axes of the fingers were almost vertical and the the sum of the estimated $f_x$ of the thumb and index finger was indeed approximately equal to the force required to hold the object stationary in air.

In low force frames, the fingernail images of different surfaces were very similar to each other,
which caused the surface curvature predictions of $c_1$ and $c_2$ to be less accurate than in high-force frames.

\subsection{Surface cross validation}

\begin{figure}
\centering
  \subfigure[RMS error in force components $f_x$, $f_y$, $f_z$.]{
    \begin{minipage}[b]{0.47\columnwidth}
      \centering
      \includegraphics[width=\textwidth]{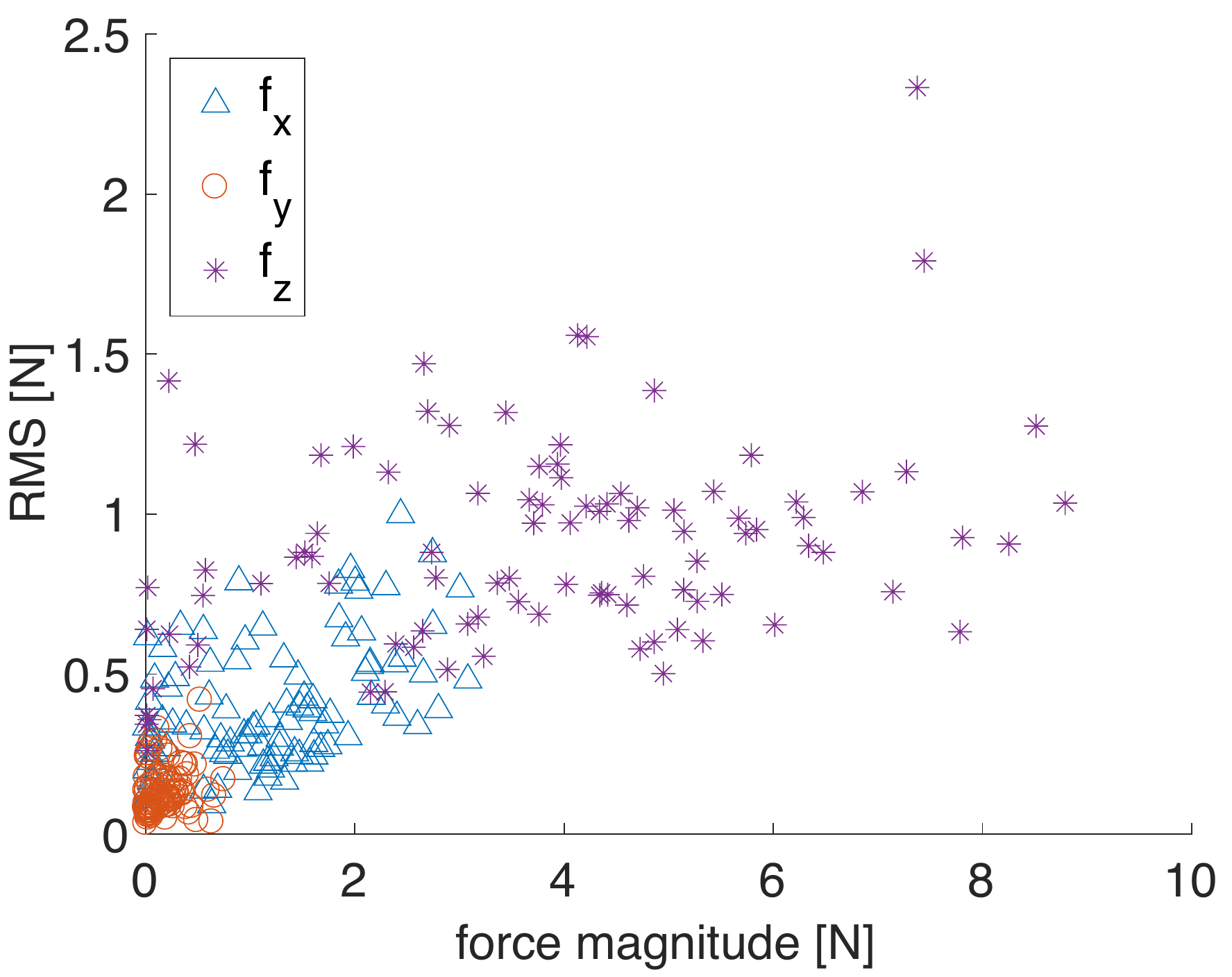}
    \end{minipage}}\hfill
  \subfigure[RMS error in predicted force length magnitude.]{
    \begin{minipage}[b]{0.47\columnwidth}
      \centering
      \includegraphics[width=\columnwidth]{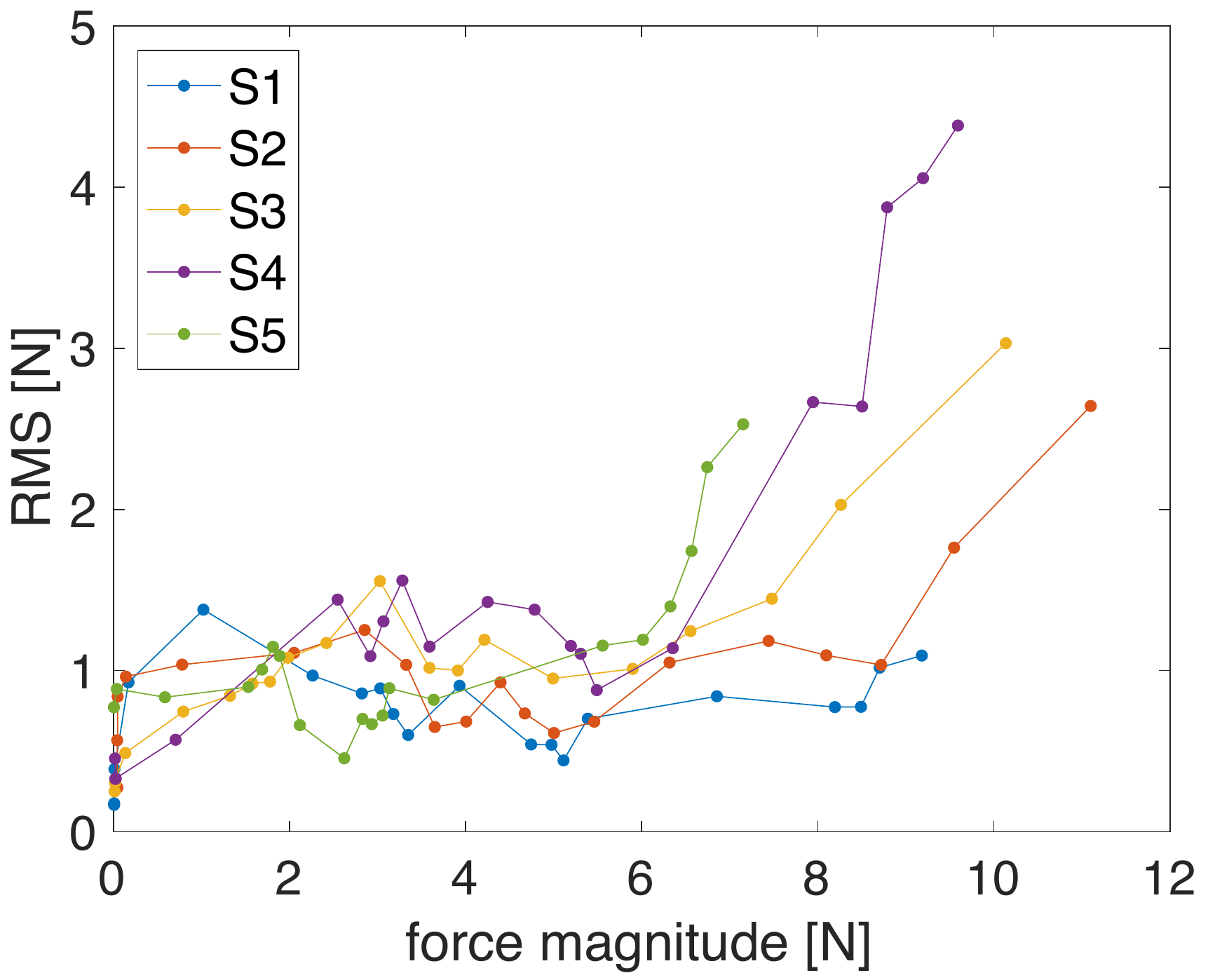}
    \end{minipage}}\\
  \subfigure[RMS error in angle between predicted and surface normal ($f_z$).]{
    \begin{minipage}[b]{0.47\columnwidth}
      \centering
      \includegraphics[width=\textwidth]{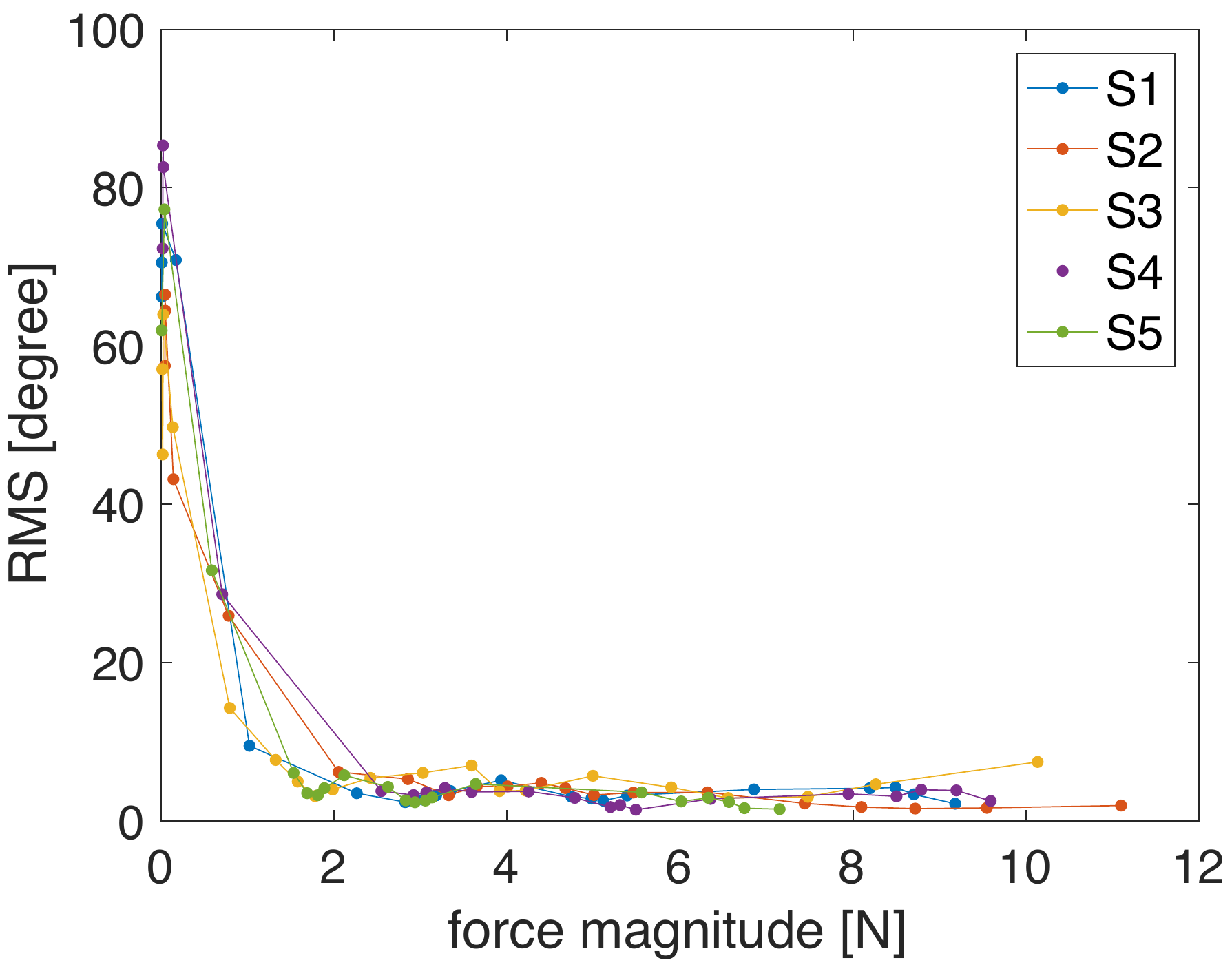}
    \end{minipage}} \hfill
  \subfigure[RMS error in Normal:Tangential force ratio.]{
    \begin{minipage}[b]{0.47\columnwidth}
      \centering
      \includegraphics[width=\columnwidth]{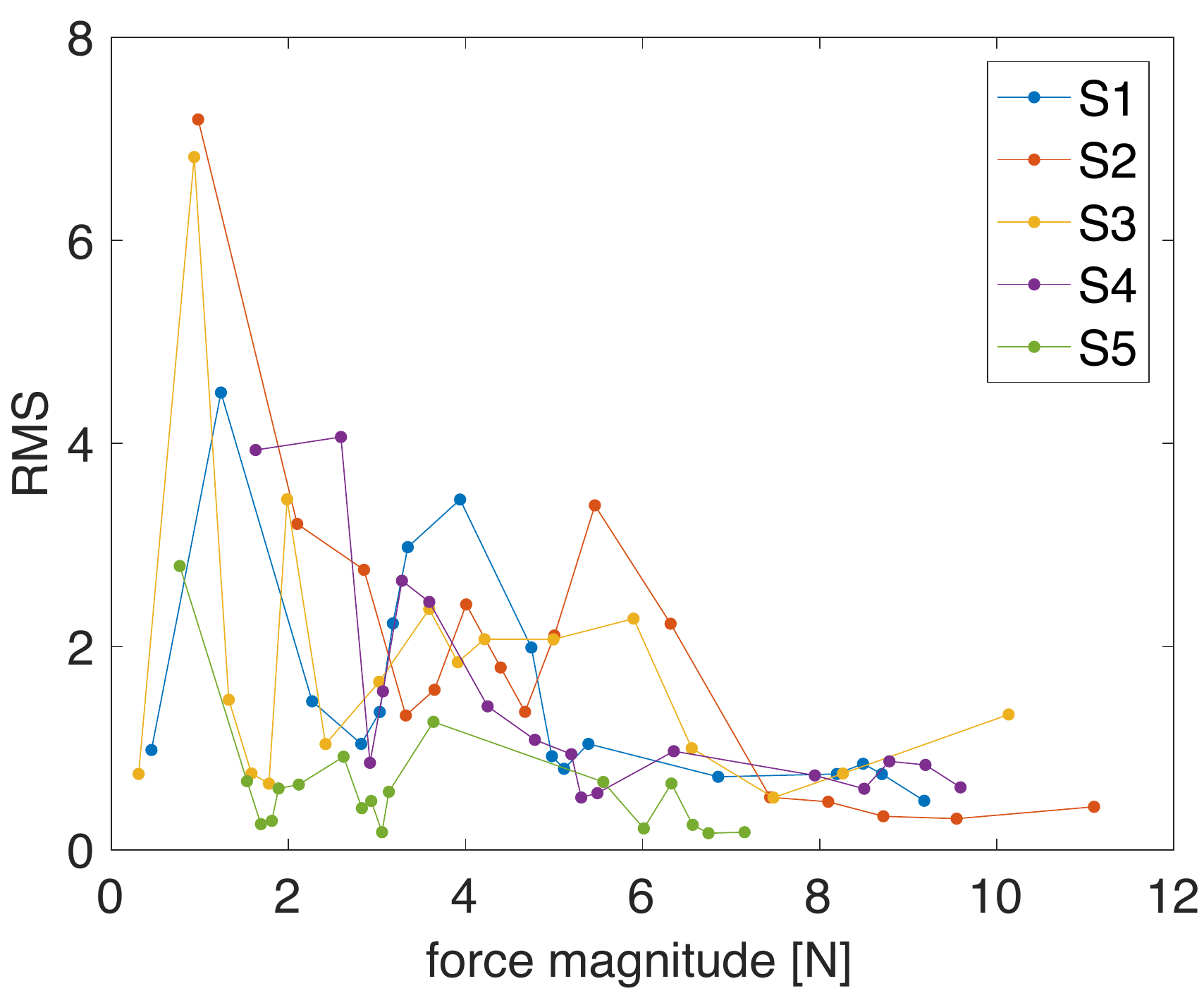}
    \end{minipage}} \\
  \subfigure[RMS error in torque components $\tau_x$, $\tau_y$, $\tau_z$.]{
    \begin{minipage}[b]{0.47\columnwidth}
      \centering
      \includegraphics[width=\columnwidth]{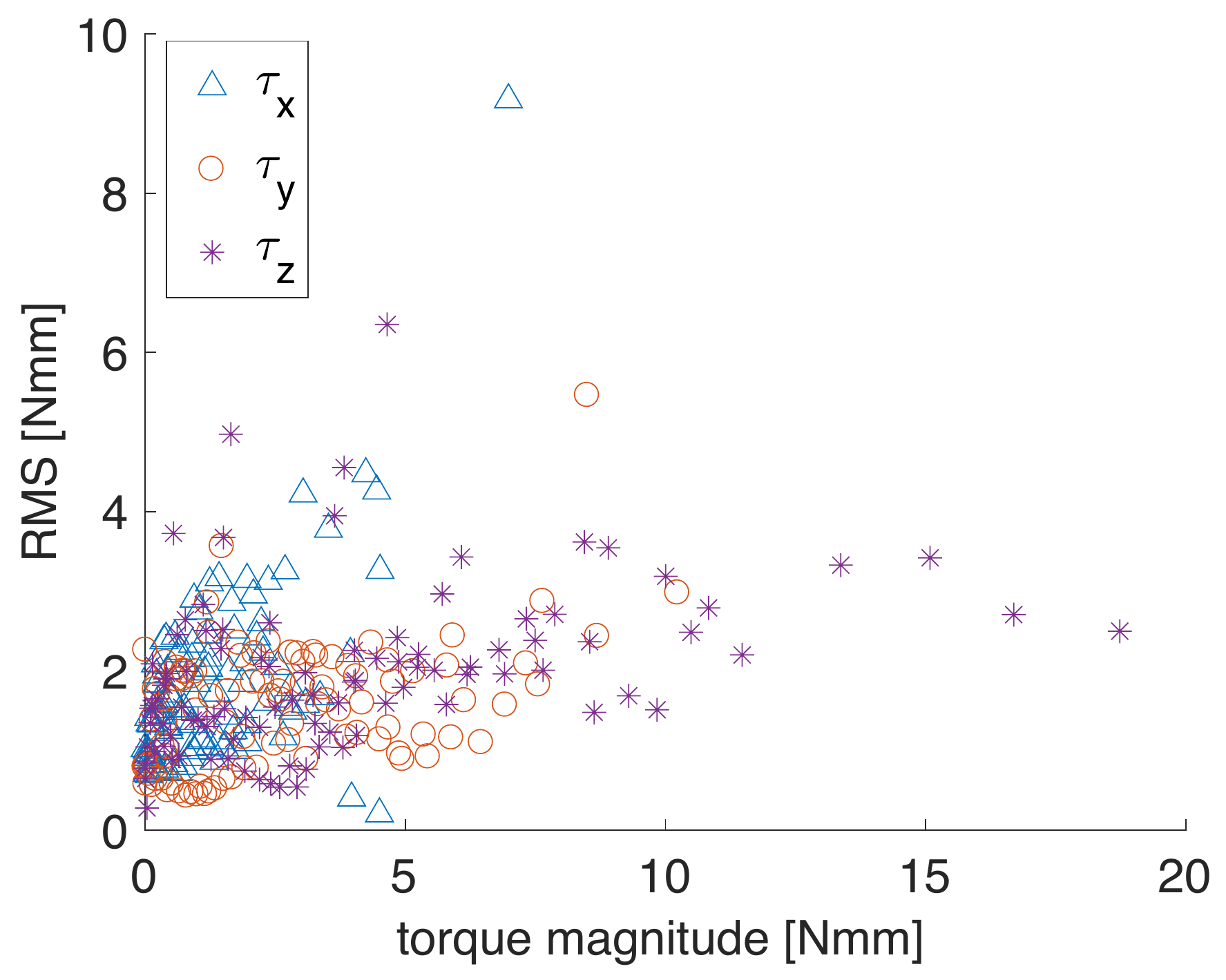}
    \end{minipage}}s
    \vspace{0 in}
  \caption{Predictions across all 5 participants for surface cross validation. The testing data set was surface 3 (cf., Fig.~\ref{setup04}) across all 3 weights. More annotations can be seen in Fig.~\ref{method_compar1}.}
  \label{plot_surface_cross1}
\end{figure}

\begin{figure}[!ht]
\centering
  \subfigure[The output error.]{
    \begin{minipage}[b]{0.45\columnwidth}
      \centering
      \includegraphics[width=\columnwidth]{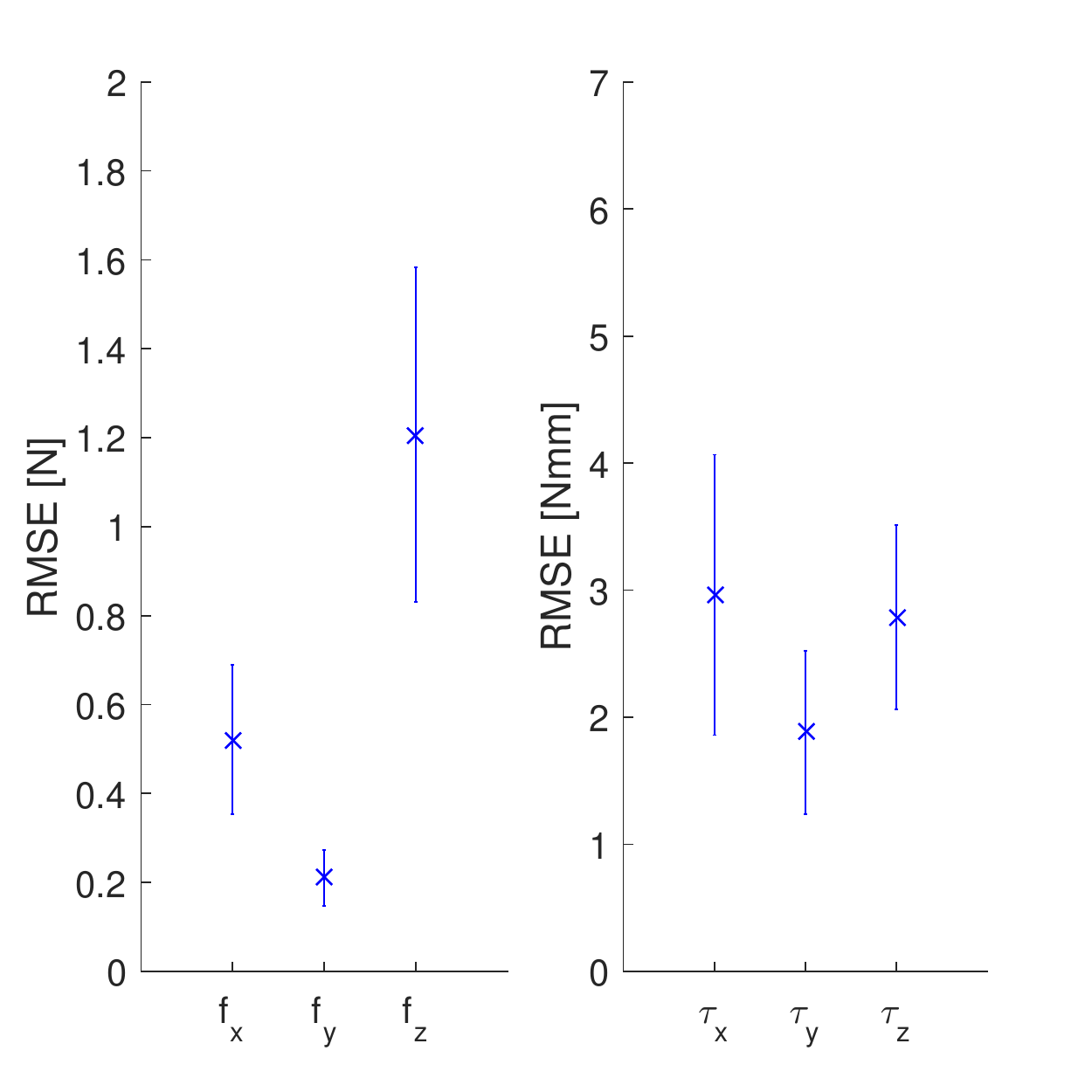}
    \end{minipage}}
  \subfigure[The output standard deviation.]{
    \begin{minipage}[b]{0.45\columnwidth}
      \centering
      \includegraphics[width=\columnwidth]{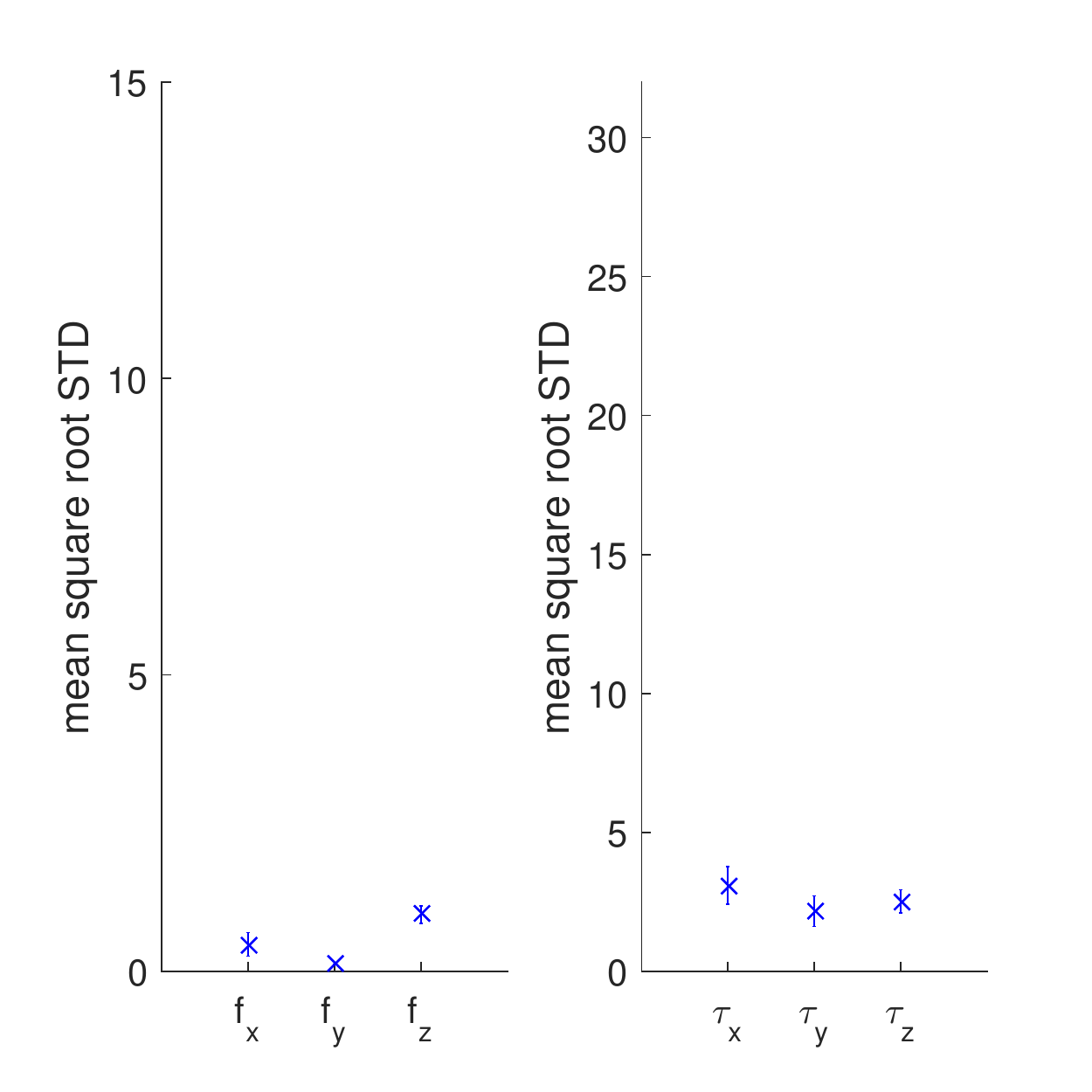}
    \end{minipage}}\\
    \vspace{0 in}
  \caption{Predictions across all 5 participants for surface cross validation using GP.}
  \label{plot_surface_cross}
\end{figure}

Our approach was capable of predicting the force and torque by surface cross validation (see Fig.~\ref{plot_surface_cross1}, \ref{plot_surface_cross}). The testing data set not used during training or validation was in this case surface\,3 (cf., Fig.~\ref{setup04}).

\subsection{A single predictor model across all participants}

\begin{figure}
\centering
  \subfigure[RMS error in force components $f_x$, $f_y$, $f_z$.]{
    \begin{minipage}[b]{0.47\columnwidth}
      \centering
      \includegraphics[width=\columnwidth]{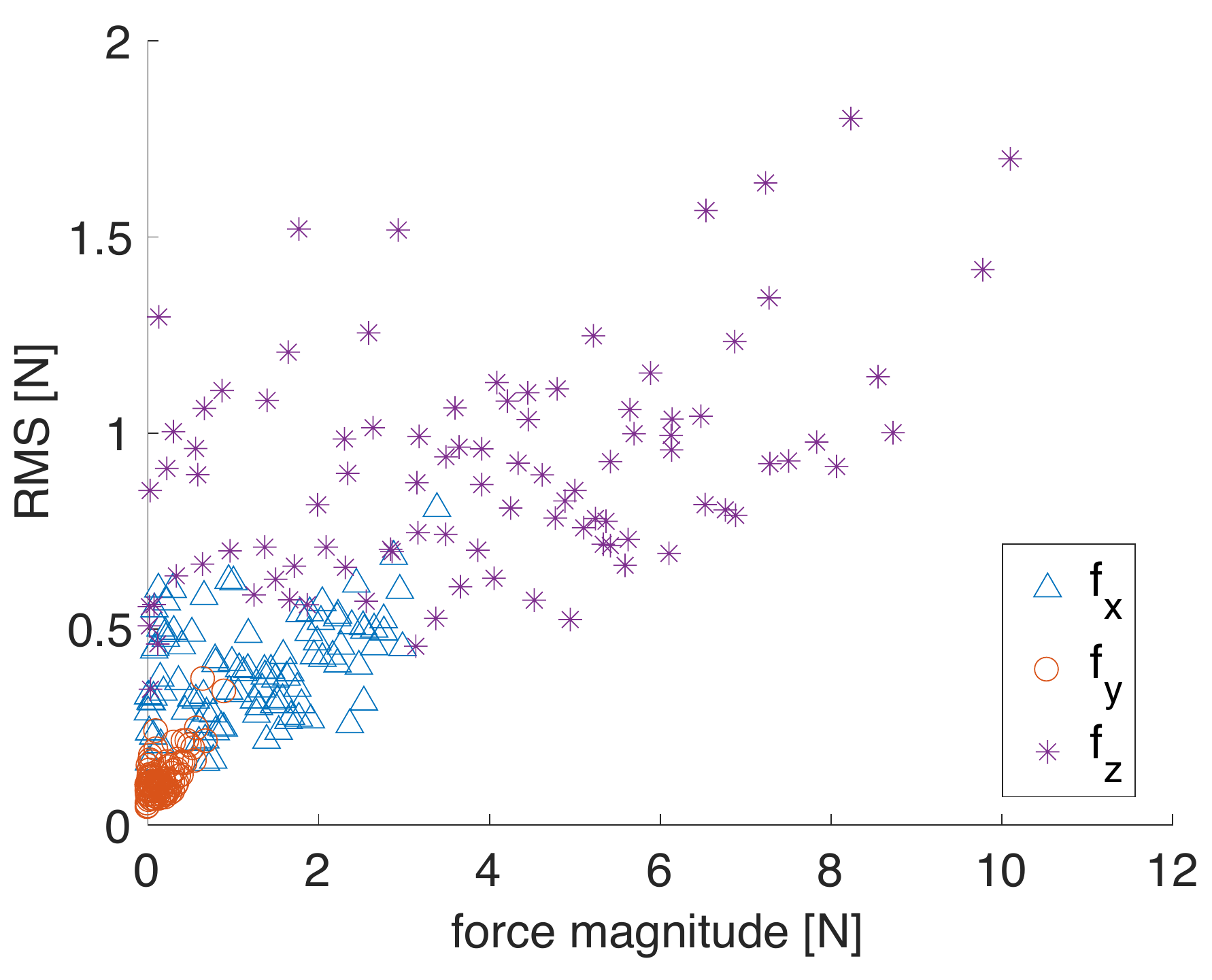}
    \end{minipage}}\hfill
  \subfigure[RMS error in predicted force length magnitude.]{
    \begin{minipage}[b]{0.47\columnwidth}
      \centering
      \includegraphics[width=\columnwidth]{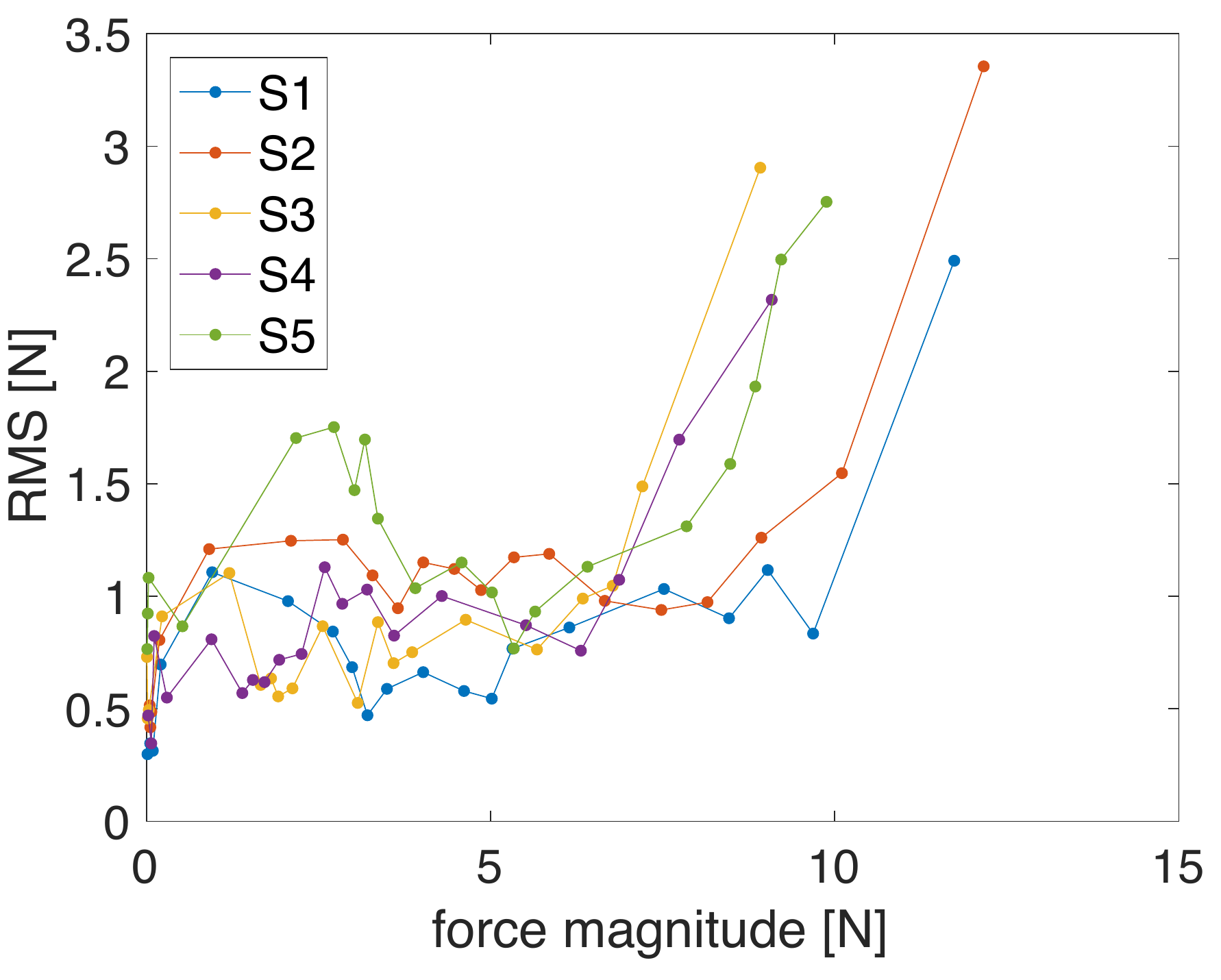}
    \end{minipage}}\\
  \subfigure[RMS error in angle between predicted and surface normal ($F_Z$).]{
    \begin{minipage}[b]{0.47\columnwidth}
      \centering
      \includegraphics[width=\columnwidth]{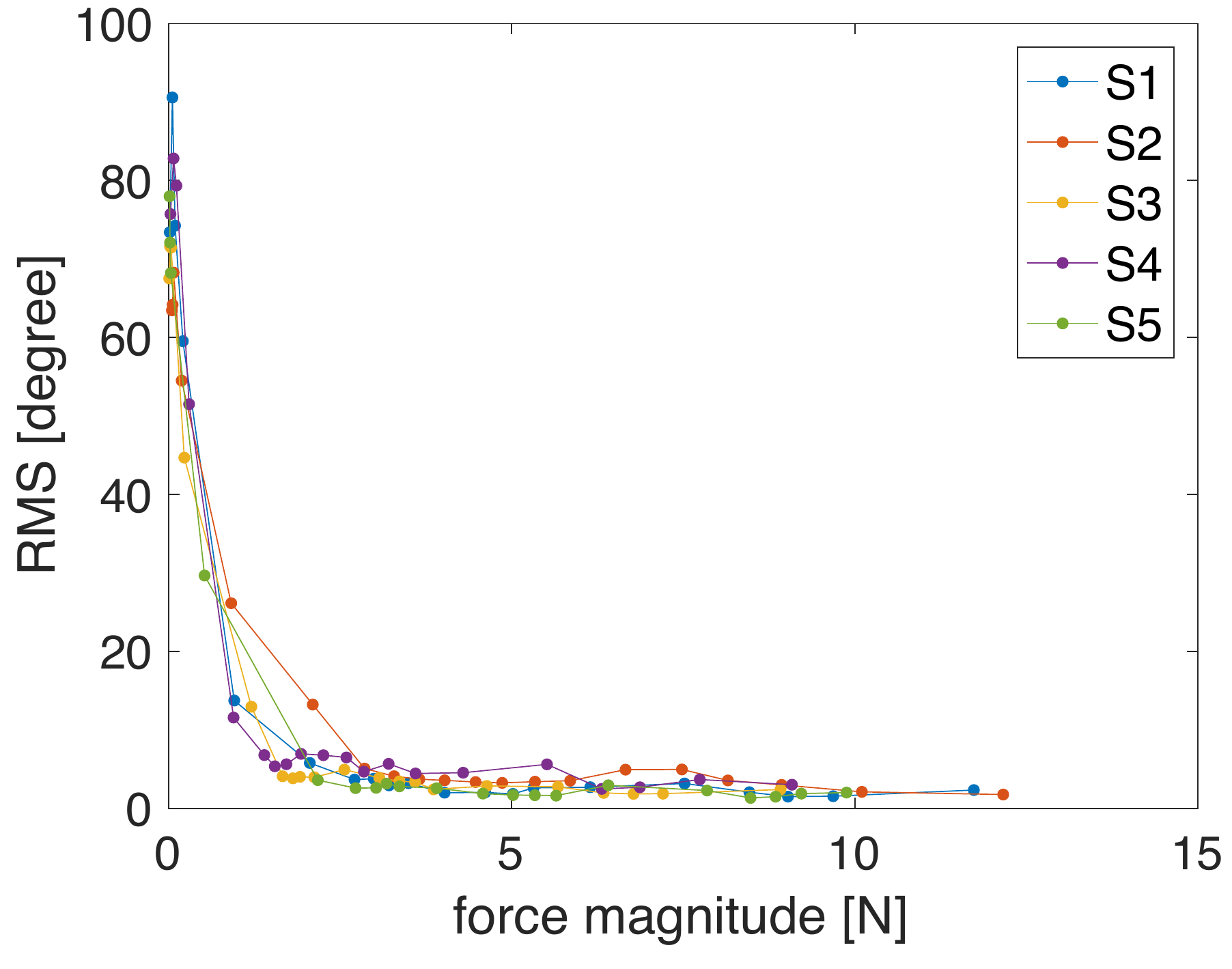}
    \end{minipage}}\hfill
  \subfigure[RMS error in Normal:Tangential force ratio.]{
    \begin{minipage}[b]{0.47\columnwidth}
      \centering
      \includegraphics[width=\columnwidth]{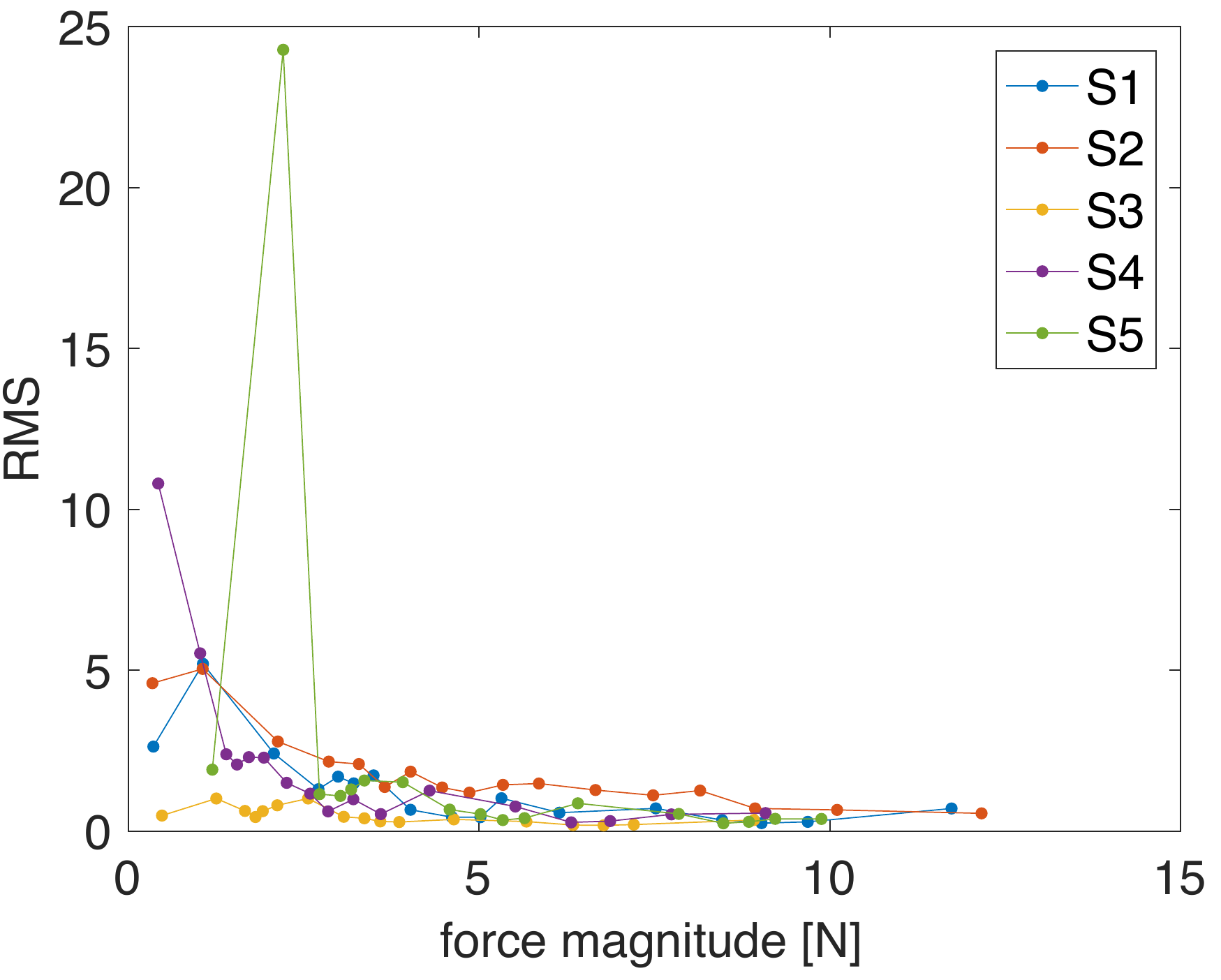}
    \end{minipage}}\\
  \subfigure[RMS error in torque components $\tau_x$, $\tau_y$, $\tau_z$.  For observation, we use the absolute value of the torque.]{
    \begin{minipage}[b]{0.47\columnwidth}
      \centering
      \includegraphics[width=\columnwidth]{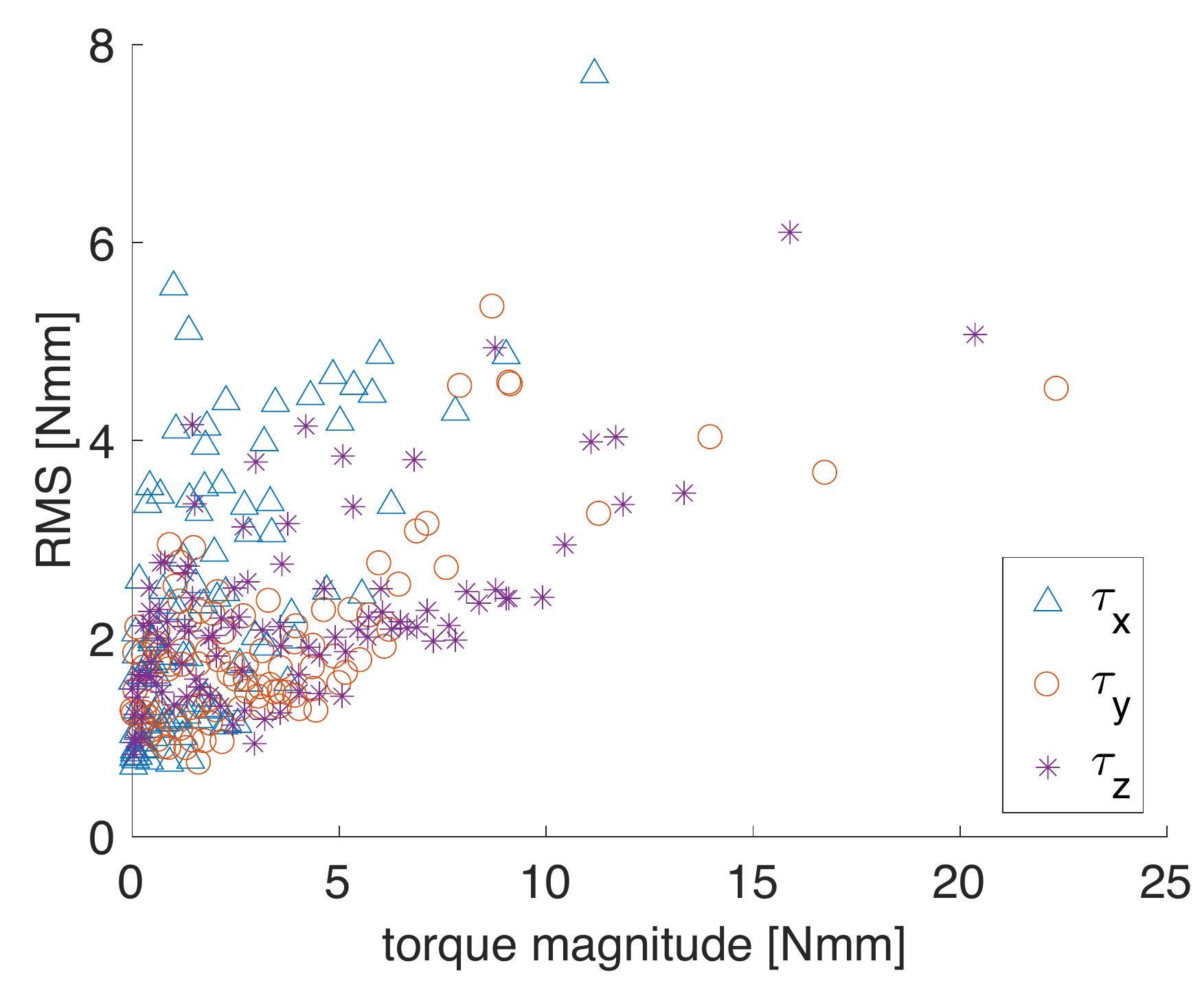}
    \end{minipage}}\\
    \vspace{0 in}
  \caption{Predictions of a single model used across all participants. Testing data was one of five repetitions of all weight-surface combinations. Annotations as in Fig.~\ref{method_compar1}.}
  \label{plot_onemodel1}
\end{figure}

\begin{figure}
\centering
  \subfigure[The output error.]{
    \begin{minipage}[b]{0.45\columnwidth}
      \centering
      \includegraphics[width=\columnwidth]{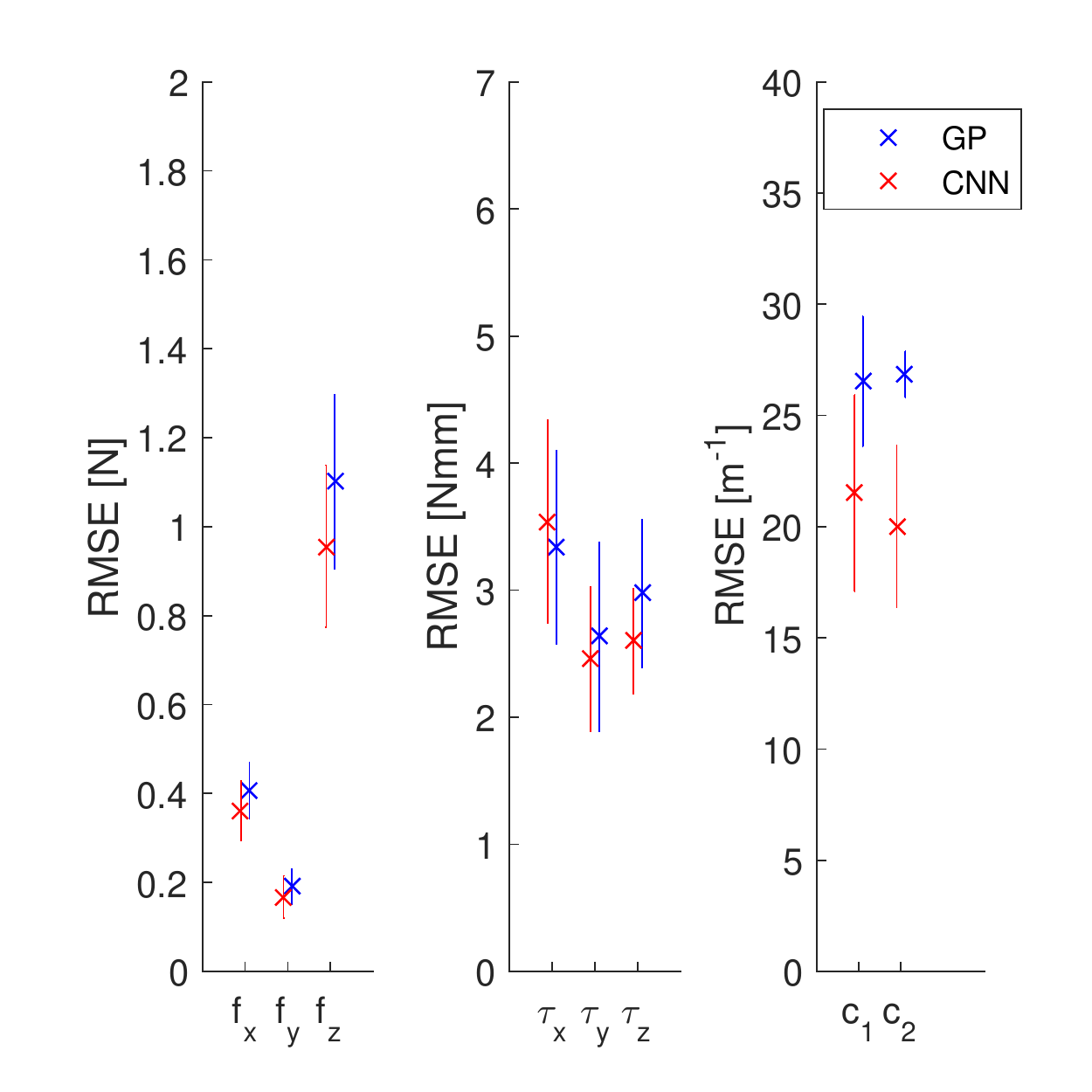}
    \end{minipage}}
  \subfigure[The mean standard deviation output by GP.]{
    \begin{minipage}[b]{0.45\columnwidth}
      \centering
      \includegraphics[width=\columnwidth]{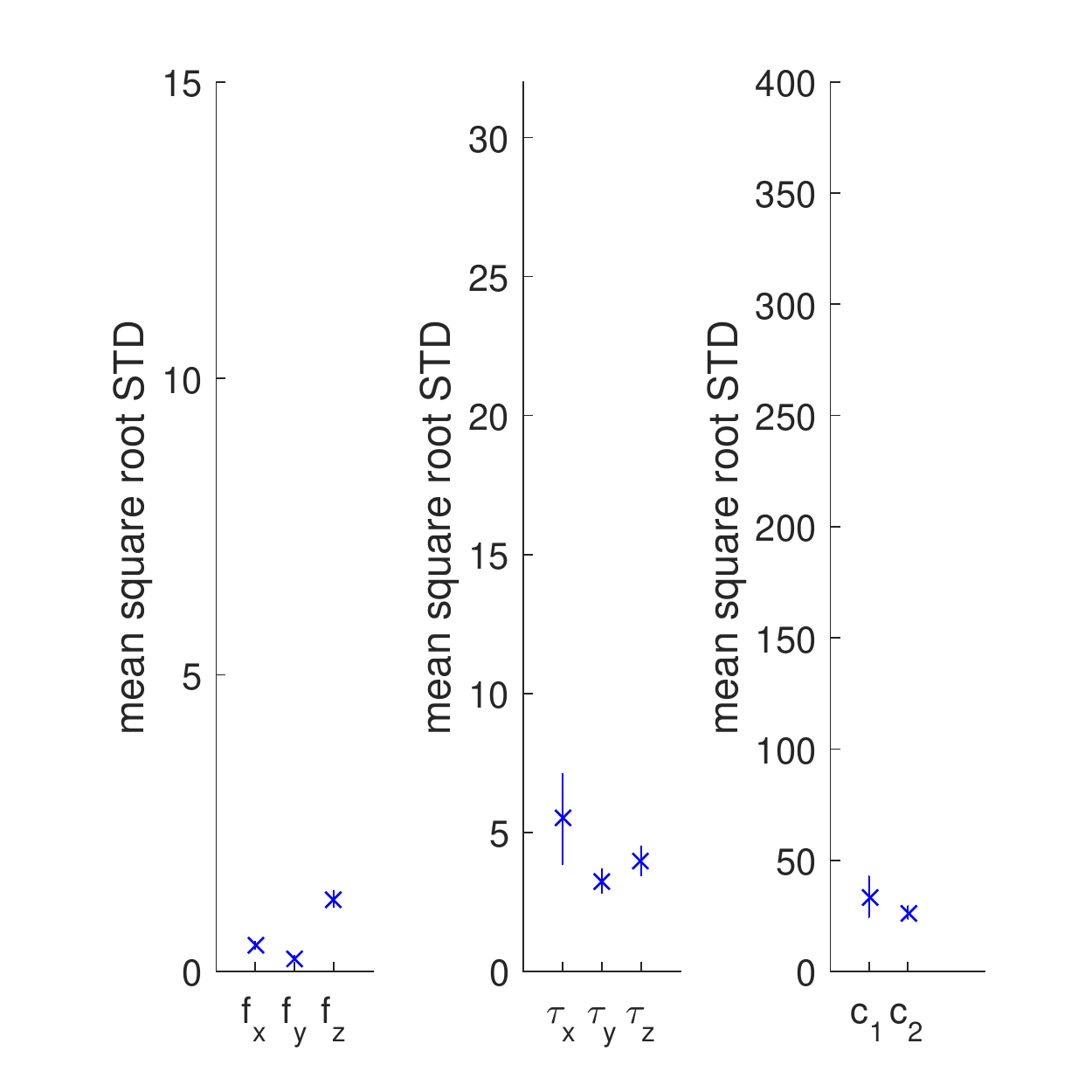}
    \end{minipage}}\\
    \vspace{0 in}
  \caption{Predictions of a single model used across all participants.}
  \label{plot_onemodel}
\end{figure}

In this experiment, the training, validation and testing data sets were the same as in Section~\ref{result1}, but all participants were combined in one model. To handle the large dataset size issue of GP, we randomly reduced the training data set to 30\% and use GP-FITC with 17\% inducing points chosen from the training data. In contrast, CNN readily used all training data. As illustrated in Fig.~\ref{plot_onemodel1},\ref{plot_onemodel}, CNN performed more accurately than GP.
GP was more accurate than other models in simple conditions, e.g., training each participant separately with small data sets. CNN, on the other hand, was more robust and faster when dealing with large data with more variables.

\subsection{Time cross validation}

\begin{table}
\caption{Time cross validation error.  The interval of the training and testing data recorded is 1 week. RMSE of force and torque estimation by GP, as well as the mean of the estimated standard deviation (in brackets) given by the GP are listed below.}
\begin{center}
\begin{tabular}{ccccccc}
 &  & $x$ & $y$ & $z$ &\\
\hline\\[-0.5ex]
\multirow{3}{*}{\centering $P_3$} & $f$    & 0.778(0.500)  & 0.197(0.179)   & 1.33(1.46) \\
                                  & $\tau$ & 3.86(4.16)   & 2.81(2.86)    & 3.80(3.29) \\
\hline
\label{table2}
\end{tabular}
\end{center}
\end{table}

\begin{figure}
\centering
  \subfigure[RMS error in force components $f_x$, $f_y$, $f_z$.]{
    \begin{minipage}[b]{0.47\columnwidth}
      \centering
      \includegraphics[width=\columnwidth]{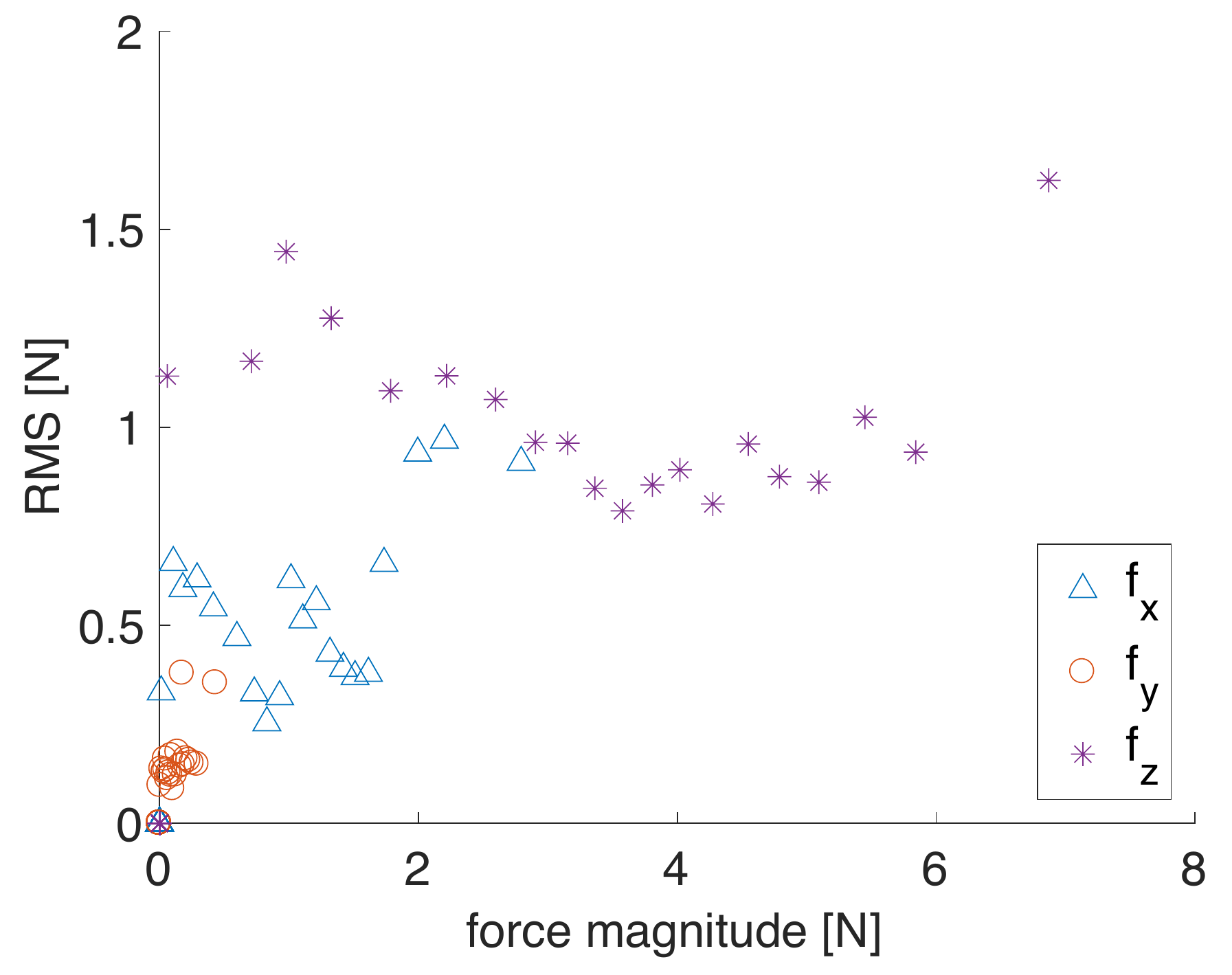}
    \end{minipage}}\hfill
  \subfigure[RMS error in predicted force length magnitude.]{
    \begin{minipage}[b]{0.47\columnwidth}
      \centering
      \includegraphics[width=\columnwidth]{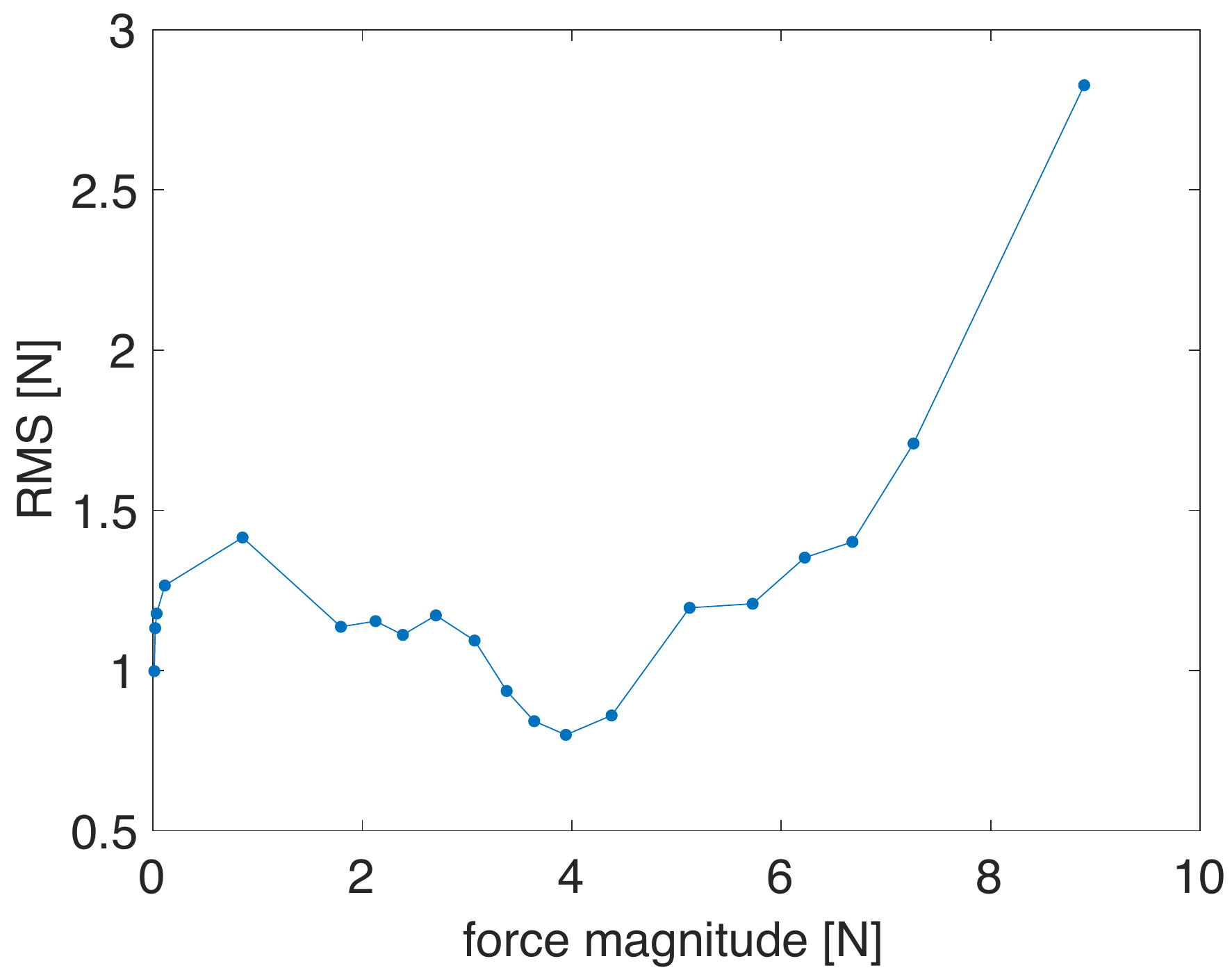}
    \end{minipage}}\\
  \subfigure[RMS error in angle between predicted and surface normal ($f_z$).]{
    \begin{minipage}[b]{0.47\columnwidth}
      \centering
      \includegraphics[width=\columnwidth]{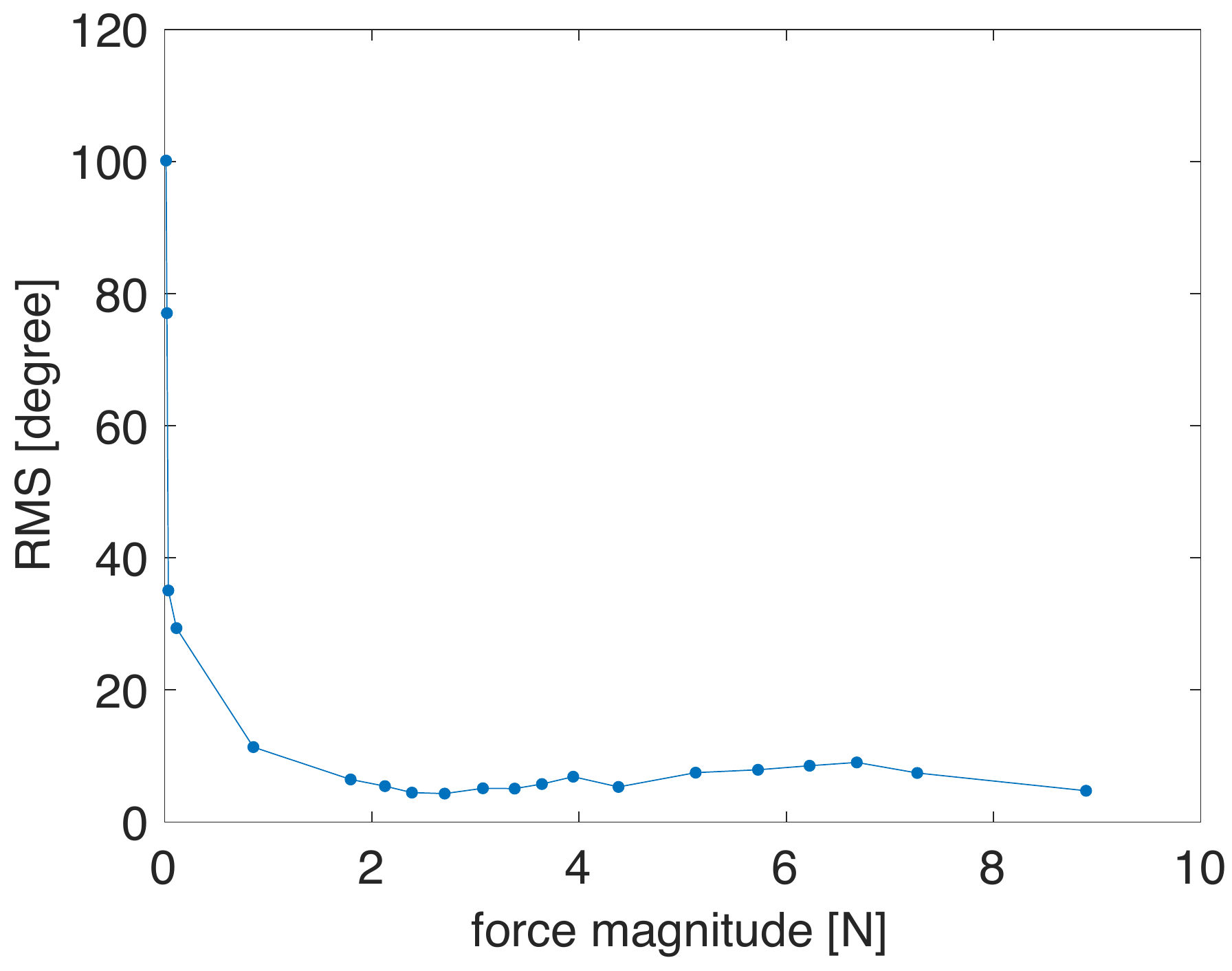}
    \end{minipage}}\hfill
  \subfigure[RMS error in Normal:Tangential force ratio.]{
    \begin{minipage}[b]{0.47\columnwidth}
      \centering
      \includegraphics[width=\columnwidth]{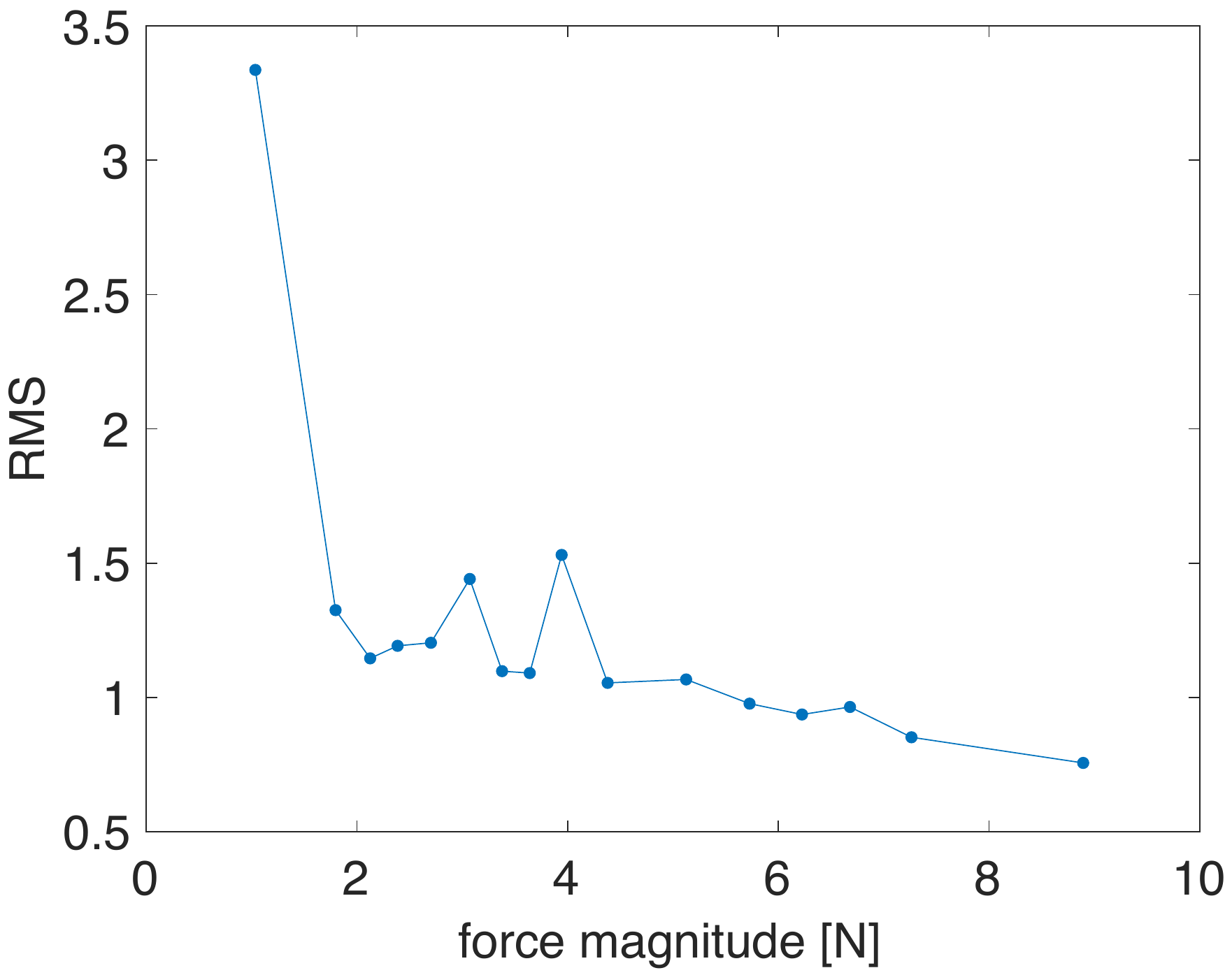}
    \end{minipage}}\\
  \subfigure[RMS error in torque components $\tau_x$, $\tau_y$, $\tau_z$.]{
    \begin{minipage}[b]{0.47\columnwidth}
      \centering
      \includegraphics[width=\columnwidth]{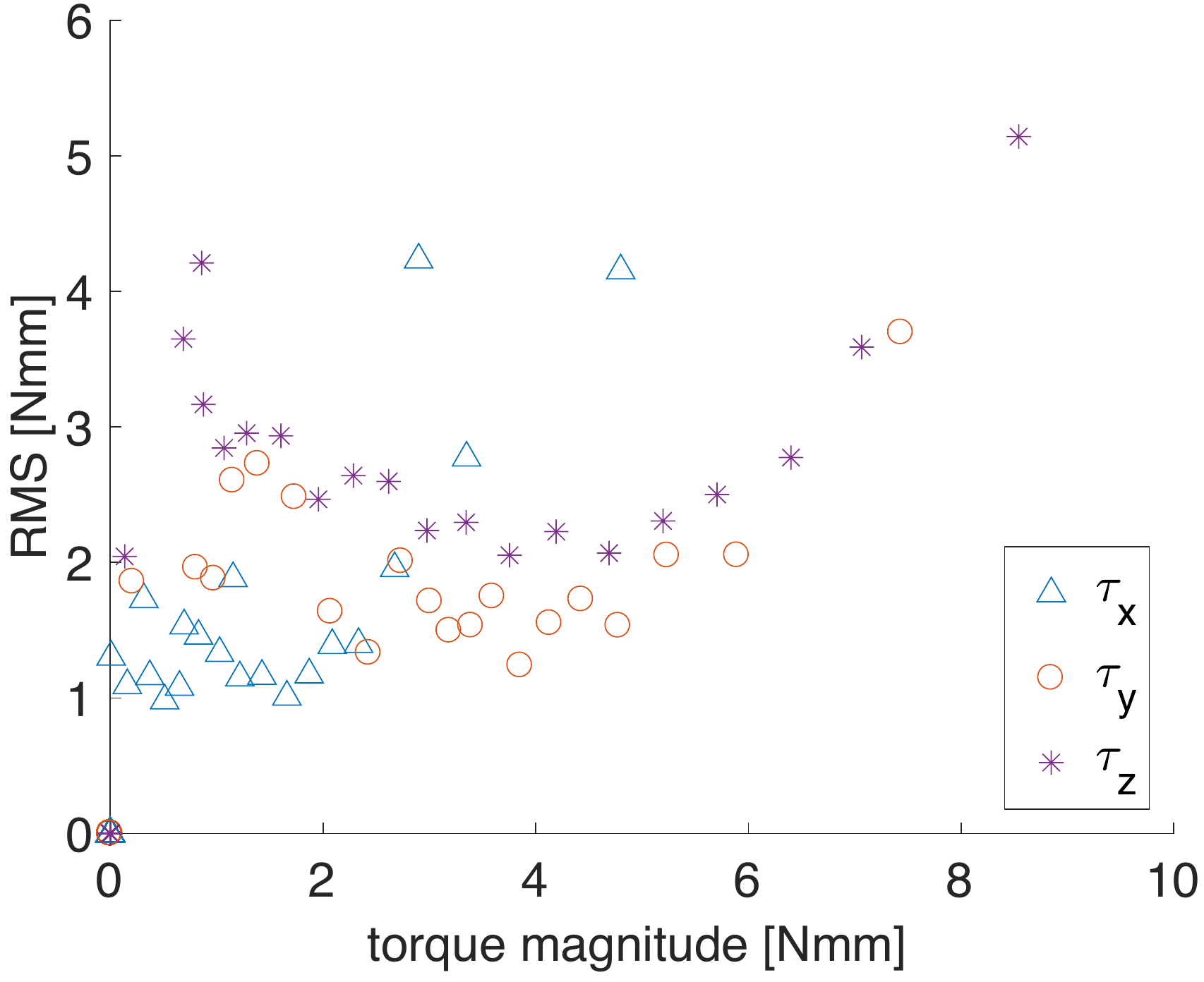}
    \end{minipage}}s\\
    \vspace{0 in}
  \caption{Predictions for participant $P_5$ for times cross validation. The interval of the training and testing data recorded wass 1 week. Annotations as in Fig.~\ref{method_compar1}.}
  \label{plot_times_cross}
\end{figure}

Variables such as finger temperature and lighting varied, of course, over time. To investigate the impact of such variations, training and testing was spaced 1 week for participant $P_5$. The result from the GP predictions are shown in 
Table~\ref{table2} and Fig.~\ref{plot_times_cross}.

\subsection{Participant cross validation}
In this experiment, we took participant $P_3$ as the testing dataset, and the other four participants as the training and/or the validation data sets. In this situation, CNN could predict whether the force increased or decreased, but could not accurately estimate the quantity of the forces: GP generated constant outputs. CNN functioned relatively well regardless of changes to lighting, rotation and shift of the images. In contrast, GP was sensitive to these variables and had difficulties when these variables were never shown in the training data. In addition, the fingernail color distribution varied between participants and was the main reason for the inaccurate predictions of both methods.

\subsection{Human grasping analysis}
\label{application}

\begin{figure}
\centering
  \subfigure[Grasping the object with flat sandpaper surface.]{
    \begin{minipage}[b]{\columnwidth}
      \centering
      \includegraphics[width=0.45\columnwidth]{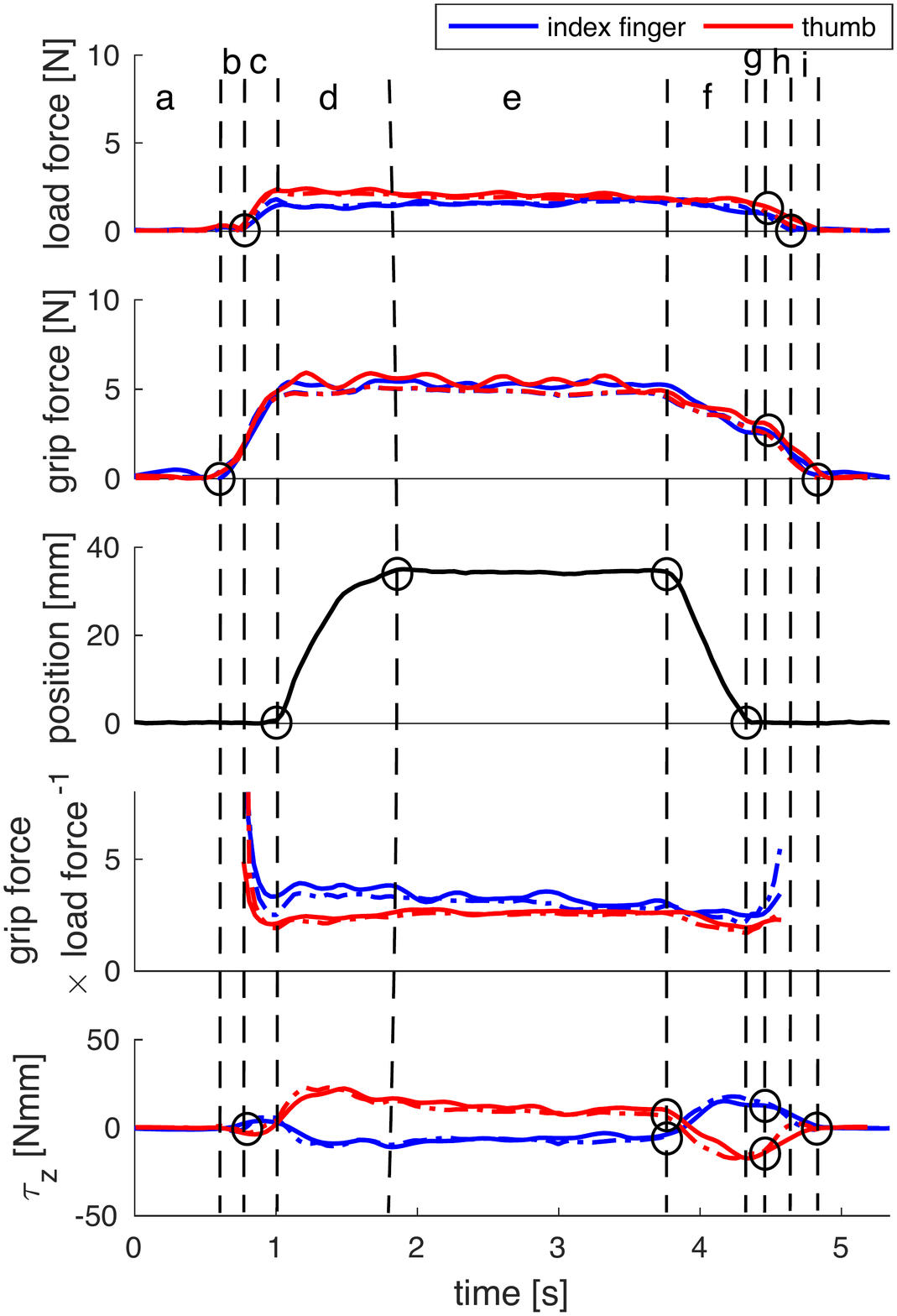}
    \end{minipage}}  \\
 \hfill
  \subfigure[Grasping the object with flat silk surface.]{
    \begin{minipage}[b]{\columnwidth}
      \centering
      \includegraphics[width=0.45\columnwidth]{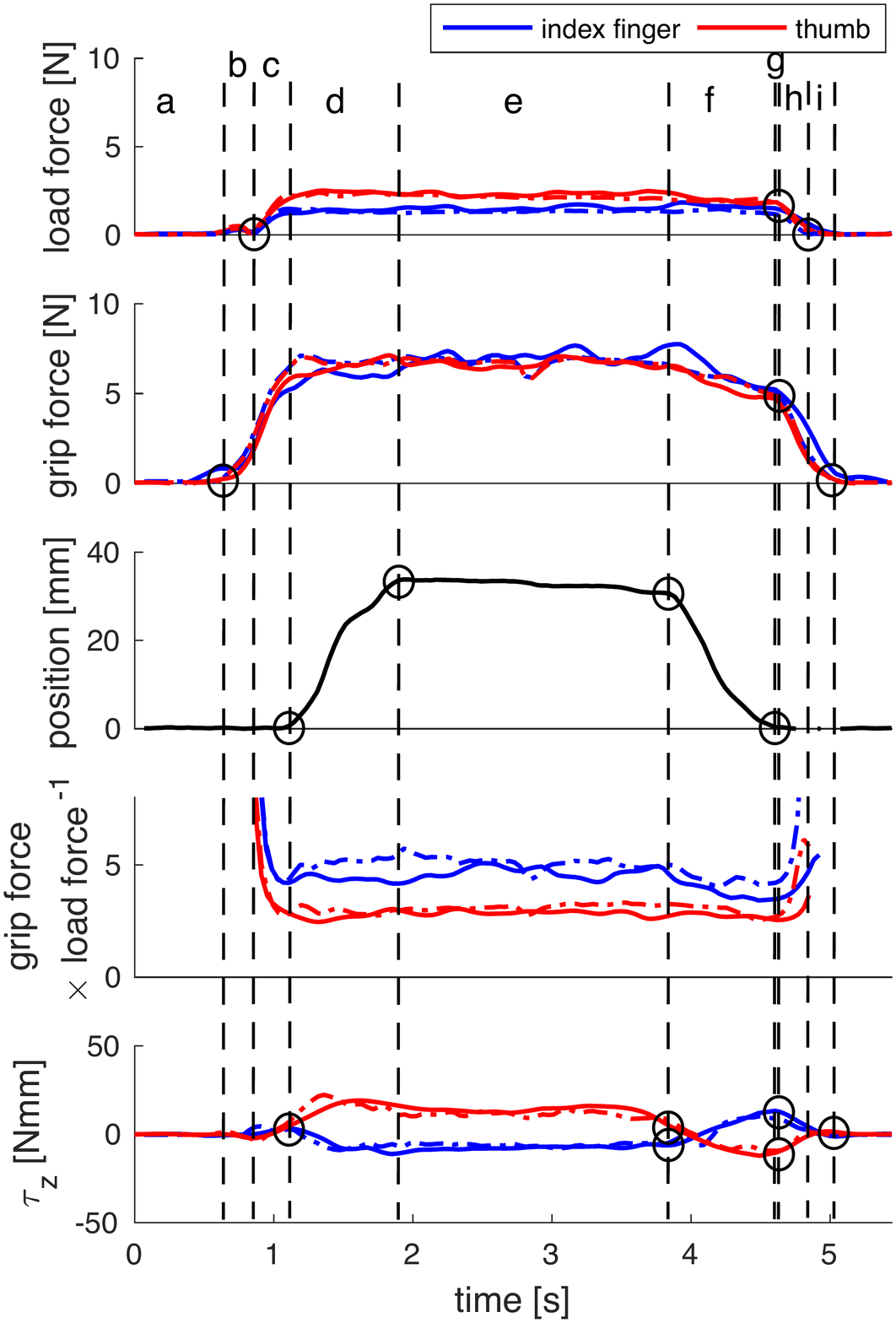}
    \end{minipage}}\\
    \vspace{0 in}
\caption{Grasping analysis of $P_1$ with 330\,g weight. The dashed lines are the ground truth and the solid lines are the prediction. The load force is the force length magnitude of $f_x$ and $f_y$ and the grip force is $f_z$. The phases of the trials \cite{Johansson240765} include: a - reaching phase; b -  preload phase; c - loading phase; d - transitional phase; e - static phase; f - replacement phase; g - delay; h - pre-unload phase; i - unloading phase. h and i phases are possible integrated into one phase, which depends on the placing behaviour of the participants. With the decreasing of the friction, the ratio (grip force $\times$ load force$^{-1}$) increases to avoid slipping. Subfigures in the third row are the position of the object.}
  \label{pick_replace_material}
\end{figure}

\begin{figure}
\centering
  \subfigure[Grasping of $P_1$ the object with 165\,g weight.]{
    \begin{minipage}[b]{0.40\columnwidth}
      \centering
      \includegraphics[width=\columnwidth]{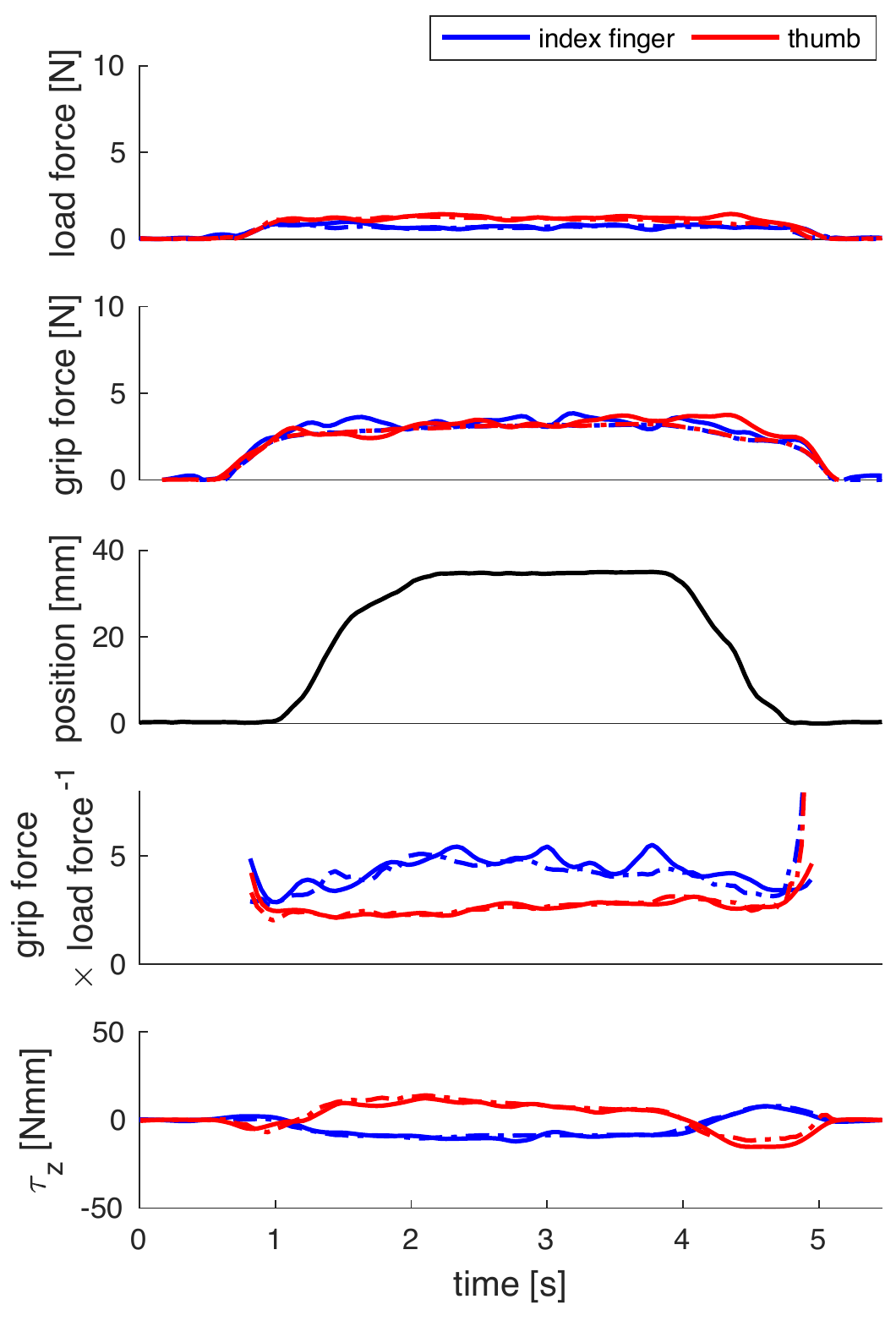}
    \end{minipage}}\hfill
  \subfigure[Grasping of $P_1$ the object with 330\,g weight.]{
    \begin{minipage}[b]{0.40\columnwidth}
      \centering
      \label{pick_place_330}
      \includegraphics[width=\columnwidth]{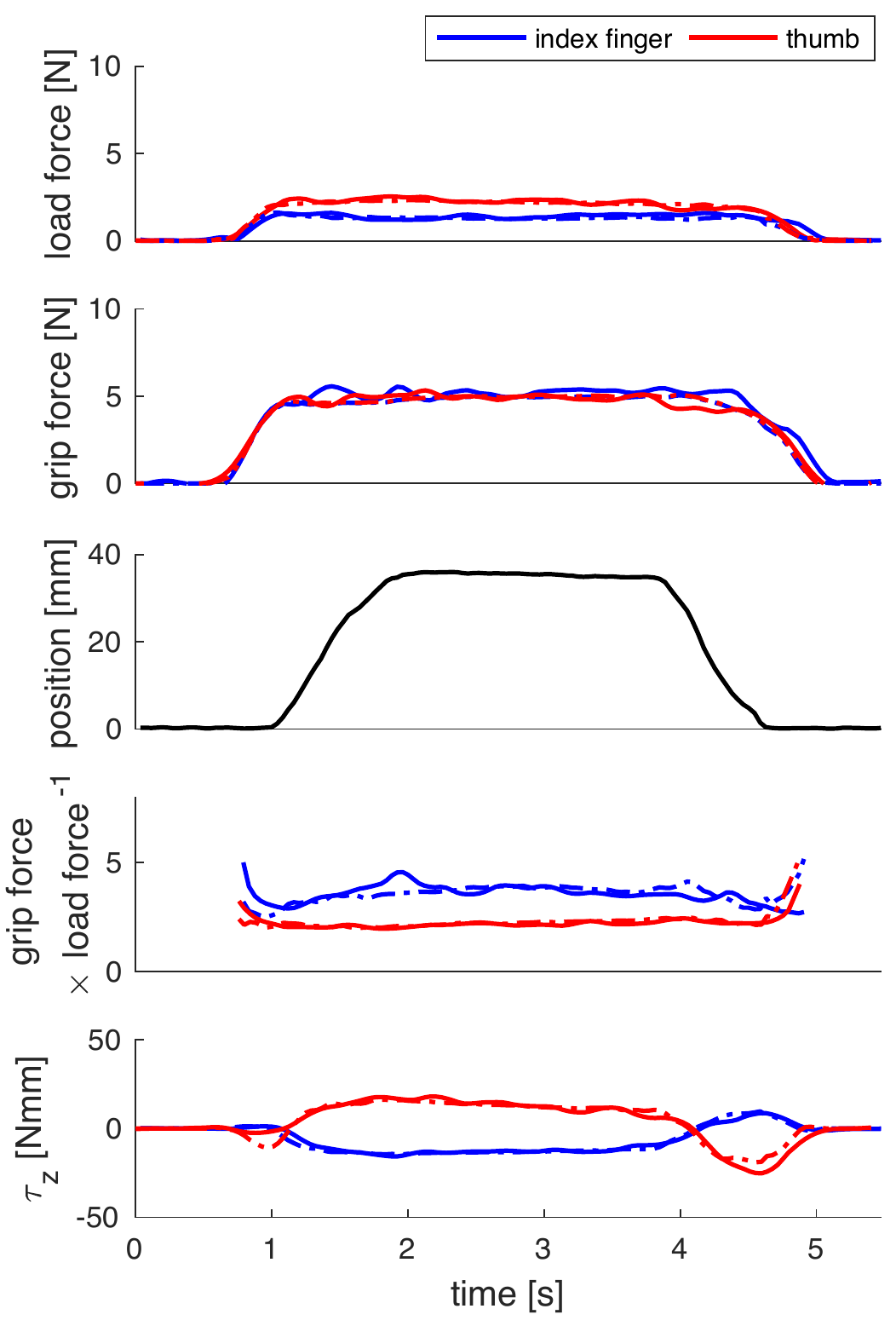}
    \end{minipage}}\\
  \subfigure[Grasping of $P_1$ the object with surface 660\,g weight.]{
    \begin{minipage}[b]{0.40\columnwidth}
      \centering
      \includegraphics[width=\columnwidth]{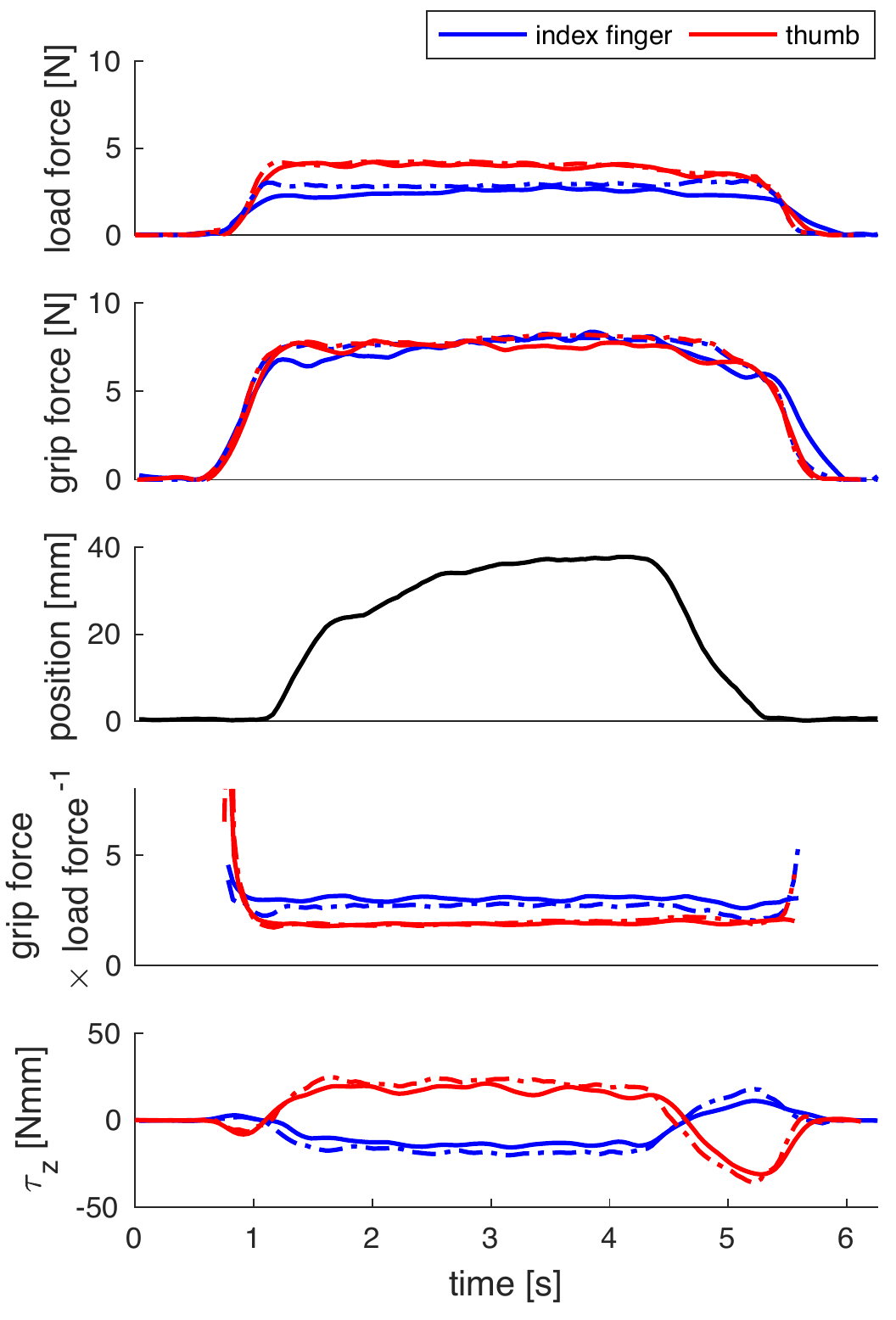}
    \end{minipage}}\hfill
  \subfigure[Grasping of $S_2$ the object with 660\,g weight.]{
    \begin{minipage}[b]{0.40\columnwidth}
      \centering
      \label{pick_place_330}
      \includegraphics[width=\columnwidth]{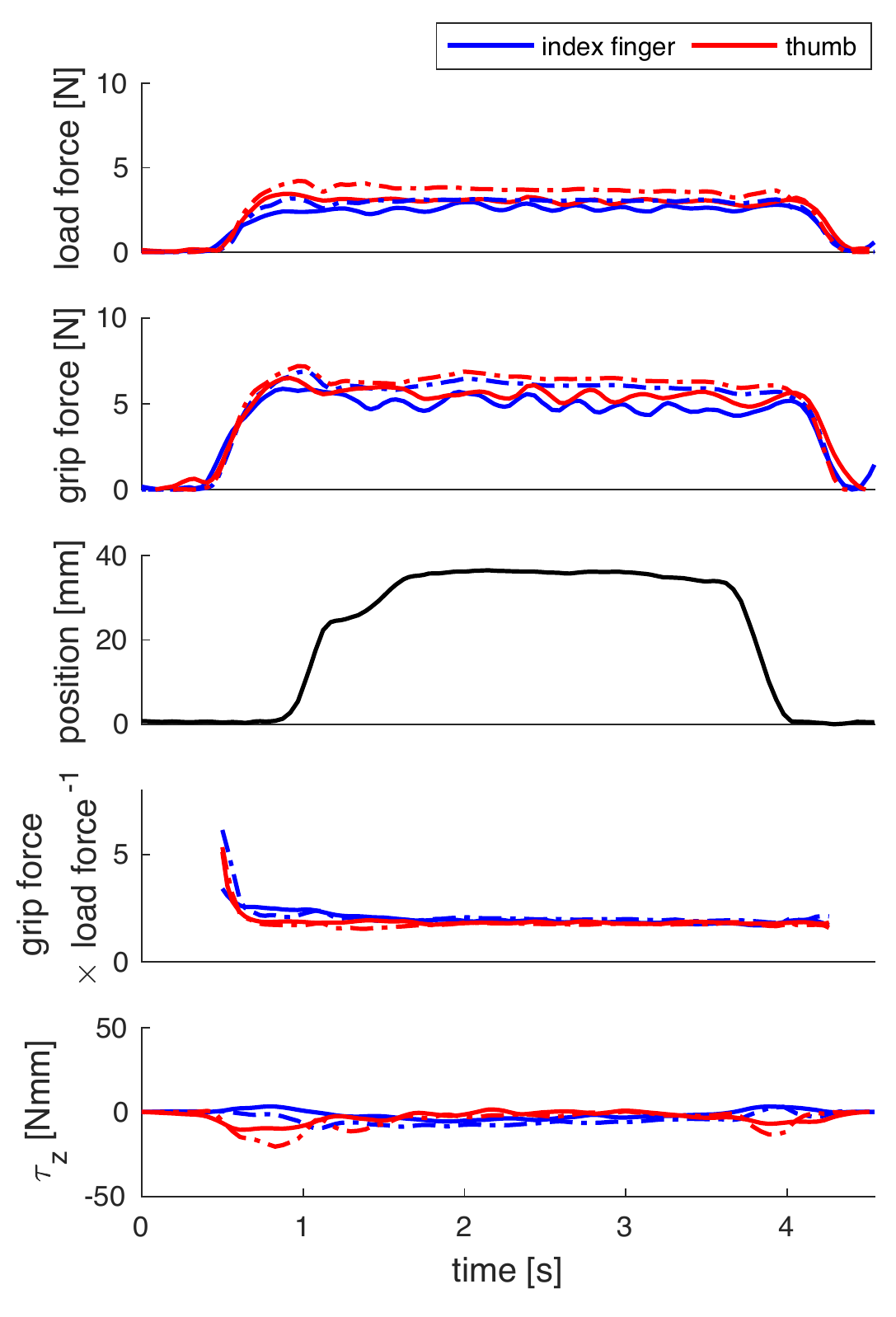}
    \end{minipage}}\\
    \vspace{0 in}
\caption{Grasping analysis of different participants with surface \,2 (cf., Fig.~\ref{setup04}) and different object weights.  The sum of the load force of the thumb and index finger during the hold phase corresponded approximately to the force required hold the object in air. The difference of the load force of the thumb and the index finger was caused by the inclination of the object. The grip forces of the thumb and the index finger were very similar to each other. For more annotations, see Fig.~\ref{pick_replace_material}. } 
  \label{pick_replace_}
\end{figure}

The approach can be used in settings such as robot teleoperation, force-based control and so on. As an example of the applications, we analyzed human grasping according to \cite{Johansson240765} using the result of our predictor.

For observation, Fig.~\ref{pick_replace_material},\ref{pick_replace_} only shows $f_x$, $f_z$ and $\tau_z$, which are most significant for picking up and replacing in our experiments. They were evaluated using partidcipant $P_1$ with two grasping surfaces in Fig.~\ref{pick_replace_material} and was evaluated for $P_1$ and $P_2$ grasping surface\,3 in Fig.~\ref{pick_replace_}. From the figures, grasping can be seen in six steps: reaching, loading, lifting, holding, replacing and unloading. Other surfaces have similar patterns for the steps.
Consider the index finger as an example: the loading starts from the critical near zero force. Lifting occurred from $f_x$ peak to torque peak. While the object was held stationary both forces and torques stable. Replacing occurred from torque peak until $f_x$ peak and unloading from the lowest point of torque to zero forces.

%% file: sections/Conclusions.tex
\section{Discussions and Conclusions}
\label{sec:3}

This paper presents a method measuring finger contact force, torque, and surface curvature using the deformation and color distribution of the fingernail.
Moving away from constrained laboratory settings with perfect conditions and restrictions, we were able to capture fingernail images when subjects made contact with varied surfaces. After conducting non-rigid finger image alignments, we trained force/torque and contact surface prediction models using the aligned images. The results show that the models accurately predicted the force/torque used for picking up and replacing an object for multiple fingers across multiple humans. The models were evaluated across surfaces and variables (e.g., finger temperature and lighting).

We compared four machine learning methods on their ability to predict force, torque and contact surface.
GP and CNN performed best out of the four methods. In simple conditions with relative small data sets, GP performed slightly better than CNN. In more complicated conditions such as cross participant validation or with a large set of training data, the CNN yielded more accurate results than the GP, and the CNN was considerably faster.

The current data processing method run at approximately 30\,Hz for an image resolution of $111\times105$ pixels, using an off-the-shelf GPU processor. Speedup would be possible by, e.g., reducing the image resolution or using more advanced hardware.

In the future, we will explore the ability of a more robust model to predict fingertip forces across participants never shown in the training data. A large data set of different subjects would probably increase the accuracy.